%% file: rotp_tpami_v0.tex
\documentclass[10pt,journal,compsoc]{IEEEtran}
\ifCLASSOPTIONcompsoc
  \usepackage[nocompress]{cite}
\else
  \usepackage{cite}
\fi
\ifCLASSINFOpdf
\else
\fi

\usepackage{times}  % DO NOT CHANGE THIS
\usepackage{helvet} % DO NOT CHANGE THIS
\usepackage{courier}  % DO NOT CHANGE THIS
\usepackage[hyphens]{url}  % DO NOT CHANGE THIS
\usepackage{graphicx} % DO NOT CHANGE THIS
\usepackage{graphicx}  % DO NOT CHANGE THIS
\usepackage{amsmath}
\usepackage{amssymb}
\usepackage{amsthm}
\usepackage{bm}
\usepackage{subfigure} 
\usepackage{multirow}
\usepackage{color}
\usepackage{url}
\usepackage{amsfonts}
\usepackage{mathtools}

\usepackage{algorithm}
\usepackage{algorithmic}
\newtheorem{theorem}{Theorem}

\newtheorem{proposition}{Proposition}

\usepackage{threeparttable}
\usepackage{enumitem}

\begin{document}
%
% paper title
% Titles are generally capitalized except for words such as a, an, and, as,
% at, but, by, for, in, nor, of, on, or, the, to and up, which are usually
% not capitalized unless they are the first or last word of the title.
% Linebreaks \\ can be used within to get better formatting as desired.
% Do not put math or special symbols in the title.
\title{Regularized Optimal Transport Layers for Generalized Global Pooling Operations}
%
%
% author names and IEEE memberships
% note positions of commas and nonbreaking spaces ( ~ ) LaTeX will not break
% a structure at a ~ so this keeps an author's name from being broken across
% two lines.
% use \thanks{} to gain access to the first footnote area
% a separate \thanks must be used for each paragraph as LaTeX2e's \thanks
% was not built to handle multiple paragraphs
%
%
%\IEEEcompsocitemizethanks is a special \thanks that produces the bulleted
% lists the Computer Society journals use for "first footnote" author
% affiliations. Use \IEEEcompsocthanksitem which works much like \item
% for each affiliation group. When not in compsoc mode,
% \IEEEcompsocitemizethanks becomes like \thanks and
% \IEEEcompsocthanksitem becomes a line break with idention. This
% facilitates dual compilation, although admittedly the differences in the
% desired content of \author between the different types of papers makes a
% one-size-fits-all approach a daunting prospect. For instance, compsoc 
% journal papers have the author affiliations above the "Manuscript
% received ..."  text while in non-compsoc journals this is reversed. Sigh.

\author{Hongteng~Xu,~\IEEEmembership{Member,~IEEE},~
        Minjie~Cheng% <-this % stops a space
\IEEEcompsocitemizethanks{
\IEEEcompsocthanksitem Hongteng Xu was with the Gaoling School of Artificial Intelligence, Renmin University of China and Beijing Key Laboratory of Big Data Management and Analysis Methods.\protect\\
E-mail: hongtengxu@ruc.edu.cn
\IEEEcompsocthanksitem Minjie Cheng was with the Gaoling School of Artificial Intelligence, Renmin University of China.\protect\\
E-mail: chengminjie@ruc.edu.cn
\IEEEcompsocthanksitem The two authors contributed equally to this work.
}
\thanks{Manuscript received XX XX, 20XX; revised XX XX, 20XX.}}

% note the % following the last \IEEEmembership and also \thanks - 
% these prevent an unwanted space from occurring between the last author name
% and the end of the author line. i.e., if you had this:
% 
% \author{....lastname \thanks{...} \thanks{...} }
%                     ^------------^------------^----Do not want these spaces!
%
% a space would be appended to the last name and could cause every name on that
% line to be shifted left slightly. This is one of those "LaTeX things". For
% instance, "\textbf{A} \textbf{B}" will typeset as "A B" not "AB". To get
% "AB" then you have to do: "\textbf{A}\textbf{B}"
% \thanks is no different in this regard, so shield the last } of each \thanks
% that ends a line with a % and do not let a space in before the next \thanks.
% Spaces after \IEEEmembership other than the last one are OK (and needed) as
% you are supposed to have spaces between the names. For what it is worth,
% this is a minor point as most people would not even notice if the said evil
% space somehow managed to creep in.

% The paper headers
\markboth{Journal of \LaTeX\ Class Files,~Vol.~XX, No.~X, XX~20XX}%
{Shell \MakeLowercase{\textit{et al.}}: Bare Demo of IEEEtran.cls for Computer Society Journals}
% The only time the second header will appear is for the odd numbered pages
% after the title page when using the twoside option.
% 
% *** Note that you probably will NOT want to include the author's ***
% *** name in the headers of peer review papers.                   ***
% You can use \ifCLASSOPTIONpeerreview for conditional compilation here if
% you desire.

% The publisher's ID mark at the bottom of the page is less important with
% Computer Society journal papers as those publications place the marks
% outside of the main text columns and, therefore, unlike regular IEEE
% journals, the available text space is not reduced by their presence.
% If you want to put a publisher's ID mark on the page you can do it like
% this:
%\IEEEpubid{0000--0000/00\$00.00~\copyright~2015 IEEE}
% or like this to get the Computer Society new two part style.
%\IEEEpubid{\makebox[\columnwidth]{\hfill 0000--0000/00/\$00.00~\copyright~2015 IEEE}%
%\hspace{\columnsep}\makebox[\columnwidth]{Published by the IEEE Computer Society\hfill}}
% Remember, if you use this you must call \IEEEpubidadjcol in the second
% column for its text to clear the IEEEpubid mark (Computer Society jorunal
% papers don't need this extra clearance.)

% use for special paper notices
%\IEEEspecialpapernotice{(Invited Paper)}

% for Computer Society papers, we must declare the abstract and index terms
% PRIOR to the title within the \IEEEtitleabstractindextext IEEEtran
% command as these need to go into the title area created by \maketitle.
% As a general rule, do not put math, special symbols or citations
% in the abstract or keywords.
\IEEEtitleabstractindextext{%
\begin{abstract}
Global pooling is one of the most significant operations in many machine learning models and tasks, which works for information fusion and structured data (like sets and graphs) representation. 
However, without solid mathematical fundamentals, its practical implementations often depend on empirical mechanisms and thus lead to sub-optimal, even unsatisfactory performance. 
In this work, we develop a novel and generalized global pooling framework through the lens of optimal transport. 
The proposed framework is interpretable from the perspective of expectation-maximization. 
Essentially, it aims at learning an optimal transport across sample indices and feature dimensions, making the corresponding pooling operation maximize the conditional expectation of input data. 
We demonstrate that most existing pooling methods are equivalent to solving a regularized optimal transport (ROT) problem with different specializations, and more sophisticated pooling operations can be implemented by hierarchically solving multiple ROT problems.
Making the parameters of the ROT problem learnable, we develop a family of regularized optimal transport pooling (ROTP) layers. 
We implement the ROTP layers as a new kind of deep implicit layer. 
Their model architectures correspond to different optimization algorithms. 
We test our ROTP layers in several representative set-level machine learning scenarios, including multi-instance learning (MIL), graph classification, graph set representation, and image classification. 
Experimental results show that applying our ROTP layers can reduce the difficulty of the design and selection of global pooling --- our ROTP layers may either imitate some existing global pooling methods or lead to some new pooling layers fitting data better. 
The code is available at \url{https://github.com/SDS-Lab/ROT-Pooling}.
\end{abstract}

% Note that keywords are not normally used for peerreview papers.
}

% make the title area
\maketitle

% To allow for easy dual compilation without having to reenter the
% abstract/keywords data, the \IEEEtitleabstractindextext text will
% not be used in maketitle, but will appear (i.e., to be "transported")
% here as \IEEEdisplaynontitleabstractindextext when the compsoc 
% or transmag modes are not selected <OR> if conference mode is selected 
% - because all conference papers position the abstract like regular
% papers do.
\IEEEdisplaynontitleabstractindextext
% \IEEEdisplaynontitleabstractindextext has no effect when using
% compsoc or transmag under a non-conference mode.

% For peer review papers, you can put extra information on the cover
% page as needed:
% \ifCLASSOPTIONpeerreview
% \begin{center} \bfseries EDICS Category: 3-BBND \end{center}
% \fi
%
% For peerreview papers, this IEEEtran command inserts a page break and
% creates the second title. It will be ignored for other modes.
\IEEEpeerreviewmaketitle

\IEEEraisesectionheading{\section{Introduction}\label{sec:introduction}}
\IEEEPARstart{A}{s} a fundamental operation of information fusion and structured data representation, global pooling achieves a global representation for a set of inputs.
It makes the representation invariant to the permutation of the inputs. 
This operation has been widely used in many set-level machine learning tasks. 
In multi-instance learning (MIL) tasks~\cite{yan2018deep,ilse2018attention}, we often leverage a global pooling operation to aggregate multiple instances into a bag-level representation.
In graph representation tasks, after passing a graph through a graph neural network, we often apply a global pooling operation (or called ``readout'') to merge its node embeddings into a global graph embedding~\cite{ying2018hierarchical,xu2018powerful}. 
Besides these two representative cases, pooling layers are also necessary for convolutional neural networks (CNN) when extracting visual features~\cite{krizhevsky2012imagenet,he2016deep}. 
For the data with multi-scale clustering structures~\cite{yu2018hierarchical,pan2021scalable}, we can stack multiple pooling layers and derive a hierarchical pooling operation accordingly.

\begin{figure*}[t]
    \centering
    \subfigure[Regularized optimal transport pooling (ROTP) layer]{
    \includegraphics[height=5cm]{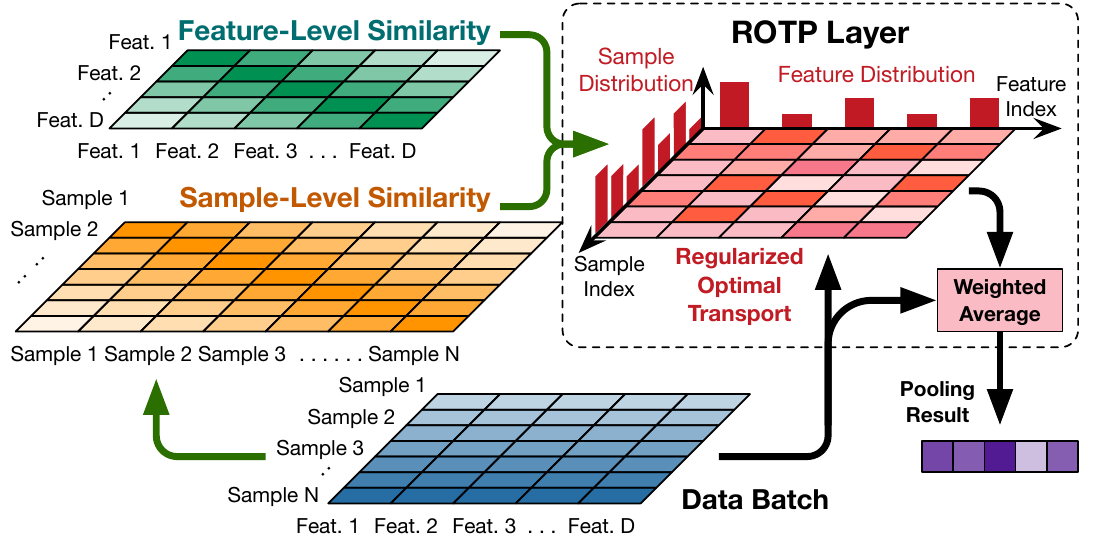}\label{fig:scheme_a}
    }
    \hspace{5mm}
    \subfigure[HROTP module]{
    \includegraphics[height=5cm]{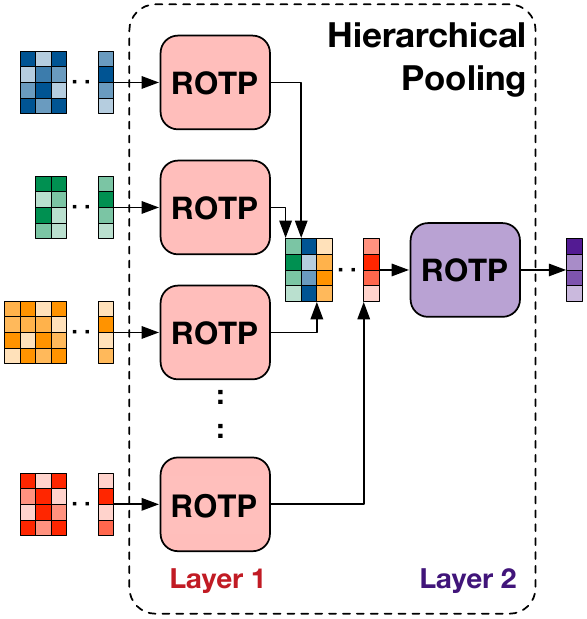}\label{fig:scheme_b}
    }
    \caption{(a) An illustration of the proposed regularized optimal transport pooling layer (ROTP). 
    Here, the green arrow indicates the optional usage of the side information like the feature-level and sample-level similarities.
    (b) An illustration of the hierarchical ROTP (HROTP) module constructed by integrating multiple ROTP layers.
    Here, the ROTP layers with different parameters are labeled by different colors.}\label{fig:scheme}
\end{figure*}

Currently, many different kinds of global pooling operations have been proposed. 
Simple operations, like add-pooling, mean-pooling (or called average-pooling), and max-pooling~\cite{boureau2010theoretical}, are commonly used because of their computational efficiency. 
The mixture~\cite{lee2016generalizing} and the concatenation~\cite{liu2016cross,you2021mc} of these simple pooling operations are also considered to improve their performance. 
More recently, some pooling methods, e.g., Network-in-Network (NIN)~\cite{lin2013network}, Set2Set~\cite{vinyals2015order}, DeepSet~\cite{zaheer2017deep}, dynamic pooling~\cite{yan2018deep}, and attention-based pooling layers~\cite{ilse2018attention,lee2019set}, are developed with learnable parameters and more sophisticated mechanisms. 
Although the above global pooling methods have been widely used in many machine learning models and tasks, their theoretical study is far lagged-behind.
In particular, the principles of these methods are not well-interpreted in Statistics, whose rationality and effectiveness are not supported in theory. 
Additionally, facing so many different pooling methods, the differences and the connections among them are not thoroughly investigated.
Without insightful theoretical guidance, the pooling methods' design and selection are empirical or depend on time-consuming enumerating, which often leads to poorly-generalizable models and sub-optimal performance in practice. 

To simplify the design of global pooling in an interpretable way and boost its performance in practice, in this study, we propose a novel and solid algorithmic pooling framework to unify and generalize most existing global pooling operations through the lens of optimal transport. 
As illustrated in Fig.~\ref{fig:scheme_a}, given a batch of data, the proposed pooling operation first optimizes the joint distribution of sample indices and feature dimensions and then weights and averages the representative ``sample-feature'' pairs. 
The optimization problem in the above pooling process corresponds to a regularized optimal transport (ROT) problem.
In the problem, the target joint distribution corresponds to an optimal transport (OT) plan derived under three regularizations: $i)$ the smoothness of the OT plan, $ii)$ the uncertainty of the OT plan's marginal distributions, and optionally $iii)$ the Gromov-Wasserstein (GW) discrepancy~\cite{peyre2016gromov} between the feature-level and sample-level similarity matrices. 
The above pooling operation provides a generalized pooling framework with theoretical guarantees. 
Specifically, we demonstrate that most existing pooling operations are specializations of the ROT problem under different parameter configurations. 
Moreover, the sophisticated pooling mechanisms like the mixed pooling methods in~\cite{lee2016generalizing} and the hierarchical pooling in~\cite{ying2018hierarchical} can be generalized by hierarchically solving multiple ROT problems, as shown in Fig.~\ref{fig:scheme_b}. 

Besides proposing the above unified global pooling framework, we further make the parameters of the ROT problem learnable and develop a family of regularized optimal transport pooling (\textbf{ROTP}) layers.
The ROTP layers can be treated as new members of deep implicit layers~\cite{agrawal2019differentiable,el2021implicit,huang2021textrm,bai2019deep}. 
Their model architectures correspond to different optimization methods of various ROT problems, and their backward computations can have closed-form solutions in some conditions~\cite{gould2021deep}.
In particular, we implement the ROTP layers by solving the ROT problems under different settings, including using the entropic or the quadratic smoothness term, with or without the GW discrepancy term, and so on. 
These ROT problems are solved in a proximal gradient algorithmic framework, in which each subproblem can be optimized by the Sinkhorn scaling algorithm~\cite{cuturi2013sinkhorn,pham2020unbalanced,sejourne2021unbalanced} and the Bregman alternating direction method of multipliers (Bregman ADMM, or BADMM for short)~\cite{wang2014bregman,xu2020gromov}, respectively. 
As a result, each ROTP layer unrolls the iterative optimization of the corresponding ROT problem to feed-forward computations, whose backpropagation adjusts the parameters controlling the optimization process. 
We analyze each ROTP layer's representation power, computational complexity, and numerical stability in depth and test its performance in various learning tasks. 

The contributions of our work can be summarized as follows. 
\begin{itemize}[noitemsep]
\item[$i)$] \textbf{A generalized global pooling framework with theoretical supports.} 
We propose a generalized global pooling framework, which unifies many existing pooling methods with theoretical guarantees and offers a new perspective based on regularized optimal transport.
From the viewpoint of statistical signal processing, the proposed pooling framework is interpretable, which yields an expectation-maximization principle. 
Furthermore, stacking multiple ROTP layers together leads to a hierarchical ROTP (HROTP) module shown in Fig.~\ref{fig:scheme_b}. 
We demonstrate that such a hierarchical pooling module can generalize the mixed pooling mechanism  in~\cite{lee2016generalizing} and represent the complex data with hierarchical clustering structures, e.g., the set of graphs.
\item[$ii)$] \textbf{Effective and flexible global pooling layers.}
Based on the proposed global pooling framework, we develop a family of ROTP layers and analyze them quantitatively. 
The proposed ROTP layers can be implemented under different regularizers and optimization algorithms, which have high flexibility and can be applied in various learning scenarios.
Additionally, based on the ROTP layers, we can build the HROTP module and provide new solutions to complicated operations like mixed pooling and set fusion. 
\item[$iii)$] \textbf{Universal effectiveness in various tasks.} 
We test our ROTP layers in several representative set-level machine learning tasks, including multi-instance learning (MIL), graph classification, graph set representation, and image classification.
These tasks correspond to real-world applications, such as medical image analysis, molecule classification, drug-drug interaction analysis, and the ImageNet challenge. 
In each task, our ROTP layers can imitate or outperform state-of-the-art global pooling methods, which simplify the design and the selection of global pooling operations in practice. 
\end{itemize}

The remainder of this paper is organized as follows:
Section~\ref{sec:related} provides a detailed literature review and explains the connections and differences between our work and existing methods. 
Section~\ref{sec:model} introduces the proposed ROTP framework and demonstrates its rationality and generalization power. 
Additionally, this section introduces the hierarchical ROTP module for mixed pooling and set fusion. 
Section~\ref{sec:alg} introduces the ROTP layers under different settings and analyzes their computational complexity and numerical stability. 
Section~\ref{sec:cmp} analyzes the generalization power, computational complexity, and numerical stability of different ROTP layers and compares them with other optimal transport-based pooling methods.
Section~\ref{sec:exp} provides the implementation details of our methods and experimental results on multiple datasets. 
Finally, Section~\ref{sec:con} concludes the paper and discusses our future work. 

\input{tex/related_work.tex}

\input{tex/models.tex}

\input{tex/algorithms.tex}

\input{tex/compare.tex}

\input{tex/experiments.tex}

\section{Conclusion and Future Work}\label{sec:con}
This study proposes a generalized pooling framework driven by the regularized optimal transport problem. 
We demonstrate that many existing pooling operations correspond to solving the ROT problem with different configurations. 
By learning the parameters of the ROT problem, we obtain an ROTP layer and propose three implementations based on different settings. 
For each implementation of the ROTP layer, we analyze its in-depth on its stability and complexity. 
Stacking the ROTP layers leads to a hierarchical pooling module for set fusion.
Our work provides a solid and effective pooling framework with theoretical support and statistical interpretability. 
Experiments on practical learning tasks and real-world datasets demonstrate the usefulness of our ROTP layers. 
In the future, we consider applying our regularized optimal transport modules to reformulate other machine learning models, e.g., the local pooling layers in convolutional neural networks and message passing layers in graph neural networks. 
Additionally, as aforementioned, we plan to develop a CUDA version of the ROTP layer to improve its computational efficiency further.

\bibliographystyle{IEEEtran}
\bibliography{refs}
% biography section
% 
% If you have an EPS/PDF photo (graphicx package needed) extra braces are
% needed around the contents of the optional argument to biography to prevent
% the LaTeX parser from getting confused when it sees the complicated
% \includegraphics command within an optional argument. (You could create
% your own custom macro containing the \includegraphics command to make things
% simpler here.)
%\begin{IEEEbiography}[{\includegraphics[width=1in,height=1.25in,clip,keepaspectratio]{mshell}}]{Michael Shell}
% or if you just want to reserve a space for a photo:
%\vspace{-2cm}
\begin{IEEEbiography}[{\includegraphics[width=1in,height=1.25in,clip,keepaspectratio]{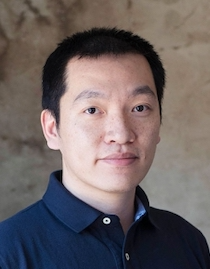}}]{Hongteng Xu}
is an Associate Professor (Tenure-Track) in the Gaoling School of Artificial Intelligence, Renmin University of China. 
From 2018 to 2020, he was a senior research scientist in Infinia ML Inc. 
In the same time period, he is a visiting faculty member in the Department of Electrical and Computer Engineering, Duke University. 
He received his Ph.D. from the School of Electrical and Computer Engineering at Georgia Institute of Technology (Georgia Tech) in 2017. 
His research interests include machine learning and its applications, especially optimal transport theory, sequential data modeling and analysis, deep learning techniques, and their applications in computer vision and data mining.
\end{IEEEbiography}
\begin{IEEEbiography}[{\includegraphics[width=1in,height=1.25in,clip,keepaspectratio]{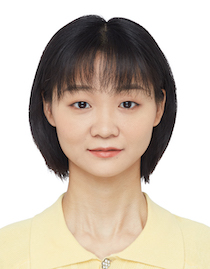}}]{Minjie Cheng}
received her B.E. degree of computer science and technology from Zhengzhou University, China, in 2016, and the M.E. degree of software engineering from Beijing University of Chemical Technology, China, in 2021. She is currently a Ph.D. student in Gaoling School of Artificial Intelligence, Renmin University of China. Her current research interests include machine learning and its applications to biochemical data analysis and modeling.
\end{IEEEbiography}
% \begin{IEEEbiography}[{\includegraphics[width=1in,height=1.25in,clip,keepaspectratio]{figures/dixin_luo.png}}]{Dixin Luo}
% received her bachelor’s and PhD degrees from Shanghai Jiao Tong University in 2010 and 2016, respectively. 
% From 2016 to 2020, she was a postdoctoral researcher in University of Toronto and Duke University. 
% Currently, she is an Assistant Professor in the School of Computer Science and Technology, Beijing Institute of Technology. 
% Her research interests include machine learning and its applications to data mining, sequential data analysis, graph analysis and healthcare.
% \end{IEEEbiography}

\end{document}

%% file: tex/related_work.tex
\section{Related Work}\label{sec:related}
\subsection{Pooling operations}
Most existing models empirically leverage simple pooling operations like add-pooling, mean-pooling, and max-pooling for convenience, whose practical performance is often sub-optimal. 
Many efforts have been made to achieve better pooling performance, which can be coarsely categorized into two strategies. 
The first strategy is applying multiple simple pooling operations jointly, e.g., concatenating the output of mean-pooling with that of max-pooling~\cite{liu2016cross,you2021mc}. 
In~\cite{lee2016generalizing}, the mixed mean-max pooling and its structured variants leverage mixture models of mean-pooling and max-pooling to improve pooling results. 
Recently, a generalized norm-based pooling (GNP) is proposed in~\cite{ko2021learning,gulcehre2014learned}. 
It can imitate max-pooling and mean-pooling under different settings.
By learning its parameters, the GNP achieves a mixture of max-pooling and mean-pooling in a nonlinear way.

The second strategy is designing pooling operations with cutting-edge neural network architectures and empowering them with additional feature transformation and extraction abilities. 
The early work following this strategy includes the Network-in-Network in~\cite{lin2013network} and the Set2Set in~\cite{vinyals2015order}, which integrate neural networks like classic multi-layer perceptrons (MLPs) and recurrent neural networks (RNNs) into pooling operations.
Recently, attention-pooling and its gated version merge multiple instances with the help of different self-attentive mechanisms~\cite{ilse2018attention}. 
Based on a similar idea, the dynamic-pooling in~\cite{yan2018deep}  applies an iterative adjustment step to improve the self-attentive mechanism.
More recently, the DeepSet in~\cite{zaheer2017deep}, the SetTransformer in~\cite{lee2019set}, and the prototype-oriented set representer in~\cite{dan2021learning} leverage and modify advanced transformer modules~\cite{vaswani2017attention} to achieve sophisticated pooling operations. 
Besides the above global pooling methods, some attempts have been made to leverage graph structures to achieve pooling operations, e.g., the DiffPooling in~\cite{ying2018hierarchical}, the ASAP in~\cite{ranjan2020asap}, and the self-attentive graph pooling (SAGP) in~\cite{lee2019self}.

Different from the above methods, we study the design of pooling operation through the lens of computational optimal transport~\cite{peyre2019computational} and propose a novel and solid algorithmic framework to unify many representative global pooling methods. 
The neural network-based implementations of our framework can be interpreted as solving a regularized optimal transport problem with learnable parameters. 
As a result, instead of designing and selecting global pooling operations empirically, we can apply our ROTP layers to approximate suitable global pooling layers automatically according to observed data or achieve new pooling mechanisms with better performance.

\subsection{Optimal transport-based machine learning}
Optimal transport (OT) theory~\cite{villani2008optimal} has proven to be useful in machine learning tasks, e.g., distribution matching~\cite{frogner2015learning,courty2016optimal,fatras2021unbalanced}, data clustering~\cite{agueh2011barycenters,cuturi2014fast}, and generative modeling~\cite{arjovsky2017wasserstein,tolstikhin2018wasserstein}. 
Given the samples of two distributions, the discrete OT problem aims at learning a joint distribution of the samples (a.k.a., the optimal transport plan) and indicating the correspondence between them accordingly. 
This discrete OT problem is a linear programming problem~\cite{kusner2015word}. 
By adding an entropic regularizer~\cite{cuturi2013sinkhorn}, the problem becomes strictly convex and can be solved efficiently by the Sinkhorn scaling in~\cite{sinkhorn1967concerning,benamou2015iterative}.
Along this direction, the logarithmic stabilized Sinkhorn scaling algorithm~\cite{chizat2018scaling,schmitzer2019stabilized} and the proximal point method~\cite{xie2020fast} make efforts to suppress the numerical instability of the classic Sinkhorn scaling algorithm and solve the entropic OT problem robustly. 
The Greenkhorn algorithm~\cite{altschuler2017near} provides a stochastic Sinkhorn scaling algorithm for batch-based optimization. 
When the marginal distributions of the optimal transport plan are unreliable or unavailable, the variants of the original OT problem, e.g., the partial OT~\cite{benamou2015iterative,chapel2020partial} and the unbalanced OT~\cite{chizat2018scaling,pham2020unbalanced} are considered, and the algorithms focusing on these variants are developed accordingly.
Besides the Sinkhorn scaling algorithm, some other algorithms are developed, e.g., the Bregman ADMM~\cite{wang2014bregman,ye2017fast,xu2020gromov}, the smoothed semi-dual algorithm~\cite{blondel2018smooth}, and the conditional gradient algorithm~\cite{titouan2019optimal,titouan2020fused}. 

Recently, some attempts have been made to design neural networks to imitate the Sinkhorn-based algorithms of OT problems, such as the Gumbel-Sinkhorn network~\cite{mena2018learning}, the sparse Sinkhorn attention model~\cite{tay2020sparse}, the Sinkhorn autoencoder~\cite{patrini2020sinkhorn}, and the Sinkhorn-based transformer~\cite{sander2021sinkformers}. 
Focusing on pooling layers, some OT-based solutions have been proposed as well. 
In~\cite{mialon2020trainable}, an OT-based feature aggregation method called OTK is proposed. 
Following a similar idea, a differentiable expectation-maximization pooling method is proposed in~\cite{kim2021differentiable}, whose implementation is based on solving an entropic OT problem.
More recently, the pooling methods based on sliced Wasserstein distance~\cite{kolouri2018sliced} are proposed, e.g., the sliced-Wasserstein embedding (SWE) in~\cite{naderializadeh2021pooling} and the WEGL in~\cite{kolouri2020wasserstein}. 
However, these methods only consider solving the pooling problems in specific tasks rather than developing a generalized pooling framework.
Moreover, they focus on the optimal transport in the sample space and are highly dependent on Sinkhorn-based algorithms and random projections, which ignore the potentials of other algorithms.

\subsection{Implicit layers and optimization-driven models}
Essentially, our ROTP layers can be viewed as new members of deep implicit layers~\cite{el2021implicit,look2020differentiable,bai2019deep} (or equivalently, called declarative neural networks~\cite{gould2021deep}), which achieve global pooling by solving optimization problems. 
Many implicit layers have been proposed, e.g., the input convex neural networks (ICNNs)~\cite{amos2017input} and the deep equilibrium (DEQ) network~\cite{bai2019deep}, and so on.
From the viewpoint of ordinary differential equations (ODEs), the DEQ network reformulates the ResNet~\cite{he2016deep} as a single implicit layer whose feed-forward computation corresponds to fixed point iterations.
The OptNet in~\cite{amos2017optnet} provides a toolbox to implement convex optimization problems as neural network layers. 
At the same time, some researchers contributed to this field from the viewpoint of signal processing, implementing the iteration optimization steps of compressive sensing by neural networks~\cite{sun2016deep,wu2019learning}. 
In general, the learning of such optimization-driven neural networks is based on auto-differentiation (AD)~\cite{baydin2018automatic}, which often owns high time and space complexity. 
Fortunately, for the layers corresponding to convex optimization problems, their gradients often have closed-form solutions, and their backpropagation steps are efficient~\cite{gould2021deep}.

Note that the study of implicit layers has been interactive with computational optimal transport methods. 
For example, an optimal transport model based on the input convex neural network has been proposed in~\cite{makkuva2020optimal}. 
More recently, the work in~\cite{xie2021hypergradient,gould2022exploiting} develops the Sinkhorn-scaling algorithm as implicit layers, whose backward computation is achieved in a closed form. 
Focusing on the design and the learning of global pooling operations, our generalized pooling framework provides a new optimization-driven solution.

%% file: tex/models.tex
\section{Proposed Global Pooling Framework}\label{sec:model}
\subsection{A generalized formulation of global pooling}
Denote $\mathcal{X}_D=\{\bm{X}\in \mathbb{R}^{D\times N}|N\in\mathbb{N}\}$ as the space of sample sets, where each set $\bm{X}=[\bm{x}_1,...,\bm{x}_N]\in\mathbb{R}^{D\times N}$ contains $N$ $D$-dimensional samples. 
In practice, $\bm{X}$ can be used to represent the instances in a bag, the node embeddings of a graph, or the local visual features in an image.
Following the work in~\cite{gulcehre2014learned,ko2021learning,li2020deepergcn}, we assume the input data $\bm{X}$ to be nonnegative in this study. 
This assumption is generally reasonable because the input data are often processed by non-negative activations, like ReLU, Sigmoid, Softplus, and so on.
For some pooling methods, e.g., the max-pooling and the generalized norm pooling~\cite{ko2021learning}, the non-negativeness of input data is even necessary.

A global pooling operation, denoted as $f:\mathcal{X}_{D}\mapsto\mathbb{R}^D$, maps each set to a single vector and ensures the output is \textit{permutation-invariant}, i.e., $f(\bm{X})=f(\bm{X}_{\pi})$ for $\bm{X},\bm{X}_{\pi}\in\mathcal{X}_D$, where $\bm{X}_{\pi}=[\bm{x}_{\pi(1)},...,\bm{x}_{\pi(N)}]$ and $\pi$ is an arbitrary permutation. 
As aforementioned, many pooling methods have been proposed to achieve this aim.
Typically, the mean-pooling takes the average of the input vectors as its output, i.e., $f(\bm{X})=\frac{1}{N}\sum_{n=1}^N\bm{x}_n$. 
Another popular pooling operation, max-pooling, concatenates the maximum of each dimension as its output, i.e., $f(\bm{X})=\|_{d=1}^{D} \max_n\{x_{dn}\}_{n=1}^{N}$, where $x_{dn}$ is the $d$-th element of $\bm{x}_n$ and ``$\|$'' represents the concatenation operator. 
The attention-pooling in~\cite{ilse2018attention} outputs the weighted summation of the input vectors, i.e., $f(\bm{X})=\bm{X}\bm{a}_{\bm{X}}$, where $\bm{a}_{\bm{X}}\in\Delta^{N-1}$ is a vector on the $(N-1)$-Simplex. 
The attention-pooling leverages a self-attention mechanism to derive $\bm{a}_{\bm{X}}$ from the input $\bm{X}$, $i.e.$, $\bm{a}_{\bm{X}}=\text{softmax}(\bm{w}^T\text{tanh}(\bm{VX}))^T$. 

The element $x_{dn}$ of the $\bm{X}$ represents the $d$-th feature of the $n$-th sample. 
From the perspective of statistical signal processing, it can be treated as the signal corresponding to a pair of the sample index and the feature dimension. 
Given all sample indices and feature dimensions, we can define a joint distribution for them, denoted as $\bm{P}=[p_{dn}]\in [0, 1]^{D\times N}$, and the element $p_{dn}$ indicates the significance of the signal $x_{dn}$. 
Accordingly, the above global pooling methods yield the following generalized formulation that calculates and concatenates the conditional expectations of $x_{dn}$'s for $d=1,...,D$:
\begin{eqnarray}\label{eq:pooling}
f(\bm{X})=(\bm{X}\odot \underbrace{\text{diag}^{-1}( \overbrace{\bm{P1}_N}^{\bm{p}=[p_d]})\bm{P}}_{\tilde{\bm{P}}=[p_{n|d}]})\bm{1}_N
=\big\Vert_{d=1}^{D}\mathbb{E}_{n\sim p_{n|d}}[x_{dn}],
\end{eqnarray}
where $\odot$ is the Hadamard product, $\text{diag}(\cdot)$ converts a vector to a diagonal matrix, and $\bm{1}_N$ represents the $N$-dimensional all-one vector. 
$\bm{P1}_N=\bm{p}$ is the distribution of feature dimensions. 
$\text{diag}^{-1}(\bm{p})\bm{P}=\tilde{\bm{P}}=[p_{n|d}]$ normalizes the rows of $\bm{P}$, and the $d$-th row leads to the distribution of sample indices conditioned on the $d$-th feature dimension. 

Based on the generalized formulation in~\eqref{eq:pooling}, we can find that the above global pooling methods apply different mechanisms to derive the $\bm{P}$.
Given $\bm{X}$, the mean-pooling treats each element evenly and $\bm{P}=[\frac{1}{DN}]$. 
The max-pooling sets $\bm{P}\in \{0, \frac{1}{D}\}^{D\times N}$ and $p_{dn}=\frac{1}{D}$ if and only if $n=\arg\max_{m}\{x_{dm}\}_{m=1}^{N}$. 
The attention-pooling derives $\bm{P}$ as a learnable rank-one matrix parameterized by the input $\bm{X}$, i.e., $\bm{P}=\frac{1}{D}\bm{1}_D\bm{a}^{T}_{\bm{X}}$. 
All these operations set the marginal distribution of feature dimensions to be uniform, i.e., $\bm{p}=\bm{P1}_N=[\frac{1}{D}]$. 
For the other marginal distribution $\bm{q}=\bm{P}^T\bm{1}_D$, some pooling methods impose specific constraints, e.g., $\bm{q}=\frac{1}{N}\bm{1}_N$ for the mean-pooling and $\bm{q}=\bm{a}_{\bm{X}}$ for the attention-pooling, while the max-pooling makes $\bm{q}$ unconstrained.
In the following content, we will show that in general scenarios, we can relax the constraints to some regularization terms when the marginal distributions have some uncertainties.

\subsection{A regularized optimal transport pooling operation} 
The above analysis indicates that we can unify typical pooling operations in an interpretable algorithmic framework based on the expectation-maximization principle. 
In particular, the pooling operation in~\eqref{eq:pooling} is determined by the joint distribution $\bm{P}$. 
To keep the fused data as informative as possible, we would like to obtain a joint distribution that maximizes the expectations in~\eqref{eq:pooling}, which leads to the following optimization problem:
\begin{eqnarray}\label{eq:ot}
\begin{aligned}
\bm{P}^*&=\arg\sideset{}{_{\bm{P}\in \Pi(\bm{p},\bm{q})}}\max \sideset{}{_{d=1}^{D}}\sum p_d \mathbb{E}_{n\sim p_{n|d}}[x_{dn}]\\
&=\arg\sideset{}{_{\bm{P}\in \Pi(\bm{p},\bm{q})}}\max\underbrace{\mathbb{E}_{(d, n)\sim \bm{P}}[x_{dn}]}_{\langle\bm{X},\bm{P}\rangle}
\end{aligned}
\end{eqnarray}
where $\langle\cdot,\cdot\rangle$ represents the inner product of matrices. 
$\bm{p}=[p_d]\in\Delta^{D-1}$ and $\bm{q}\in\Delta^{N-1}$ are predefined distributions for feature dimensions and sample indices, respectively. 
Accordingly, the marginal distributions of $\bm{P}$ are restricted to be $\bm{p}$ and $\bm{q}$, i.e., $\bm{P}\in \Pi(\bm{p},\bm{q})=\{\bm{P}\geq \bm{0}|\bm{P}\bm{1}_N=\bm{p},\bm{P}^T\bm{1}_D=\bm{q}\}$. 

As shown in~\eqref{eq:ot}, the objective function is the weighted summation of the expectations conditioned on different feature dimensions, which leads to the expectation of all $x_{dn}$'s. 
Here, we have connected the global pooling operation to the theory of optimal transport --- \eqref{eq:ot} is a classic optimal transport problem~\cite{villani2008optimal}, which learns the optimal joint distribution $\bm{P}^*$ to maximize the expectation of $x_{dn}$.
Plugging $\bm{P}^*$ into~\eqref{eq:pooling} leads to a global pooling result of $\bm{X}$.

However, solving~\eqref{eq:ot} directly often leads to undesired pooling results because of the following three reasons:

\textbf{$i)$ The nature of linear programming.} 
Solving~\eqref{eq:ot} is time-consuming and always leads to a sparse $\bm{P}^*$ because it is a constrained linear programming problem. 
A sparse $\bm{P}^*$ tends to filter out some weak but possibly-informative signals in $\bm{X}$, which may have negative influences on downstream tasks.

\textbf{$ii)$ The uncertainty of marginal distributions.} 
Solving~\eqref{eq:ot} requires us to set the marginal distributions $\bm{p}$ and $\bm{q}$ in advance. 
However, such exact marginal distributions are often unavailable or unreliable.\footnote{That is why most existing pooling methods either assume the marginal distributions to be uniform or make them unconstrained.}

\textbf{$iii)$ The lack of structural information.} 
The optimal transport problem in~\eqref{eq:ot} did not consider the structural relations among samples and those among features. 
However, real-world samples, like the node embeddings of a graph and the instances in a bag, are non-i.i.d. in general, and their features can be correlated. 
The state-of-the-art pooling methods, especially those for graph representation~\cite{yuan2020structpool,wang2020second,ying2018hierarchical,ranjan2020asap}, often take such structural information into account.

According to the above analysis, to make the optimal transport-based pooling applicable, we need to $i)$ improve the smoothness of the optimization problem, $ii)$ take the uncertainties of the marginal distributions into account, and $iii)$ leverage the sample-level and feature-level structural information hidden in the input data. 
Therefore, we extend~\eqref{eq:ot} to a regularized optimal transport (ROT) problem:
\begin{eqnarray}\label{eq:uot}
\begin{aligned}
&\bm{P}_{\text{rot}}^*(\bm{X};\bm{\theta})
=\arg\sideset{}{_{\bm{P}\in\Omega}}\min\overbrace{\underbrace{\langle-\bm{X},\bm{P}\rangle}_{\text{OT term}} + \underbrace{\alpha_0\langle C(\bm{X},\bm{P}),\bm{P}\rangle}_{\text{Structural Reg.}}}^{\text{Fused Gromov-Wasserstein discrepancy}} \\
&\quad+\underbrace{\alpha_1 \text{R}(\bm{P})}_{\text{Smoothness Reg.}} +\underbrace{\alpha_2\text{KL}(\bm{P1}_N | \bm{p}_0)+\alpha_3 \text{KL}(\bm{P}^T\bm{1}_D | \bm{q}_0)}_{\text{Marginal Reg.}},
\end{aligned}
\end{eqnarray}
where $\Omega=\{\bm{P}>\bm{0}|\bm{1}_D^T\bm{P}\bm{1}_N=1\}$. 
In~\eqref{eq:uot}, we introduce the following three regularizers, each of which solves one of the above three challenges.
    
\textbf{$i)$ Smoothness regularization.} 
$\text{R}(\bm{P})$ is a regularizer of $\bm{P}$, which is used to improve the smoothness of the objective function. 
Typically, we can set $\text{R}(\bm{P})$ as the negative entropy of $\bm{P}$ ($\text{R}(\bm{P})=\langle\bm{P},\log\bm{P}-\bm{1}\rangle=\sum_{d,n}p_{dn}(\log p_{dn} -1)$)~\cite{cuturi2013sinkhorn} or the quadratic regularizer of $\bm{P}$ ($\text{R}(\bm{P})=\|\bm{P}\|_F^2$)~\cite{blondel2018smooth}. 
The parameter $\alpha_1$ controls the significance of $\text{R}(\bm{P})$. 

\textbf{$ii)$ Marginal prior regularization.} 
Instead of imposing strict constraints, we leverage two KL divergence terms in~\eqref{eq:uot} to penalize the difference between the marginals of $\bm{P}$ and the predefined prior distributions (denoted as $\bm{p}_0$ and $\bm{q}_0$, respectively).
Here, $\text{KL}(\bm{a}|\bm{b})=\langle\bm{a},\log\bm{a}-\log\bm{b}\rangle-\langle\bm{a}-\bm{b},\bm{1}\rangle$ represents the KL-divergence between $\bm{a}$ and $\bm{b}$. 
The strength of these two terms is controlled by the weights $\alpha_2$ and $\alpha_3$, respectively. 
This regularization helps us achieve a trade-off between the utilization of prior information and the robustness to its uncertainty.

\textbf{$iii)$ Gromov-Wasserstein discrepancy-based structural regularization.} 
Given $\bm{X}$, we construct its feature-level and sample-level covariance matrices,\footnote{In practice, besides the covariance matrices, we can also leverage other methods to define the feature-level and sample-level similarities, e.g., the cosine similarity and other kernel matrices.} respectively, i.e., $\bm{\Sigma}_1=\frac{1}{N}(\bm{X}-\bm{\mu}_1\bm{1}_N^T)(\bm{X}-\bm{\mu}_1\bm{1}_N^T)^T$ and $\bm{\Sigma}_2=\frac{1}{D}(\bm{X}-\bm{1}_D\bm{\mu}_2^T)^T(\bm{X}-\bm{1}_D\bm{\mu}_2^T)$, where $\bm{\mu}_1=\frac{1}{N}\bm{X1}_N$ and $\bm{\mu}_2=\frac{1}{D}\bm{X}^T\bm{1}_D$. 
Following the work in~\cite{wang2020second,yuan2020structpool}, we would like to make the feature-level covariance highly correlated with the sample-level covariance such that the features can preserve the structural relations among the samples.
Therefore, we construct a structural cost as follows: 
\begin{eqnarray}\label{eq:cov}
    \begin{aligned}
        C(\bm{X},\bm{P}) = -\bm{\Sigma}_1\bm{P}\bm{\Sigma}_2^T,
    \end{aligned}
\end{eqnarray}
and $\langle C(\bm{X},\bm{P}),\bm{P}\rangle=-\text{tr}(\bm{\Sigma}_1\bm{P}\bm{\Sigma}_2^T\bm{P}^T)$, whose significance is controlled by $\alpha_0$. 
It is easy to find that this structural regularization is the same as the objective of the Gromov-Wasserstein discrepancy problem in~\cite{memoli2011gromov,peyre2016gromov}, penalizing the difference between the sample-level covariance and the feature-level covariance. 
As shown in~\eqref{eq:uot}, combining the original optimal transport term with the structural regularizer leads to the well-known fused Gromov-Wasserstein (FGW) discrepancy~\cite{titouan2019optimal}, which is an optimal transport-based metric for structured data (like graphs and sets)~\cite{titouan2020fused,xu2020learning}.

Compared with~\eqref{eq:ot}, \eqref{eq:uot} is an unconstrained optimization problem, and thus we can apply differentiable algorithms to optimize it.\footnote{When $\alpha_0=0$ and $\text{R}(\cdot)$ is a strictly-convex function, the objective function in~\eqref{eq:uot} is strictly-convex.}
The optimal transport matrix $\bm{P}_{\text{rot}}^*$ can be viewed as a function of $\bm{X}$, whose parameters are the weights of the regularizers and the prior distributions, i.e., $\bm{P}_{\text{rot}}^*(\bm{X};\bm{\theta})$, where $\bm{\theta}=\{\alpha_0,\alpha_1,\alpha_2,\alpha_3,\bm{p}_0,\bm{q}_0\}$ represents the model parameters for convenience. 
Plugging it into~\eqref{eq:pooling}, we obtain the proposed regularized optimal transport pooling (ROTP) operation:
\begin{eqnarray}\label{eq:uotp}
\begin{aligned}
f_{\text{rot}}(\bm{X};\bm{\theta})
= (\bm{X}\odot \text{diag}^{-1}( \bm{P}_{\text{rot}}^*(\bm{X};\bm{\theta})\bm{1}_N)\bm{P}_{\text{rot}}^*(\bm{X};\bm{\theta}))\bm{1}_N.
\end{aligned}
\end{eqnarray}

\subsection{The rationality and generalizability of ROTP}
Our ROTP is a feasible global pooling operation. 
In particular, it satisfies the requirement of permutation-invariance under mild conditions. 
\begin{theorem}\label{thm:pi}
The ROTP in~\eqref{eq:uotp} is permutation-invariant, i.e., $f_{\text{rot}}(\bm{X})=f_{\text{rot}}(\bm{X}_\pi)$ for an arbitrary permutation $\pi$ when the following two conditions are satisfied.
\begin{itemize}
    \item[$i)$] The $\text{R}(\cdot)$ in~\eqref{eq:uot} is a permutation-invariant function of $\bm{P}$.
    \item[$ii)$] The $\bm{q}_0$ in~\eqref{eq:uot} is either a permutation-equivariant function of $\bm{X}$ or a uniform distribution, i.e., $\bm{q}_0=\frac{1}{N}\bm{1}_N$.
\end{itemize}
\end{theorem} 
\begin{proof}
For convenience, we ignore the notation $\bm{\theta}$ in the following derivation.
Let $\bm{P}^*$ be the optimal solution of~\eqref{eq:uot} given $\bm{X}$.
Denote $\pi: \{1,...,N\}\mapsto\{1,....,N\}$ as an arbitrary permutation and $\bm{X}_{\pi}$ as the column-wise permuted data. 
We have the following six equations:
\begin{eqnarray}\label{eq:cond}
\begin{aligned}
   i)~~&\langle-\bm{X},\bm{P}^*\rangle= \langle-\bm{X}_{\pi},\bm{P}_{\pi}^*\rangle,\\
   ii)~~&\langle C(\bm{X},\bm{P}^*),\bm{P}^*\rangle = -\text{tr}(\bm{\Sigma}_1\bm{P}^*\bm{\Sigma}_2^T(\bm{P}^*)^T)\\
   &=-\text{tr}(\bm{\Sigma}_1\bm{P}_{\pi}^*\bm{\Sigma}_{2,\pi,\pi}^T(\bm{P}_{\pi}^*)^T)=\langle C(\bm{X}_{\pi},\bm{P}_{\pi}^*),\bm{P}_{\pi}^*\rangle\\
   iii)~~&\text{R}(\bm{P}^*)=\text{R}(\bm{P}_{\pi}^*),\\
   iv)~~&\text{KL}(\bm{P}^*\bm{1}_N|\bm{p}_0)=\text{KL}(\bm{P}_{\pi}^*\bm{1}_N|\bm{p}_0),\\
   v)~~&\text{When $\bm{q}_0$ is a permutation-equivariant function of $\bm{X}$:}\\
   &\text{KL}((\bm{P}^*)^T\bm{1}_D|\bm{q}_0(\bm{X}))=\underbrace{\text{KL}((\bm{P}_{\pi}^*)^T\bm{1}_D|\bm{q}_{0}(\bm{X}_{\pi}))}_{\text{KL}((\bm{P}_{\pi}^*)^T\bm{1}_D|\bm{q}_{0,\pi}(\bm{X}))}\\
   vi)~~&\text{When $\bm{q}_0$ is uniform:}\\
   &\text{KL}((\bm{P}^*)^T\bm{1}_D|\bm{q}_0)=\text{KL}((\bm{P}_{\pi}^*)^T\bm{1}_D|\bm{q}_{0}).
\end{aligned}
\end{eqnarray}
where $\bm{P}^*_{\pi}$ is the column-wise permutation result of $\bm{P}^*$.
In the second equation, $\bm{\Sigma}_{2,\pi,\pi}$ means permuting $\bm{\Sigma}_2$ row-wisely and column-wisely based on $\pi$. 
The third equation is based on the condition 1, and the fifth and the sixth equations are based on the condition 2.
According to~\eqref{eq:cond}, $\bm{P}^*_{\pi}$ must be the optimal solution of~\eqref{eq:uot} given $\bm{X}_{\pi}$.
Therefore, $\bm{P}^*$ is a permutation-equivariant function of $\bm{X}$, i.e., $\bm{P}^*_{\pi}(\bm{X})=\bm{P}^*(\bm{X}_{\pi})$.

Plugging $\bm{P}^*_{\pi}(\bm{X})=\bm{P}^*(\bm{X}_{\pi})$ into~\eqref{eq:uotp}, we have
\begin{eqnarray}
\begin{aligned}
f_{\text{rot}}(\bm{X}_{\pi})
=&(\bm{X}_{\pi}\odot (\text{diag}^{-1}( \bm{P}^*(\bm{X}_{\pi})\bm{1}_N)\bm{P}^*(\bm{X}_{\pi})))\bm{1}_N\\
=&(\bm{X}_{\pi}\odot (\text{diag}^{-1}( \bm{P}_{\pi}^*(\bm{X})\bm{1}_N)\bm{P}_{\pi}^*(\bm{X})))\bm{1}_N\\
=&(\bm{X}_{\pi}\odot (\text{diag}^{-1}( \bm{P}^*(\bm{X})\bm{1}_N)\bm{P}_{\pi}^*(\bm{X})))\bm{1}_N\\
=&(\bm{X}\odot (\text{diag}^{-1}( \bm{P}^*(\bm{X})\bm{1}_N)\bm{P}^*(\bm{X})))\bm{1}_N\\
=&f_{\text{rot}}(\bm{X}),
\end{aligned}
\end{eqnarray}
which completes the proof.
\end{proof}
Theorem~\ref{thm:pi} provides us with sufficient conditions to ensure the permutation-invariance of the proposed ROTP framework. 
Note that the two conditions in Theorem~\ref{thm:pi} are common in practice. 
When $\text{R}(\bm{P})$ is an entropic or a quadratic function of $\bm{P}$, Condition 1 is always held.
The $\bm{q}_0$ is a permutation-equivariant function of $\bm{X}$ for the max-pooling and the attention-pooling, and it is uniform for the mean-pooling.

Moreover, our ROTP operation provides a generalized global pooling framework that unifies many representative pooling operations.
In particular, the mean-pooling, the max-pooling, and the attention-pooling can be formulated as the specializations of~\eqref{eq:uotp} under different parameter configurations.
\begin{proposition}\label{prop:equiv}
Given an arbitrary $\bm{X}\in\mathbb{R}^{D\times N}$, the mean-pooling, the max-pooling, and the attention-pooling with attention weights $\bm{a}_{\bm{X}}$ can be equivalently achieved by the $f_{\text{rot}}(\bm{X};\bm{\theta})$ in~\eqref{eq:uotp} under the following configurations:

\begin{center}
\begin{tabular}{@{\hspace{3pt}}c@{\hspace{5pt}}|
@{\hspace{5pt}}c@{\hspace{6pt}}c@{\hspace{6pt}}c@{\hspace{3pt}}}
\hline
$f_{\text{rot}}(\bm{X};\bm{\theta})$
&Mean-pooling
&Max-pooling
&Attention-pooling\\
\hline
$\alpha_0$
&0
&0
&0\\
$\alpha_1$
&$\rightarrow\infty$
&0
&$\rightarrow\infty$\\
$\alpha_2$
&$\rightarrow\infty$
&$\rightarrow\infty$
&$\rightarrow\infty$\\
$\alpha_3$
&$\rightarrow\infty$
&0
&$\rightarrow\infty$\\
$\bm{p}_0$
&$\frac{1}{D}\bm{1}_D$
&$\frac{1}{D}\bm{1}_D$
&$\frac{1}{D}\bm{1}_D$\\
$\bm{q}_0$
&$\frac{1}{N}\bm{1}_N$
&---
&$\bm{a}_{\bm{X}}$\\
\hline
\end{tabular}
\end{center}
Here, ``$\bm{q}_0=-$'' means that $\bm{q}_0$ can be arbitrary vectors.  
$\alpha_2,\alpha_3\rightarrow \infty$ means that the marginal prior regularizers become strict marginal constraints.
$\alpha_1\rightarrow\infty$ means that the smoothness regularizer is dominant and thus the OT term becomes ignorable.
\end{proposition}
\begin{proof}
\textbf{Equivalence to mean-pooling operation:}
For~\eqref{eq:uot}, $\alpha_2,\alpha_3\rightarrow\infty$, $\bm{p}_0=\frac{1}{D}\bm{1}_D$ and $\bm{q}_0=\frac{1}{N}\bm{1}_N$, we require the marginals of $\bm{P}^*$ to match with uniform distributions strictly. 
Additionally, $\alpha_0=0$ and $\alpha_1\rightarrow\infty$ mean that the smoothness regularizer is dominant and both the OT term and the GW-based structural regularizer are ignored.
Therefore, the optimization problem in~\eqref{eq:uot} degrades to
\begin{eqnarray}\label{eq:mean}
\bm{P}^*=\arg\sideset{}{_{\bm{P}\in \Pi(\frac{1}{D}\bm{1}_D, \frac{1}{N}\bm{1}_N)}}\min \text{R}(\bm{P}).
\end{eqnarray}
when $\text{R}(\bm{P})=\langle\bm{P},\log\bm{P}-\bm{1}\rangle$ or $\|\bm{P}\|_F^2$, the optimal solution of~\eqref{eq:mean} is $\bm{P}^*=[\frac{1}{DN}]$, and thus the corresponding $f_{\text{rot}}$ becomes the mean-pooling operation.

\textbf{Equivalence to max-pooling operation:}
For~\eqref{eq:uot}, when $\alpha_0=\alpha_1=0$, both the structural and the smoothness regularizers are ignored. 
$\alpha_2\rightarrow\infty$ and $\bm{p}_0=\frac{1}{D}\bm{1}_D$ mean that $\bm{P1}_N=\frac{1}{D}\bm{1}_D$ strictly, while $\alpha_3=0$ and $\bm{q}_0=-$ mean that $\bm{P}^T\bm{1}_D$ is unconstrained.
In this case, the problem in~\eqref{eq:uot} becomes
\begin{eqnarray}
\bm{P}^*=\arg\sideset{}{_{\bm{P}\in \Pi(\frac{1}{D}\bm{1}_D, \cdot)}}\max \langle\bm{X},\bm{P}\rangle,
\end{eqnarray}
whose optimal solution obviously corresponds to setting $p^*_{dn}=\frac{1}{D}$ if and only if $n=\arg\max_{m}\{x_{dm}\}_{m=1}^M$.
Therefore, the corresponding $f_{\text{rot}}$ becomes the max-pooling operation.

\textbf{Equivalence to attention-pooling operation:}
Similar to the case of mean-pooling, given the configuration shown in the above table, the problem in~\eqref{eq:uot} becomes
\begin{eqnarray}
\bm{P}^*=\arg\sideset{}{_{\bm{P}\in \Pi(\frac{1}{D}\bm{1}_D, \bm{a}_{\bm{X}})}}\min \text{R}(\bm{P}),
\end{eqnarray}
whose optimal solution is $\bm{P}^*=\frac{1}{D}\bm{1}_D\bm{a}_{\bm{X}}^T$. 
Accordingly, the corresponding $f_{\text{rot}}$ becomes the attention-pooling operation.
\end{proof}

\subsection{Hierarchical ROTP operations}\label{sec:extend}
Proposition~\ref{prop:equiv} demonstrates that a single ROTP operation can imitate various global pooling methods.
Moreover, the hierarchical combination of multiple ROTP operations can reproduce more complicated pooling mechanisms, including mixed pooling operation and set fusion. 

\subsubsection{Hierarchical ROTP for mixed pooling}
Typically, a mixed pooling operation first applies multiple different pooling operations to the same input data and then aggregates the outputs of the pooling operations. 
It often has more substantial representation power than a single pooling operation because of considering different pooling mechanisms.
For example, the mixed mean-max pooling operation in~\cite{lee2016generalizing} is defined as follows: 
\begin{eqnarray}\label{eq:mmp}
f_{\text{mix}}(\bm{X})=\omega \text{MeanPool}(\bm{X}) + (1-\omega)\text{MaxPool}(\bm{X}).
\end{eqnarray}
When $\omega\in (0,1)$ is a single learnable scalar,~\eqref{eq:mmp} is called ``Mixed mean-max pooling". 
When $\omega$ is parameterized as a sigmoid function of $\bm{X}$,~\eqref{eq:mmp} is called ``Gated mean-max pooling". 

It is easy to demonstrate that such a mixed pooling operation can be equivalently achieved by hierarchically integrating three ROTP operations.
\begin{proposition}[Hierarchical ROTP for mixed pooling]\label{prop:mix}
Given an arbitrary $\bm{X}\in\mathbb{R}^{D\times N}$, the $f_{\text{mix}}(\bm{X})$ in~\eqref{eq:mmp} can be equivalently implemented by $f_{\text{rot}}([f_{\text{rot}}(\bm{X};\bm{\theta}_1),f_{\text{rot}}(\bm{X};\bm{\theta}_2)];\bm{\theta}_3)$, 
where $\bm{\theta}_1=\{0,\infty,\infty,\infty,\frac{1}{D}\bm{1}_D,\frac{1}{N}\bm{1}_N\}$, $\bm{\theta}_2=\{0,0, \infty,0,\frac{1}{D}\bm{1}_D,-\}$, and $\bm{\theta}_3=\{0,\infty,\infty,\infty,\frac{1}{D}\bm{1}_D,[\omega,1-\omega]^T\}$.
\end{proposition}
\begin{proof}
In particular, given $\bm{X}\in\mathbb{R}^{D\times N}$, we have
\begin{eqnarray}\label{eq:hot}
\begin{aligned}
&f_{\text{mix}}(\bm{X})
=\omega f_{\text{rot}}(\bm{X};\bm{\theta}_1) + (1-\omega)f_{\text{rot}}(\bm{X};\bm{\theta}_2)\\
&=\underbrace{[f_{\text{rot}}(\bm{X};\bm{\theta}_1),f_{\text{rot}}(\bm{X};\bm{\theta}_2)]}_{\bm{Y}\in\mathbb{R}^{D\times 2}}[\omega, 1-\omega]^T\\
&=\Bigl(\bm{Y}\odot \text{diag}^{-1}\bigl( \underbrace{\overbrace{\tfrac{1}{D}\bm{1}_D}^{\bm{p}_0}\overbrace{[\omega,1-\omega]}^{\bm{q}_0^T}}_{\bm{P}^*}\bm{1}_2\bigr)\bigl(\tfrac{1}{D}\bm{1}_D[\omega,1-\omega]\bigr)\Bigr)\bm{1}_2\\
&=f_{\text{rot}}(\bm{Y};\bm{\theta}_3).
\end{aligned}
\end{eqnarray}
Here, the first equation is based on Proposition~\ref{prop:equiv} --- we can replace $\text{MeanPool}(\bm{X})$ and $\text{MaxPool}(\bm{X})$ with $f_{\text{rot}}(\bm{X};\bm{\theta}_1)$ and $f_{\text{rot}}(\bm{X};\bm{\theta}_2)$, respectively, where $\bm{\theta}_1=\{0,\infty,\infty,\infty,\frac{1}{D}\bm{1}_D,\frac{1}{N}\bm{1}_N\}$ and $\bm{\theta}_2=\{0,0, \infty,0,\frac{1}{D}\bm{1}_D,-\}$. 
The concatenation of $f_{\text{rot}}(\bm{X};\bm{\theta}_1)$ and $f_{\text{rot}}(\bm{X};\bm{\theta}_2)$ is a matrix with size $D\times 2$, denoted as $\bm{Y}=[f_{\text{rot}}(\bm{X};\bm{\theta}_1),f_{\text{rot}}(\bm{X};\bm{\theta}_2)]$. 
As shown in the third equation of~\eqref{eq:hot}, the $f_{\text{mix}}(\bm{X})$ in~\eqref{eq:mmp} can be rewritten based on $\bm{p}_0=\frac{1}{D}\bm{1}_D$, $\bm{q}_0=[\omega,1-\omega]^T$, and the rank-1 matrix $\bm{P}^*=\bm{p}_0\bm{q}_0^T$.
The formulation corresponds to passing $\bm{Y}$ through the third ROTP operation, i.e., $f_{\text{rot}}(\bm{Y};\bm{\theta}_3)$, where $\bm{\theta}_3=\{0, \infty,\infty,\infty,\frac{1}{D}\bm{1}_D,[\omega,1-\omega]^T\}$. 
\end{proof}
Proposition~\ref{prop:mix} means that the mixed pooling in~\eqref{eq:mmp} corresponds to a simple hierarchical ROTP (HROTP) operation, in which the outputs of two ROTP operations are fused through the third ROTP operation. 
In more general scenarios, given $\bm{X}\in\mathbb{R}^{D\times N}$, we can achieve an $M$-head mixed pooling operation via integrating $M+1$ ROTP operations as follows: 
\begin{eqnarray}\label{eq:mrotp}
\begin{aligned}
f_{\text{mrot}}(\bm{X};\bm{\Theta})=f_{\text{rot}}(\|_{m=1}^{M}f_{\text{rot}}(\bm{X};\bm{\theta}_m);\bm{\theta}_{M+1}),
\end{aligned}
\end{eqnarray}
where $\bm{\Theta}=\{\bm{\theta}_m\}_{m=1}^{M+1}$, and the $\bm{\theta}_m$'s are different from each other in general.

\subsubsection{Hierarchical ROTP for set fusion}
Many real-world data have hierarchical set-level structures. 
A typical example is combinatorial drug analysis, in which each sample is a set of drugs, and each drug can be modeled as a graph and represented as a set of node embeddings. 
Therefore, when predicting the property of a drug set, we need to first obtain a graph embedding by pooling the node embeddings of each drug and then represent the drug set by pooling the graph embeddings of different drugs. 

Such a set fusion operation can also be achieved by our HROTP operation, as illustrated in Fig.~\ref{fig:scheme_b}. 
Denote a set of $M$ sets as $\mathcal{X}=\{\bm{X}\}_{m=1}^{M}$. 
The proposed HROTP module for fusing the $M$ sets is defined as follows:
\begin{eqnarray}\label{eq:hrotp}
\begin{aligned}
f_{\text{hrot}}(\mathcal{X};\bm{\Theta},g)=f_{\text{rot}}(\|_{m=1}^{M}g(f_{\text{rot}}(\bm{X}_m;\bm{\theta}_1));\bm{\theta}_{2}),
\end{aligned}
\end{eqnarray}
where $\bm{\Theta}=\{\bm{\theta}_1,\bm{\theta}_2\}$. 
As shown in~\eqref{eq:hrotp}, $f_{\text{rot}}(\cdot;\bm{\theta}_1)$ works for pooling the elements within each set, which is reused for all the sets, and $f_{\text{rot}}(\cdot;\bm{\theta}_2)$ works for pooling the representations of different sets. 
Additionally, to enhance the representation power of the HROTP module, we can plug more neural network layers between the above two pooling steps. 
In~\eqref{eq:hrotp}, the neural network $g:\mathbb{R}^{D}\mapsto\mathbb{R}^{D'}$ works for feature extraction, which can be a multi-layer perceptron (MLP) or more complicated modules.

%% file: tex/algorithms.tex
\section{ROTP Layers and Their Implementations}\label{sec:alg}
Besides imitating existing global pooling operations based on manually-selected parameters, we further implement the proposed ROTP operation $f_{\text{rot}}(\bm{X};\bm{\theta})$ as a learnable neural network layer, whose parameters $\bm{\theta}$ include the prior distributions $\{\bm{p}_0\in\Delta^{D-1}, \bm{q}_0\in\Delta^{N-1}\}$, and the regularization weights $\bm{\alpha}=[\alpha_{i}]_{i=0}^3\in (0,\infty)^{4}$. 
These parameters are constrained parameters: $\{\bm{\alpha}_i\geq 0\}_{i=0}^3$, $\bm{p}_0\in\Delta^{D-1}$, and $\bm{q}_0\in\Delta^{N-1}$. 
We apply the following parametrization strategy to make the proposed ROTP layer with unconstrained parameters. 
In particular, we set $\{\bm{\alpha}_i=\text{softplus}(\bm{\beta}_i)\}_{i=0}^3$, where $\{\bm{\beta}_i\}_{i=0}^{3}$ are unconstrained parameters. 
For the prior distributions, we can either fix them as uniform distributions, i.e., $\bm{p}_0=\frac{1}{D}\bm{1}_D$ and $\bm{q}_0=\frac{1}{N}\bm{1}_N$, or implement them as learnable attention modules, i.e., $\bm{p}_0=\text{softmax}(\bm{U}\bm{X}\bm{1}_N)$ and $\bm{q}_0=\text{softmax}(\bm{w}^T\text{tanh}(\bm{VX}))$~\cite{ilse2018attention}, where $\bm{U},\bm{V}\in\mathbb{R}^{D\times D}$ and $\bm{w}\in\mathbb{R}^D$ are unconstrained. 
As a result, our ROTP layers can be learned by stochastic gradient descent.

The feed-forward step of the ROTP layer corresponds to solving the ROT problem in~\eqref{eq:uot} and obtaining $f_{\text{rot}}(\bm{X};\bm{\theta})$ via~\eqref{eq:uotp}. 
Its backward step corresponds to the updating of the model parameters, which adjusts the objective function in~\eqref{eq:uot} and changes the optimum $\bm{P}_{\text{rot}}^*$ accordingly.
In this work, we implement the backward step by auto-differentiation.
By learning the model parameters based on observed data, our ROTP layer may fit the data better and outperform the global pooling methods that are designed empirically. 

The architecture of our ROTP layer is determined by the optimization algorithm of~\eqref{eq:uot}. 
In the following two subsections, we introduce two algorithms to solve~\eqref{eq:uot}, which lead to two ROTP layers with different architectures.

\subsection{Sinkhorn-based ROTP layer}
We first consider leveraging the proximal point method~\cite{peyre2016gromov,xu2019scalable,xie2020fast} to solve~\eqref{eq:uot} iteratively. 
In particular, in the $t$-th iteration, given the current OT plan $\bm{P}^{(t)}$, we solve the following sub-problem:
\begin{eqnarray}\label{eq:sub-prob}
\begin{aligned}
    \bm{P}^{(t+1)}= &\arg\sideset{}{_{\bm{P}\in\Omega}}\min\langle-\bm{X},\bm{P}\rangle + \alpha_0\langle C(\bm{X},\bm{P}^{(t)}),\bm{P}\rangle
    \\ 
    &+\alpha_1 \text{R}(\bm{P})+\alpha_2\text{KL}(\bm{P1}_N | \bm{p}_0)
    \\
    &+\alpha_3 \text{KL}(\bm{P}^T\bm{1}_D | \bm{q}_0)+\underbrace{\tau\text{KL}(\bm{P}|\bm{P}^{(t)})}_{\text{Proximal term}}.
\end{aligned}
\end{eqnarray}
Here, $\text{KL}(\bm{P}|\bm{P}^{(t)})$ is the proximal term based on the current variable, which is implemented as a KL-divergence.
The weight $\tau$ controls its significance.
As a result, our ROTP layer is built by stacking $T$ feed-forward modules, and each module corresponds to the optimization of~\eqref{eq:sub-prob}.

When the smoothness regularizer is entropic, i.e., $\text{R}(\bm{P})=\langle \bm{P},\log\bm{P}-\bm{1}\rangle$, \eqref{eq:sub-prob} becomes the following entropic unbalanced optimal transport (EUOT) problem:
\begin{eqnarray}\label{eq:euot_prob}
\begin{aligned}
    \sideset{}{_{\bm{P}\in\Omega}}\min &\langle\bm{C}^{(t)},\bm{P}\rangle + (\alpha_1+\tau)\langle\log\bm{P},\bm{P}\rangle \\
    &+\alpha_2\text{KL}(\bm{P1}_N | \bm{p}_0)+\alpha_3 \text{KL}(\bm{P}^T\bm{1}_D | \bm{q}_0).
\end{aligned}
\end{eqnarray}
where the matrix $\bm{C}^{(t)}=[c^{(t)}_{dn}]=-\bm{X}-\alpha_0\bm{\Sigma}_1\bm{P}^{(t)}\bm{\Sigma}_2^T-\tau\log\bm{P}^{(t)}$ is determined by the input data and the current variable $\bm{P}^{(t)}$. 
According to~\cite{chizat2018scaling,pham2020unbalanced}, we consider the Fenchel's dual form of the EUOT problem:
\begin{eqnarray}\label{eq:dual3}
\begin{aligned}
    &\sideset{}{_{\bm{a}\in\mathbb{R}^D,\bm{b}\in\mathbb{R}^N}}\min(\alpha_1+\tau)\sideset{}{_{d,n=1}^{D,N}}\sum \exp\left( \frac{a_d+b_n + c^{(t)}_{dn}}{\alpha_1+\tau} \right)\\
    &+\alpha_2\Bigl\langle\exp\Bigl(-\frac{1}{\alpha_2}\bm{a}\Bigr), \bm{p}_0\Bigr\rangle +
    \alpha_3\Bigl\langle\exp\Bigl(-\frac{1}{\alpha_3}\bm{b}\Bigr), \bm{q}_0\Bigr\rangle.
\end{aligned}
\end{eqnarray}

\begin{figure}[t]
\centering
\includegraphics[width=0.85\linewidth]{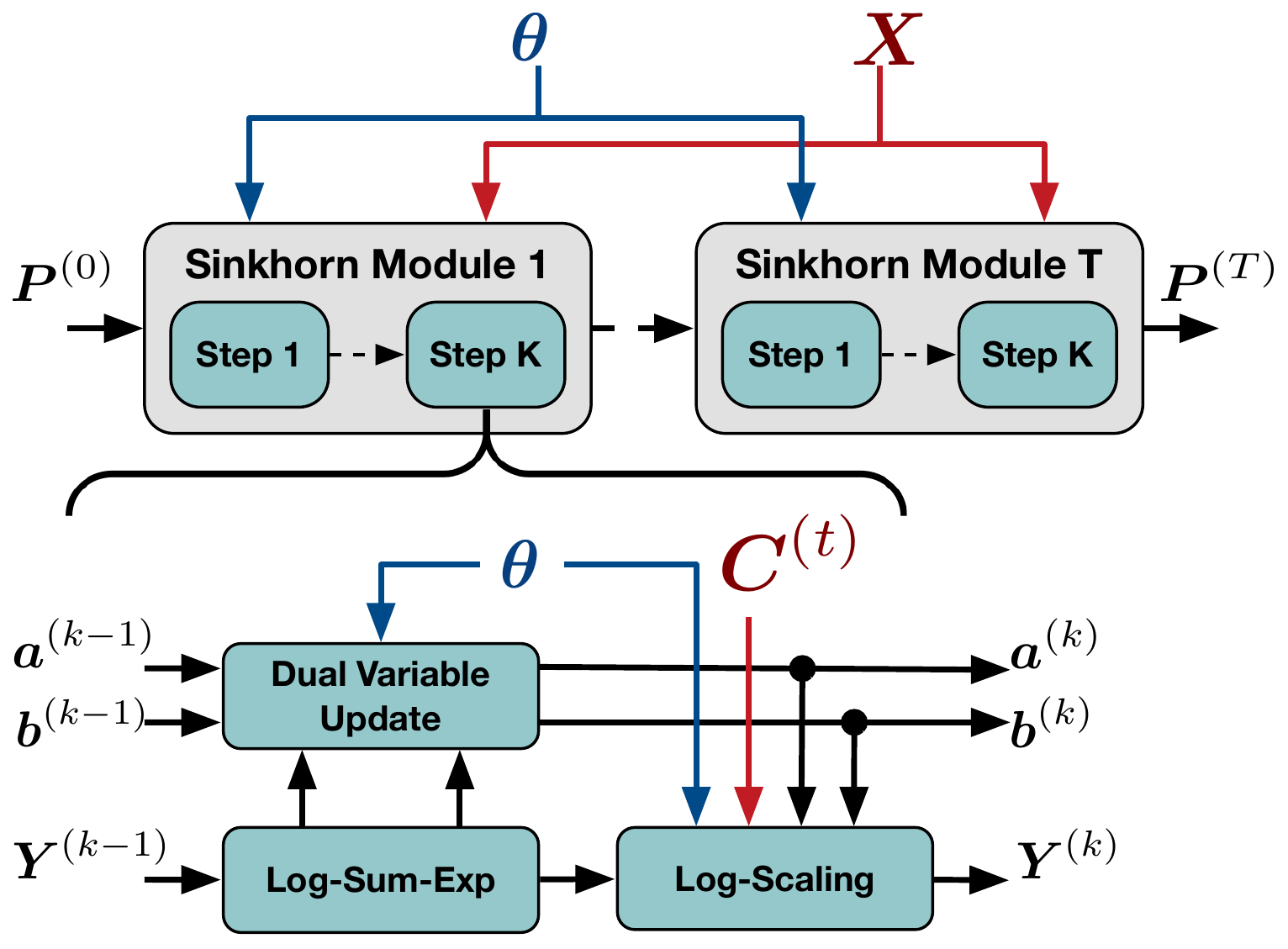}
\caption{An illustration of the Sinkhorn-based ROTP layer. 
The red, blue, and black arrows correspond to the computational flow of data, model parameters, intermediate results.}\label{fig:ff1}
\end{figure}

\begin{algorithm}[t]
    \caption{$f_{\text{rot}}(\bm{X};\bm{\theta})$ based on Sinkhorn-scaling}\label{alg:sinkhorn_uotp}
    \label{alg:sinkhorn}
	\begin{algorithmic}[1]
	    \REQUIRE Data $\bm{X}$, model parameters $\bm{\theta}$.
	    \STATE $\bm{P}^{(0)}=\bm{p}_0\bm{q}_0^T$.
	    \FOR{$t=0,...,T-1$}
        \STATE $\bm{C}^{(t)}=-\bm{X}-\alpha_0\bm{\Sigma}_1\bm{P}^{(t)}\bm{\Sigma}_2^T-\tau \log \bm{P}^{(t)}$.
        \STATE Set $\bm{a}^{(0)}=\bm{0}_D$, $\bm{b}^{(0)}=\bm{0}_N$, $\bm{Y}^{(0)}=-\frac{1}{\alpha_{1}'}\bm{C}^{(t)}$.
        \FOR{$k=0,...,K-1$}
            \STATE $\log\bm{p} =\text{LogSumExp}_{\text{row}}(\bm{Y}^{(k)})$, 
	        \STATE $\log\bm{q} =\text{LogSumExp}_{\text{col}}(\bm{Y}^{(k)})$.
	        \STATE \textbf{Dual Variable Update:}\\
	       \quad$\bm{a}^{(k+1)}=\frac{\alpha_{2}}{\alpha_{1}'+\alpha_{2}}\left(\frac{1}{\alpha_{1}'}\bm{a}^{(k)}+\log\bm{p}_0-\log\bm{p}\right)$.\\
	        \quad$\bm{b}^{(k+1)}=\frac{\alpha_{3}}{\alpha_{1}'+\alpha_{3}}\left(\frac{1}{\alpha_{1}'}\bm{b}^{(k)}+\log\bm{q}_0-\log\bm{q}\right)$.
	        \STATE \textbf{Logarithmic Scaling:}\\
	        \quad$\bm{Y}^{(k+1)}=-\frac{1}{\alpha_{1}'}\bm{C}^{(t)} + \bm{a}^{(k+1)}\bm{1}_N^T + \bm{1}_D\bm{b}^{(k+1)T}$.
        \ENDFOR
        \STATE $\bm{P}^{(t+1)}:=\exp(\bm{Y}^{(K)})$
	    \ENDFOR
	    \STATE $\bm{P}^*:=\bm{P}^{(T)}$, and  apply~\eqref{eq:uotp} to obtain $f_{\text{rot}}(\bm{X};\bm{\theta})$.
	\end{algorithmic}
\end{algorithm}

This problem can be solved by the Sinkhorn-scaling algorithm~\cite{sinkhorn1967concerning,cuturi2013sinkhorn}. 
In particular, the Sinkhorn-scaling algorithm solves this dual problem by the following iterative steps: 
$i)$ Initialize dual variables as $\bm{a}^{(0)}=\bm{0}_D$ and $\bm{b}^{(0)}=\bm{0}_N$. 
$ii)$ In the $k$-th iteration, the current dual variables $\bm{a}^{(k)}$ and $\bm{b}^{(k)}$ are updated by
\begin{eqnarray}\label{eq:sinkhorn}
\begin{aligned}
&\bm{T}(\bm{a}^{(k)},\bm{b}^{(k)})=\exp\Bigl(\bm{a}^{(k)}\bm{1}_N^T + \bm{1}_D(\bm{b}^{(k)})^T -\frac{\bm{C}^{(t)}}{\alpha_1'}\Bigr),\\
&\bm{p}^{(k)} = \bm{T}(\bm{a}^{(k)},\bm{b}^{(k)})\bm{1}_N,\quad\bm{q}^{(k)} = \bm{T}(\bm{a}^{(k)},\bm{b}^{(k)})^T\bm{1}_D,\\
&\bm{a}^{(k+1)}=\frac{\alpha_2}{\alpha_1'+\alpha_2}\left(\frac{1}{\alpha_1'}\bm{a}^{(k)}+\log\bm{p}_0-\log\bm{p}^{(k)}\right)\\
&\bm{b}^{(k+1)}=\frac{\alpha_3}{\alpha_1'+\alpha_3}\left(\frac{1}{\alpha_1'}\bm{b}^{(k)}+\log\bm{q}_0-\log\bm{q}^{(k)}\right),
\end{aligned}
\end{eqnarray}
where $\alpha_1'=\alpha_1+\tau$.
$iii)$ After $K$ steps, the variables converges and the optimal transport plan is updated as
\begin{eqnarray}
\bm{P}^{(t+1)}=\bm{T}(\bm{a}^{(K)},\bm{b}^{(K)}).
\end{eqnarray}
The convergence of the algorithm has been proven in~\cite{xu2019gromov} --- with the increase of $t$, $\bm{P}^{(t)}$ converges to a stationary point. 
Therefore, after repeating the above process $T$ times, we set $\bm{P}^*:=\bm{P}^{(T)}$.

The algorithm above leads to a Sinkhorn-based ROTP layer $f_{\text{rot}}(\bm{X};\bm{\theta})$. 
As illustrated in Fig.~\ref{fig:ff1}, this layer is implemented by unrolling the above iterative scheme by stacking $T$ proximal point modules, and each module is implemented by $K$ Sinkhorn-scaling steps. 
Furthermore, we can leverage the logarithmic stabilization strategy~\cite{chizat2018scaling,schmitzer2019stabilized} to improve the numerical stability of the Sinkhorn-scaling algorithm --- instead of updating $\bm{T}(\bm{a}^{(k)},\bm{b}^{(k)})$ directly, we can update $\log \bm{T}(\bm{a}^{(k)},\bm{b}^{(k)})$. 
Accordingly, the exponential and scaling operations in~\eqref{eq:sinkhorn} are integrated as ``LogSumExp'' operations, which helps to avoid numerical instability issues. 
Algorithm~\ref{alg:sinkhorn} shows the Sinkhorn-based ROTP layer implemented based on the stabilized Sinkhorn-scaling algorithm.
Here, LogSumExp$_{\text{col}}$ and LogSumExp$_{\text{row}}$ apply column-wise and row-wise summation, respectively.

\subsection{Bregman ADMM-based ROTP layer}
The Sinkhorn-based ROTP layer requires the smoothness regularizer $\text{R}(\bm{P})$ to be entropic, which limits its generalizability. 
Additionally, even if applying the logarithmic stabilization strategy, the Sinkhorn-scaling algorithm still has a high risk of numerical instability because the parameters $\alpha_0$ and $\alpha_1$ may change in a wide range during training. 
To overcome the challenges, we apply a Bregman alternating direction method of multipliers (Bregman ADMM or BADMM) to build another ROTP layer. 
In particular, we first rewrite~\eqref{eq:uot} in an equivalent format by introducing three auxiliary variables $\bm{S}$, $\bm{\mu}$ and $\bm{\eta}$:
\begin{eqnarray}\label{eq:badmm1}
\begin{aligned}
    &\sideset{}{_{\bm{P},\bm{S},\bm{\mu},\bm{\eta}}}\min \langle -\bm{X},\bm{P}\rangle
    +\alpha_0\langle C(\bm{X},\bm{S}),\bm{P} \rangle\\ 
    &\quad\quad+\alpha_1\text{R}(\bm{S}) + \alpha_2\text{KL}(\bm{\mu}|\bm{p}_0) + \alpha_3\text{KL}(\bm{\eta}|\bm{q}_0)\\
    &s.t.~\bm{P}=\bm{S},~\bm{P}\bm{1}_N=\bm{\mu},~\bm{S}^T\bm{1}_D=\bm{\eta}.
\end{aligned}
\end{eqnarray}
These three auxiliary variables correspond to the joint distribution $\bm{P}$ and its marginals. 
This problem can be further rewritten in a Bregman augmented Lagrangian form by introducing three dual variables $\bm{Z}$, $\bm{z}_1$, $\bm{z}_2$ for the three constraints in~\eqref{eq:badmm1}, respectively. 
For the ROT problem with auxiliary variables, we can write its Bregman augmented Lagrangian form as
\begin{eqnarray}\label{eq:bregmanAL}
\begin{aligned}
&\sideset{}{_{\bm{P},\bm{S},\bm{\mu},\bm{\eta},\bm{Z},\bm{z}_1,\bm{z}_2}}\min\langle-\bm{X} + \alpha_0 C(\bm{X},\bm{S}),\bm{P}\rangle \\
&+\underbrace{\alpha_1\text{R}(\bm{S},\bm{P})}_{\text{Regularizer 1}} + 
\underbrace{\langle\bm{Z},\bm{P}-\bm{S}\rangle + \rho\text{Div}(\bm{P},\bm{S})}_{\text{Constraint 1, for $\bm{P}$ and $\bm{S}$}} \\ 
&+\underbrace{\alpha_2\text{KL}(\bm{\mu}|\bm{p}_0)}_{\text{Regularizer 2}} 
+\underbrace{\langle\bm{z}_1,\bm{\mu}-\bm{P}\bm{1}_N\rangle +
\rho\text{Div}(\bm{\mu},\bm{P}\bm{1}_N)}_{\text{Constraint 2, for $\bm{\mu}$ and $\bm{P}$}} \\
&+\underbrace{ \alpha_3\text{KL}(\bm{\eta}|\bm{q}_0)}_{\text{Regularizer 3}}
+\underbrace{\langle\bm{z}_2,\bm{\eta}-\bm{S}^T\bm{1}_D\rangle +
\rho\text{Div}(\bm{\eta},\bm{S}^T\bm{1}_N)}_{\text{Constraint 3, for $\bm{\eta}$ and $\bm{S}$}}.
\end{aligned}
\end{eqnarray}
Here, $\text{Div}(\cdot,\cdot)$ represents the Bregman divergence term, which is implemented as the KL-divergence as the work in~\cite{wang2014bregman,xu2020gromov} did.
Its significance is controlled by $\rho>0$.
The last three lines of~\eqref{eq:bregmanAL} contain the Bregman augmented Lagrangian terms, which correspond to the three constraints in~\eqref{eq:badmm1}. 
Here, $\text{R}(\bm{S},\bm{P})$ corresponds to the smoothness regularizer. 
When applying the entropic regularizer, we set $\text{R}(\bm{S},\bm{P})=\langle\bm{S},\log\bm{S}-\bm{1}\rangle$. 
When applying the quadratic regularizer, we set $\text{R}(\bm{S},\bm{P})=\langle\bm{S},\bm{P}\rangle$.

In particular, we solve the ROT problem by alternating optimization: 
At the $t$-th iteration, we first update $\bm{P}$ while fix other variables. 
We can ignore Constraint 3 and the three regularizers (because they are irrelevant to $\bm{P}$) and write Constraint 2 explicitly. 
The problem becomes:
\begin{eqnarray}\label{eq:updateP}
\begin{aligned}
&\sideset{}{_{\bm{P}\in\Pi(\bm{\mu}^{(t)},\cdot)}}\min\langle-\bm{X} - \alpha_0\bm{\Sigma}_1\bm{S}^{(t)}\bm{\Sigma}_2^T,\bm{P}\rangle + \alpha_1\text{R}(\bm{S}^{(t)},\bm{P}) \\ &+\langle\bm{Z}^{(t)},\bm{P}-\bm{S}^{(t)}\rangle + \rho\text{KL}(\bm{P}|\bm{S}^{(t)}),
\end{aligned}
\end{eqnarray}
where $\Pi(\bm{\mu}^{(t)},\cdot)=\{\bm{P}>\bm{0}|\bm{P}\bm{1}_N=\bm{\mu}^{(t)}\}$ is the one-side constraint of $\bm{P}$. 
We can derive the closed-form solution of this problem based on the first-order optimality condition:
\begin{eqnarray}\label{eq:deriveP}
\begin{aligned}
    \log\bm{P}^{(t+1)}=(\log\bm{\mu}^{(t)}-\text{LogSumExp}_{\text{row}}(\bm{Y}))\bm{1}_N^T+\bm{Y},
\end{aligned}
\end{eqnarray}
where $\bm{Y}=$
\begin{eqnarray*}
    \begin{cases}
    \frac{\bm{X}+\alpha_0\bm{\Sigma}_1\bm{S}^{(t)}\bm{\Sigma}_2^T-\bm{Z}^{(t)}+\rho\log\bm{S}^{(t)}}{\rho}, & \text{entropic }\text{R}(\bm{S},\bm{P})\\
    \frac{\bm{X}+\alpha_0\bm{\Sigma}_1\bm{S}^{(t)}\bm{\Sigma}_2^T-\alpha_1\bm{S}^{(t)}-\bm{Z}^{(t)} + \rho\log\bm{S}^{(t)}}{\rho}, & \text{quadratic }\text{R}(\bm{S},\bm{P}).\\
    \end{cases}    
\end{eqnarray*}

Given $\bm{P}^{(t+1)}$, we can update the auxiliary variables $\bm{S}$ in a similar manner: we ignore Constraint 2, Regularizers 2 and 3, and write Constraint 3 explicitly.
Then, the optimization problem of $\bm{S}$ becomes
\begin{eqnarray}\label{eq:updateS}
\begin{aligned}
&\sideset{}{_{\bm{S}\in\Pi(\cdot,\bm{\eta}^{(t)})}}\min \langle-\alpha_0\bm{\Sigma}_1^T\bm{P}^{(t+1)}\bm{\Sigma}_2,\bm{S}\rangle +\alpha_1\text{R}(\bm{S},\bm{P}^{(t+1)})\\
&+\langle\bm{Z}^{(t)},\bm{P}^{(t+1)}-\bm{S}\rangle + \rho\text{KL}(\bm{S}|\bm{P}^{(t+1)}),
\end{aligned}
\end{eqnarray}
where $\Pi(\cdot,\bm{\eta}^{(t)})=\{\bm{S}>\bm{0}|\bm{S}^T\bm{1}_D=\bm{\eta}^{(t)}\}$. 
Similarly, we have
\begin{eqnarray}\label{eq:deriveS}
\begin{aligned}
\log\bm{S}^{(t+1)}=\bm{1}_D(\log\bm{\eta}^{(t)}-\text{LogSumExp}_{\text{col}}(\bm{Y}))^T +\bm{Y},
\end{aligned}
\end{eqnarray}
where $\bm{Y}=$
\begin{eqnarray*}
    \begin{cases}    \frac{\bm{Z}^{(t)}+\alpha_0\bm{\Sigma}_1^T\bm{P}^{(t+1)}\bm{\Sigma}_2+\rho\log\bm{P}^{(t+1)}}{\alpha_1+\rho}, &\text{entropic }\text{R}(\bm{S},\bm{P})\\
    \frac{\alpha_0\bm{\Sigma}_1^T\bm{P}^{(t+1)}\bm{\Sigma}_2-\alpha_1 \bm{P}^{(t+1)}+\bm{Z}^{(t)} + \rho\log\bm{P}^{(t+1)}}{\rho}, &\text{quadratic }\text{R}(\bm{S},\bm{P}).
    \end{cases}    
\end{eqnarray*}

Given $\bm{P}^{(t+1)}$ and $\bm{S}^{(t+1)}$, we can update $\bm{\mu}$ and $\bm{\eta}$ by solving their corresponding Bregman augmented Lagrangian optimization problems, whose solutions have closed forms as well based on the first-order optimality condition. 
In particular, when updating $\bm{\mu}$, we ignore the terms irrelevant to $\bm{\mu}$ in~\eqref{eq:bregmanAL} and leverage the constraint $\bm{S}^{(t+1)}\bm{1}_N=\bm{\mu}^{(t)}$ explicitly. 
Accordingly, the problem becomes
\begin{eqnarray}\label{eq:update_mu}
\begin{aligned}
&\sideset{}{}\min_{\bm{\mu}\in\Delta^{D-1}} \alpha_2\text{KL}(\bm{\mu}|\bm{p}_0) + \langle\bm{z}_1^{(t)},\bm{\mu}-\bm{\mu}^{(t)}\rangle+ \rho\text{KL}(\bm{\mu}|\bm{\mu}^{(t)})\\
&\Rightarrow\bm{\mu}^{(t+1)}=\sigma\Bigl( \underbrace{\frac{\rho \log\bm{\mu}^{(t)} + \alpha_2\log\bm{p}_0 - \bm{z}_1^{(t)}}{\rho + \alpha_2}}_{\text{Denoted as~}\bm{y}} \Bigr)\\
&\xRightarrow{\text{Logarithmic Update}}\log\bm{\mu}^{(t+1)}=\bm{y} - \text{LogSumExp}(\bm{y}).
\end{aligned}
\end{eqnarray}
Here, the softmax $\sigma(\cdot)$ and $\text{LogSumExp}(\cdot)$ are operations for vectors.

Similarly, when updating $\bm{\eta}$, we ignore the terms irrelevant to $\bm{\eta}$ in~\eqref{eq:bregmanAL} and leverage the constraint $(\bm{P}^{(t+1)})^T\bm{1}_D=\bm{\eta}^{(t)}$ explicitly. 
The problem becomes
\begin{eqnarray}\label{eq:update_eta}
\begin{aligned}
&\sideset{}{}\min_{\bm{\eta}\in\Delta^{N-1}} \alpha_3\text{KL}(\bm{\eta}|\bm{q}_0) + \langle\bm{z}_2^{(t)},\bm{\eta}-\bm{\eta}^{(t)}\rangle + \rho\text{KL}(\bm{\eta}|\bm{\eta}^{(t)})\\
&\Rightarrow \bm{\eta}^{(t+1)}=\sigma\Bigl( \underbrace{\frac{\rho \log\bm{\eta}^{(t)} + \alpha_3\log\bm{q}_0 - \bm{z}_2^{(t)}}{\rho + \alpha_3}}_{\text{Denoted as~}\bm{y}} \Bigr)\\
&\xRightarrow{\text{Logarithmic Update}}\log\bm{\eta}^{(t+1)}=\bm{y} - \text{LogSumExp}(\bm{y}).
\end{aligned}
\end{eqnarray}

Finally, we update the dual variables by
\begin{eqnarray}\label{eq:zs}
\begin{aligned}
    &\bm{Z}^{(t+1)}=\bm{Z}^{(t)} + \rho(\bm{P}^{(t+1)}-\bm{S}^{(t+1)}),\\
    &\bm{z}_1^{(t+1)}=\bm{z}_1^{(t)} + \rho(\bm{\mu}^{(t+1)} -\bm{P}^{(t+1)}\bm{1}_{N}),\\
    &\bm{z}_2^{(t+1)}=\bm{z}_2^{(t)} + \rho(\bm{\eta}^{(t+1)} -(\bm{S}^{(t+1)})^T\bm{1}_{D}).
\end{aligned}
\end{eqnarray}

\begin{figure}[t]
\centering
    \includegraphics[width=0.9\linewidth]{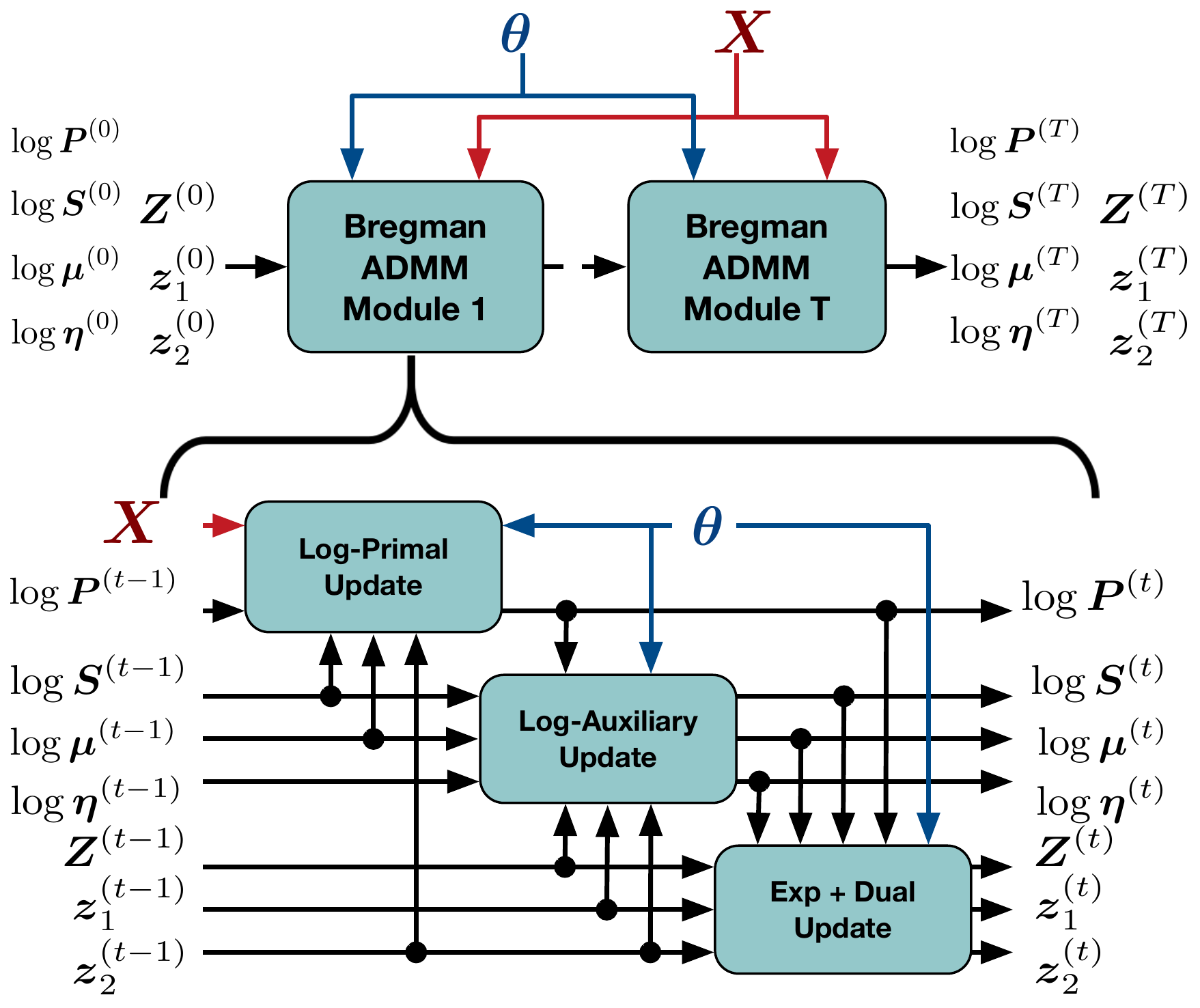}
\caption{An illustration of the BADMM-based ROTP layer. 
The red, blue, and black arrows correspond to the computational flow of data, model parameters, intermediate results.}\label{fig:ff}
\end{figure}

\begin{algorithm}[t]
    \caption{$f_{\text{rot}}(\bm{X};\bm{\theta})$ based on Bregman ADMM}\label{alg:badmm_uotp}
	\begin{algorithmic}[1]
	    \REQUIRE Data $\bm{X}$ and parameters $\bm{\theta}$. 
	    \STATE Primal and auxiliary variable $\log\bm{P}^{(0)}=\log\bm{S}^{(0)}=\log(\bm{p}_0\bm{q}_0^T)$, $\log\bm{\mu}^{(0)}=\log\bm{p}_0$, $\log\bm{\eta}^{(0)}=\log\bm{q}_0$.
	    \STATE Dual variables $\bm{Z}^{(0)}=\bm{0}_{D\times N}$, $\bm{z}_1^{(0)}=\bm{0}_D$, $\bm{z}_2^{(0)}=\bm{0}_N$.
	    \FOR{$t=0,...,T-1$}
	        \STATE Derive $\log\bm{P}^{(t+1)}$ by~\eqref{eq:deriveP}.
	        \STATE Derive $\log\bm{S}^{(t+1)}$ by~\eqref{eq:deriveS}. 
	        \STATE Derive $\log\bm{\mu}^{(t+1)}$ by~\eqref{eq:update_mu}.
	        \STATE Derive $\log\bm{\eta}^{(t+1)}$ by~\eqref{eq:update_eta}.
	        \STATE Update dual variables $\bm{Z},\bm{z}_1,\bm{z}_2$ by~\eqref{eq:zs}.
	   \ENDFOR
	   \STATE $\bm{P}^{*}:=\bm{P}^{(T)}$, and apply~\eqref{eq:uotp} to obtain $f_{\text{rot}}(\bm{X};\bm{\theta})$.
	\end{algorithmic}
\end{algorithm}

It is easy to find that the above Bregman ADMM algorithm can be viewed as a variant of the proximal point method, in which the proximal term is implemented as a set of Bregman divergence terms. 
When deriving $\bm{P}^{(t+1)}$, our Bregman ADMM applies the auxiliary variable $\bm{S}^{(t)}$, rather than the previous estimation $\bm{P}^{(t)}$, to regularize it. 
Taking the above steps in an iteration as a module, we can implement our ROTP layer by stacking such modules, as shown in Fig.~\ref{fig:ff}. 
Additionally, as shown in~\eqref{eq:deriveP}-\eqref{eq:update_eta}, both the primal and auxiliary variables can be updated in their logarithmic formats, which improves the numerical stability of our algorithm. 
In summary, Algorithm~\ref{alg:badmm_uotp} shows the scheme of our BADMM-based ROTP layer.

%% file: tex/compare.tex
\section{Further Analysis and Comparisons}\label{sec:cmp}
\subsection{Comparisons for various ROTP layers}\label{ssec:implement}

Applying different smoothness regularizers and algorithms, we consider three ROTP layers: The Sinkhorn-based ROTP layer is denoted as \textbf{ROTP$_{\text{S}}$}, 
the BADMM-based ROTP layer with the entropic smoothness regularizer is denoted as \textbf{ROTP$_{\text{B-E}}$}, and the BADMM-based ROTP layer with the quadratic smoothness regularizer is denoted as \textbf{ROTP$_{\text{B-Q}}$}. 
In this subsection, we will analyze their convergence, complexity, approximation power, and numerical stability in depth.

\subsubsection{Convergence}
\begin{figure}[t]
    \centering
    \includegraphics[height=4.5cm]{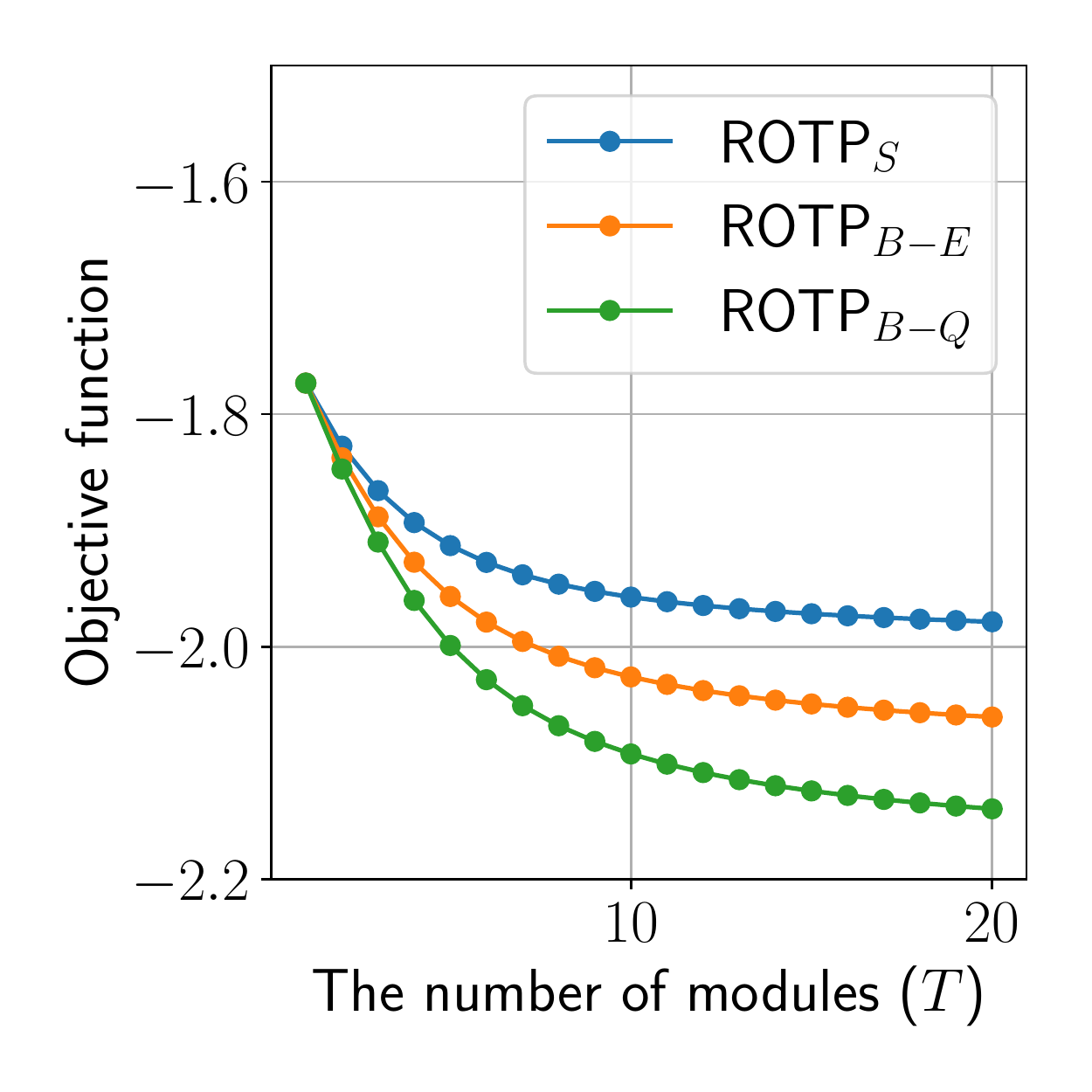}
    \caption{Given a batch of 50 sample sets, in which each sample set contains five hundred 100-dimensional samples, we compare the ROTP layers on their convergence.}
    \label{fig:convergence}
\end{figure}

As illustrated in Figs.~\ref{fig:ff1} and~\ref{fig:ff}, each ROTP layer is implemented by stacking $T$ feed-forward modules. 
Given the same data matrix and fixed model parameters, we solve the ROT problem in~\eqref{eq:uot} by different ROTP layers and record the change of the expectation term $\langle -\bm{X},\bm{P}\rangle$ with the increase of $T$. 
The comparison is shown in Fig.~\ref{fig:convergence}. 
We can find that all three layers make their objective functions converge when using more than $16$ feed-forward modules. 
However, applying different smoothness regularizers and algorithms lead to different convergence rates and optimization trajectories --- given the same number of modules, the three layers often converge to different optimums.

\subsubsection{Computational complexity} 
\begin{figure}[t]
    \centering
    \subfigure[Runtime w.r.t. the number of iterations (with $D=5$, $N=50$, $K=5$, $\alpha_0>0$)]{
    \includegraphics[height=3.5cm]{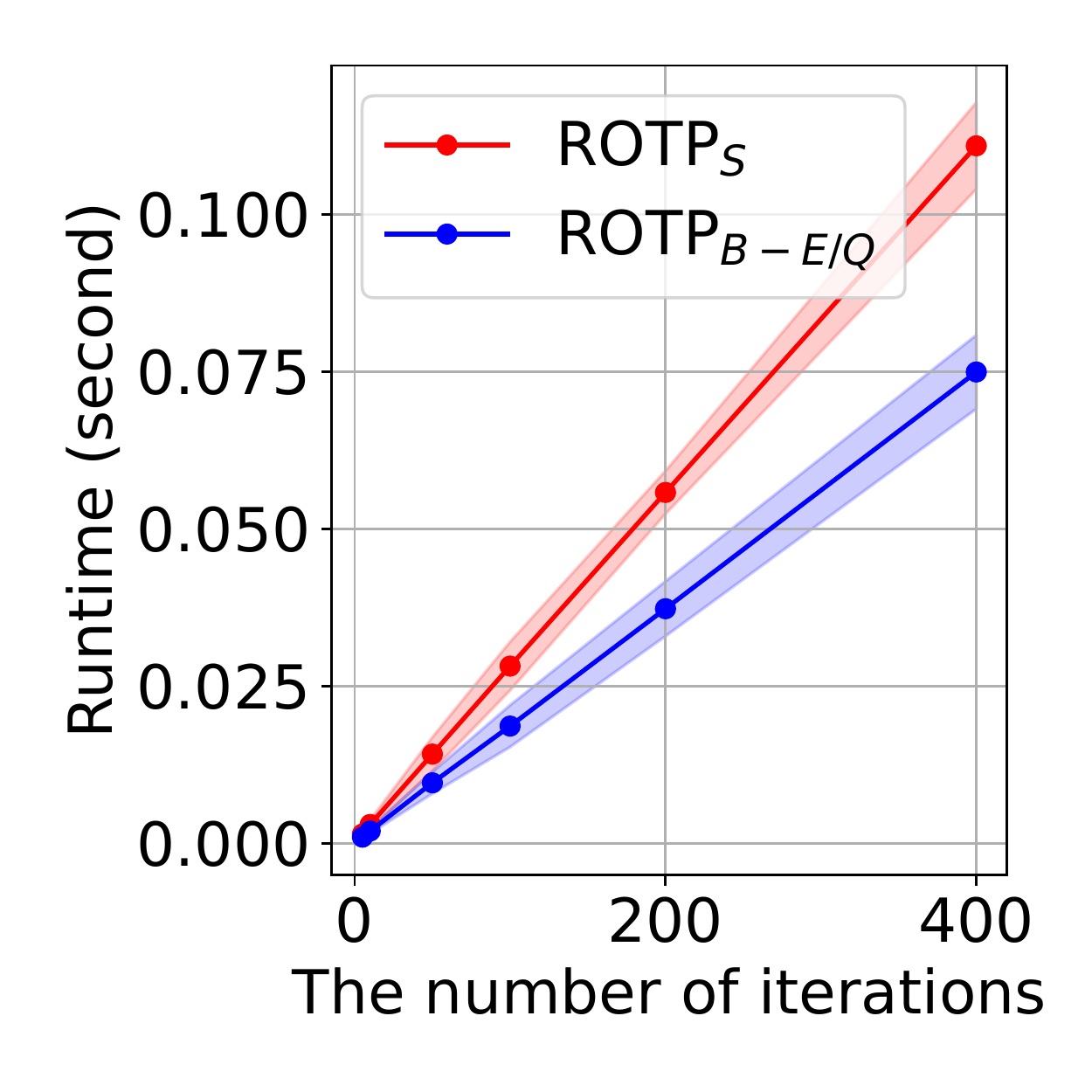}\label{fig:rot_kt}
    }\quad
    \subfigure[Runtime w.r.t. the number of samples (with $D=5$, $T=50$, $\alpha_0>0$)]{
    \includegraphics[height=3.5cm]{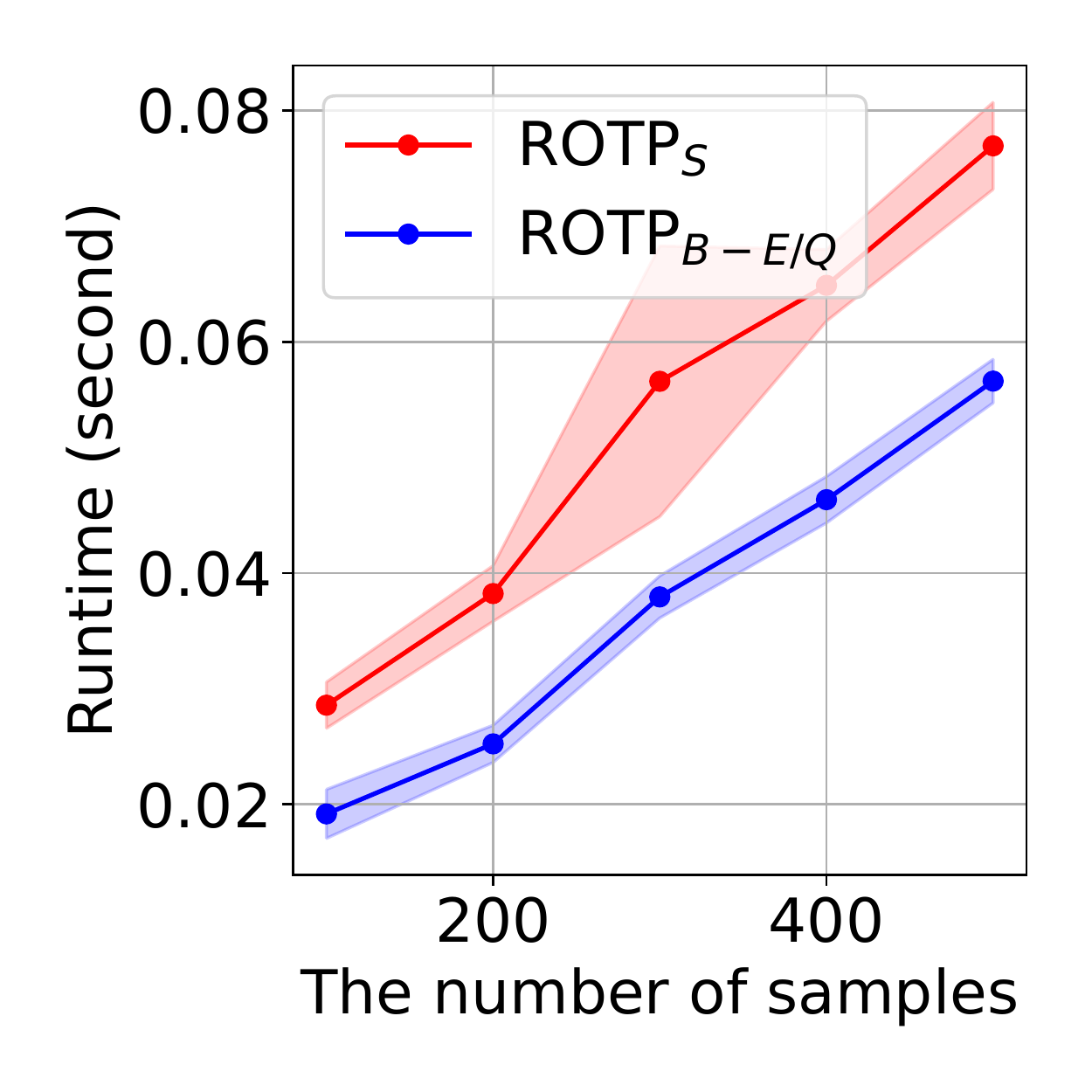}\label{fig:rot_nt}
    }\\
    \subfigure[Runtime w.r.t. the number of iterations (with $D=5$, $N=50$, $\alpha_0=0$)]{
    \includegraphics[height=3.5cm]{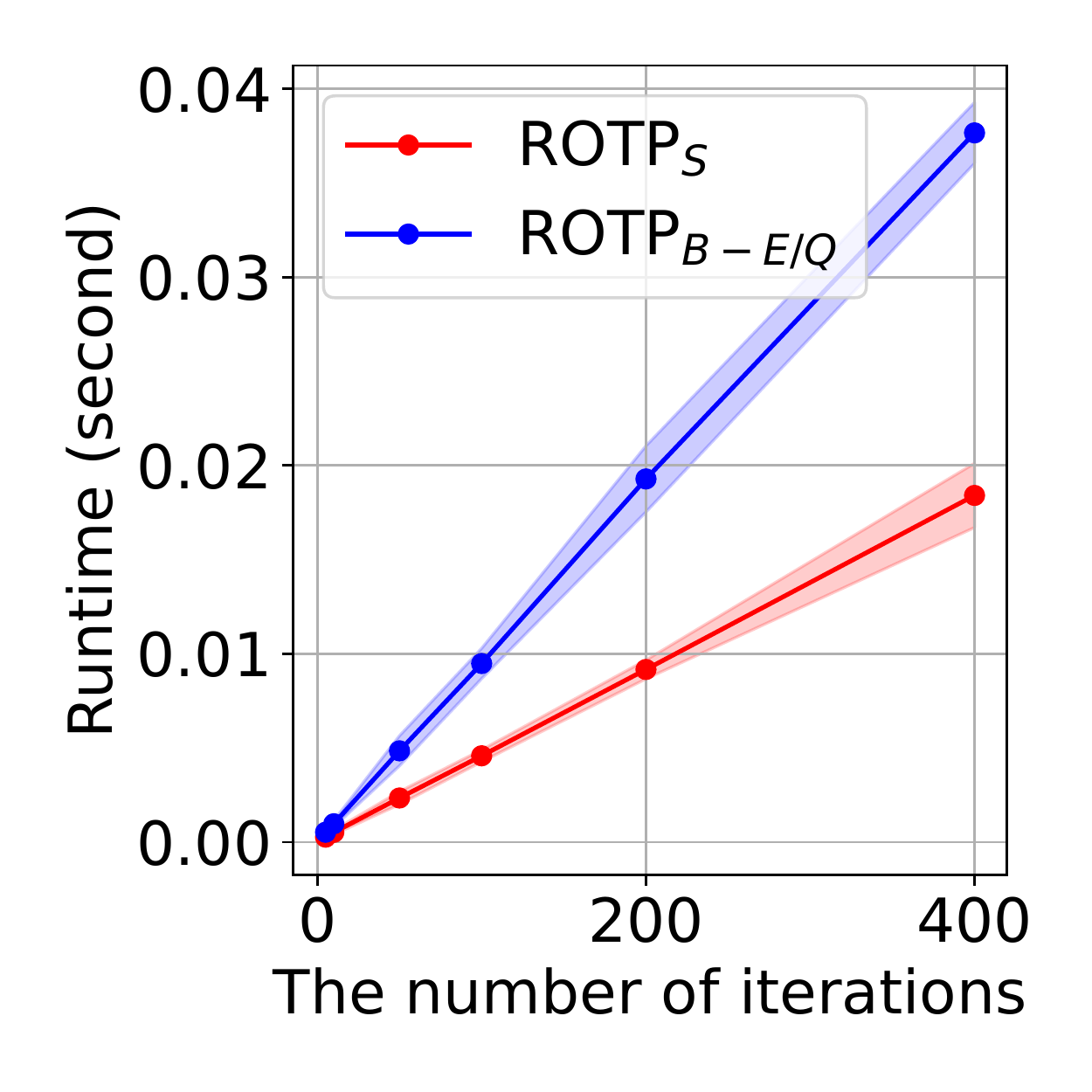}\label{fig:uot_kt}
    }\quad
    \subfigure[Runtime w.r.t. the number of samples (with $D=5$, $T~\text{or}~K=50$, $\alpha_0=0$)]{
    \includegraphics[height=3.5cm]{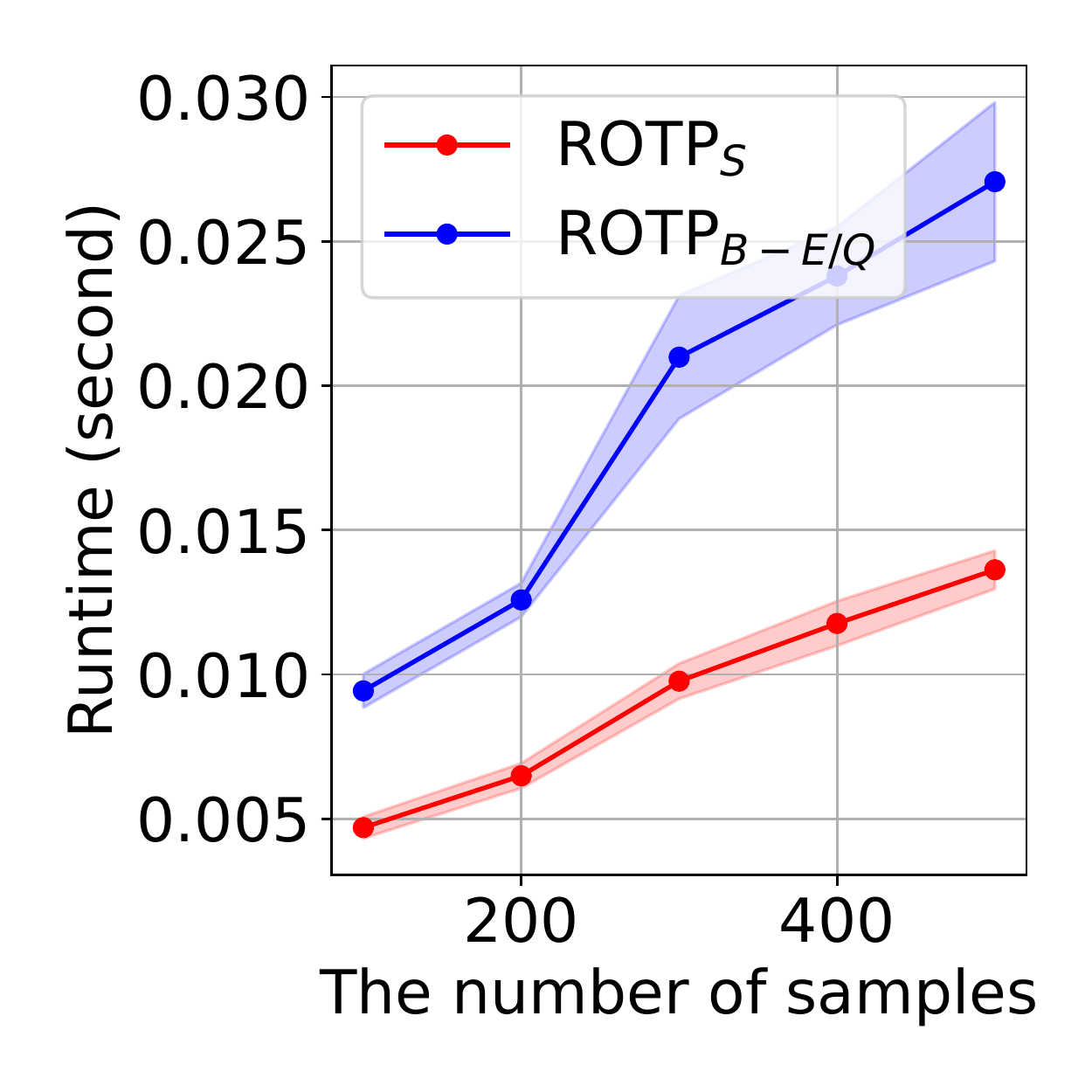}\label{fig:uot_nt}
    }
    \caption{Comparisons of the two ROTP layers on their feed-forward runtime on a single CPU.
    We plot the averaged runtime and its standard deviation in $20$ trials.
    For the BADMM-based ROTP layer, we just consider the case using the entropic smoothness regularizer because its runtime is stable with respect to the type of the regularizer.}
    \label{fig:runtime}
\end{figure}

Each Sinkhorn-scaling module contains $K$ Sinkhorn iterations, while each Bregman ADMM module corresponds to one-step updates of the primal, auxiliary, and dual variables. 
As a result, each Sinkhorn-scaling module involves $2K$ LogSumExp functions (the most time-consuming process), while each BADMM module merely requires two LogSumExp functions. 
Based on the analysis above, the computational complexity of the BADMM-based layer is lower than that of the Sinkhorn-based layer. 
In particular, given a set of $N$ $D$-dimensional samples, the similarity matrices ($\bm{\Sigma}_1$ and $\bm{\Sigma}_2$) are computed with the complexity $\mathcal{O}(N^2D + D^2N)$. 
Taking the samples and the similarity matrices as input, the computational complexity of the Sinkhorn-based ROTP layer is $\mathcal{O}(T(N^2D + D^2N + KND))$, where $N^2D + D^2N$ corresponds to the computation of $\bm{\Sigma}_1\bm{P}^{(t)}\bm{\Sigma}_2^T$ per step, and $KND$ corresponds to the $K$ Sinkhorn iterations within each Sinkhorn-scaling module. 
The computational complexity of the BADMM-based ROTP layer is $\mathcal{O}(T(N^2D+D^2N))$, where $N^2D + D^2N$ corresponds to the computation of $\bm{\Sigma}_1\bm{P}^{(t)}\bm{\Sigma}_2^T$ (and $\bm{\Sigma}_1^T(\bm{S}^{(t)})^T\bm{\Sigma}_2$). 
Figs.~\ref{fig:rot_kt} and~\ref{fig:rot_nt} verify the above analysis further. 
The runtime of these two ROTP layers under different $T$'s and $N$'s indicates that the Sinkhorn-based ROTP layer is slower than the BADMM-based ROTP layer. 

When setting $\alpha_0=0$, the ROT problem in~\eqref{eq:uot} degrades to a classic unbalanced optimal transport (UOT) problem. 
In such a situation, the computation of $\bm{\Sigma}_1\bm{P}\bm{\Sigma}_2^T$ is avoided, which leads to lower complexity for both of the layers. 
Especially for the Sinkhorn-based ROTP layer, when $\alpha_0=0$, it only requires a single Sinkhorn-scaling module with $K$ iterations to obtain the optimal transport matrix, which avoids the nested iterative optimization. 
Accordingly, its complexity becomes $\mathcal{O}(KND)$. 
The BADMM-based ROTP layer, however, still requires $T$ BADMM module, so its complexity is $\mathcal{O}(TND)$ when $\alpha_0=0$.
Figs.~\ref{fig:uot_kt} and~\ref{fig:uot_nt} show that when $\alpha_0=0$, the Sinkhorn-based ROTP layer is faster than the BADMM-based ROTP layer.

\subsubsection{Precision on approximating existing pooling layers} 
\begin{figure}[t]
\centering
\includegraphics[width=1\linewidth]{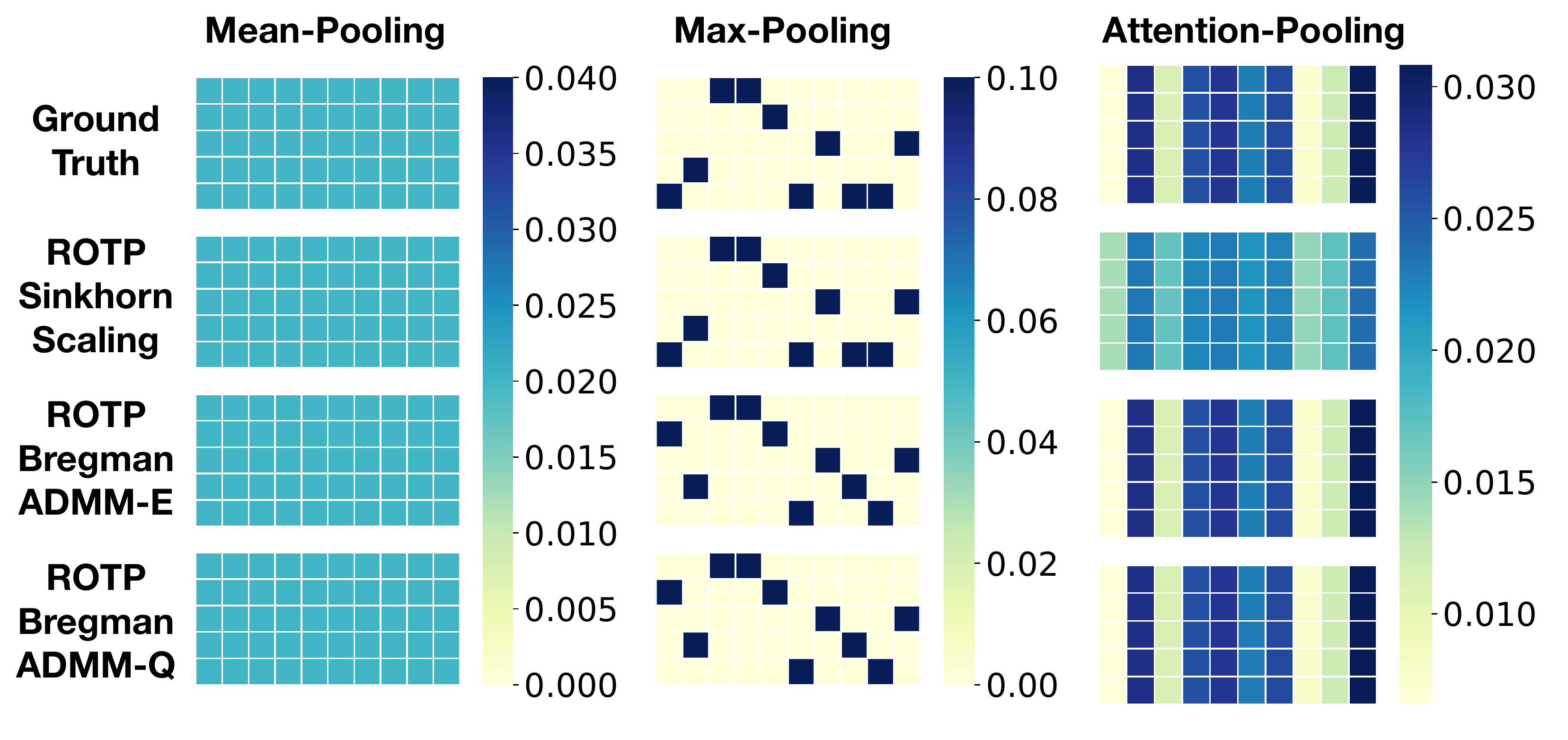}
\caption{In each column, the matrices from top to bottom are the ground truth and the $\bm{P}^{*}$'s obtained by different ROTP layers. 
Here, we set $\bm{X}\in\mathbb{R}^{5\times 10}\sim\mathcal{N}(0, 1)$, $\alpha_1=\alpha_2=\alpha_3=10^4$ in (a, c), and $\alpha_1=\alpha_3=0.01$ and $\alpha_2=10^4$ in (b).}\label{fig:approx}
\end{figure}

Proposition~\ref{prop:equiv} demonstrates that, in theory, our ROTP layer can be equivalent to some existing pooling operations under specific settings. 
In practice, however, implementing the ROTP layer by different algorithms leads to different approximation precision. 
As shown in Fig.~\ref{fig:approx}, both the Sinkhorn-based ROTP layer and the BADMM-based ROTP layer can reproduce the functionality of mean-pooling perfectly. 
However, the Sinkhorn-based ROTP layer can approximate max-pooling with higher accuracy, while the BADMM-based ROTP layer works better on approximating the attention-pooling~\cite{ilse2018attention}. 
In other words, when approximating a specific global pooling operation by an ROTP layer, we should consider the model architecture's influence.

\begin{figure}[t]
\centering
\subfigure[ROTP$_{\text{S}}$ ($\alpha_0=0$)]{
    \includegraphics[height=3.1cm]{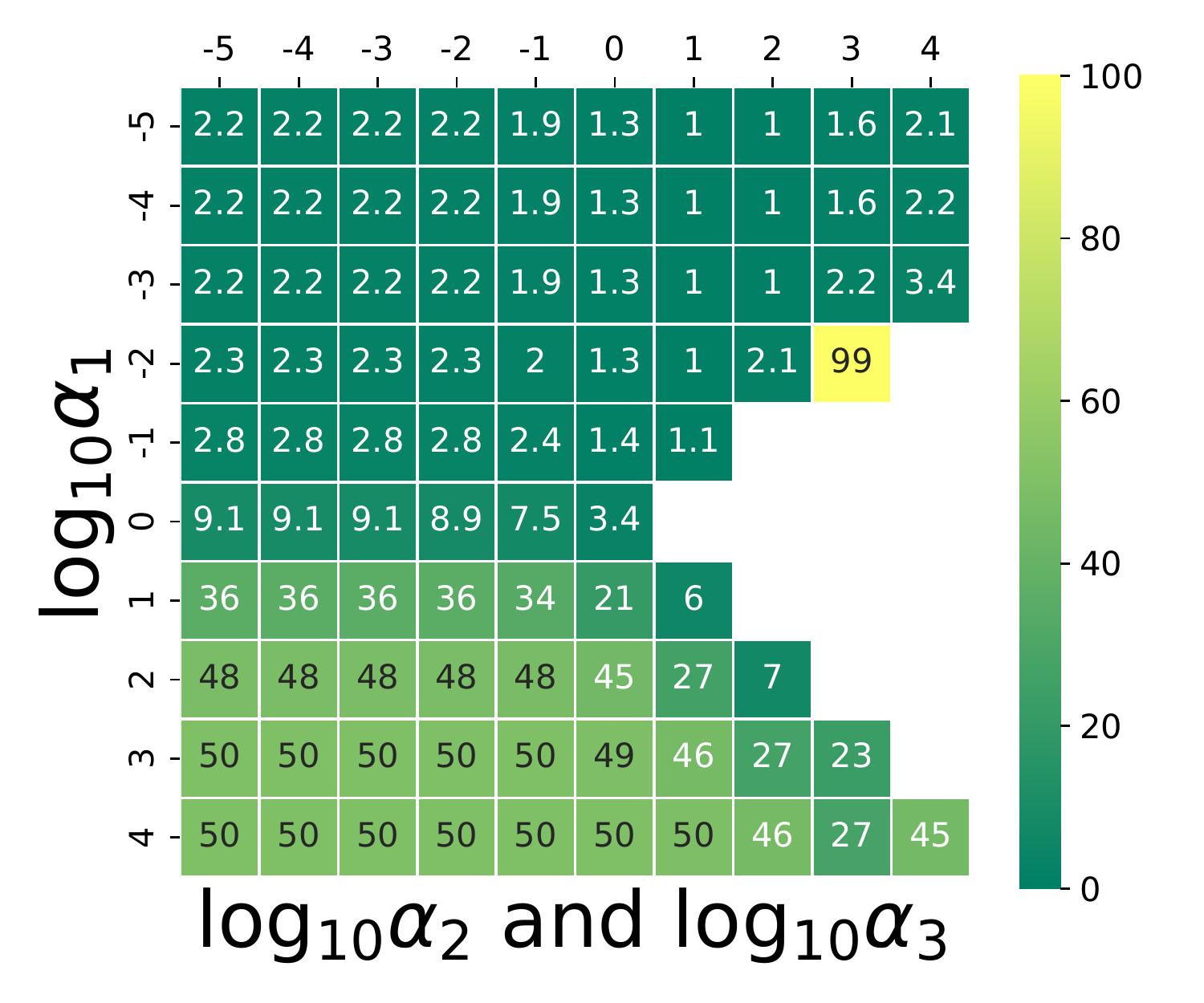}\label{fig:stability_sinkhorn1}
}
\subfigure[ROTP$_{\text{S}}$ ($\alpha_0=0.1$)]{
    \includegraphics[height=3.1cm]{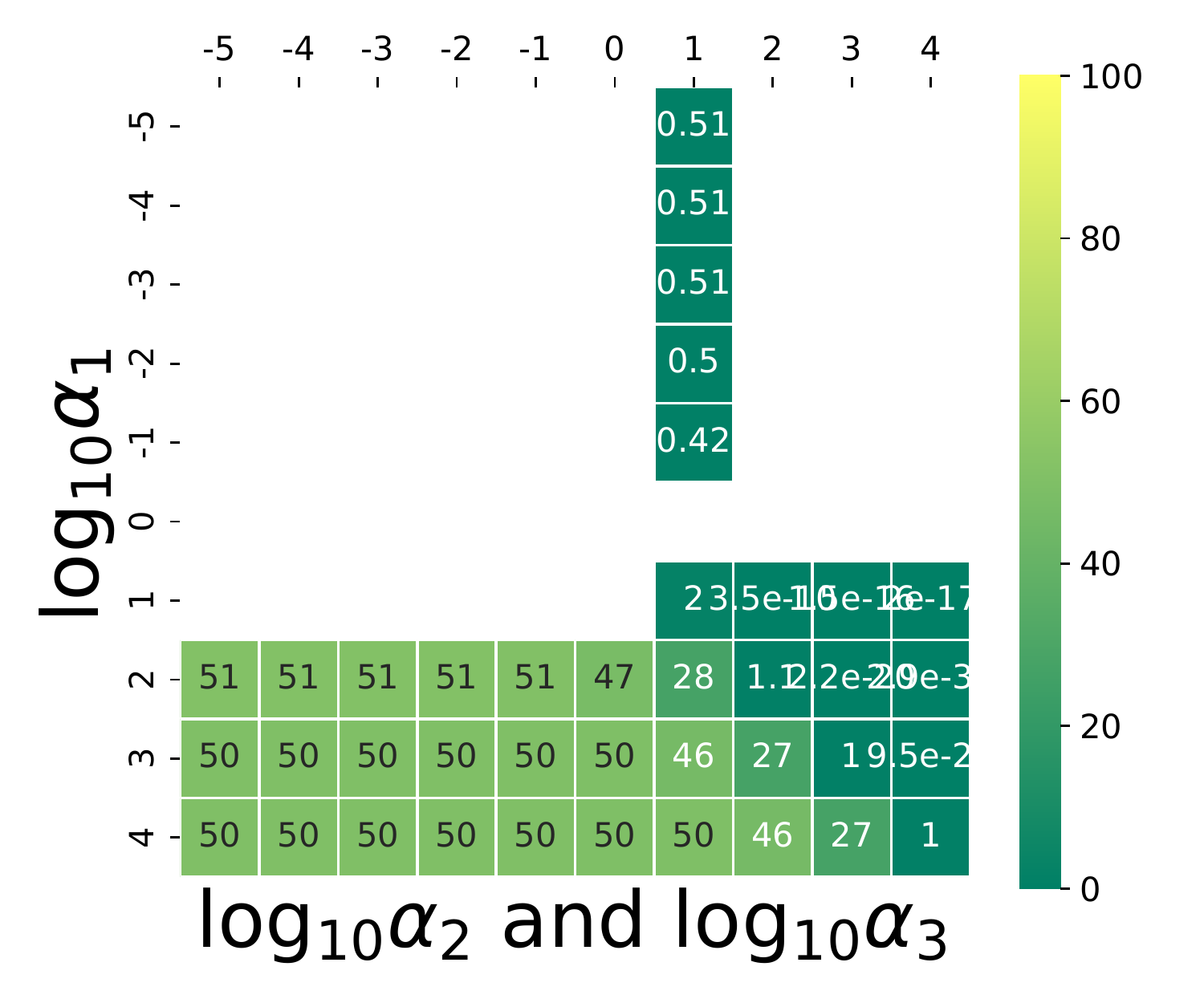}\label{fig:stability_sinkhorn2}
}
\subfigure[ROTP$_{\text{B-E/Q}}$ ($\alpha_0=0$)]{
    \includegraphics[height=3.1cm]{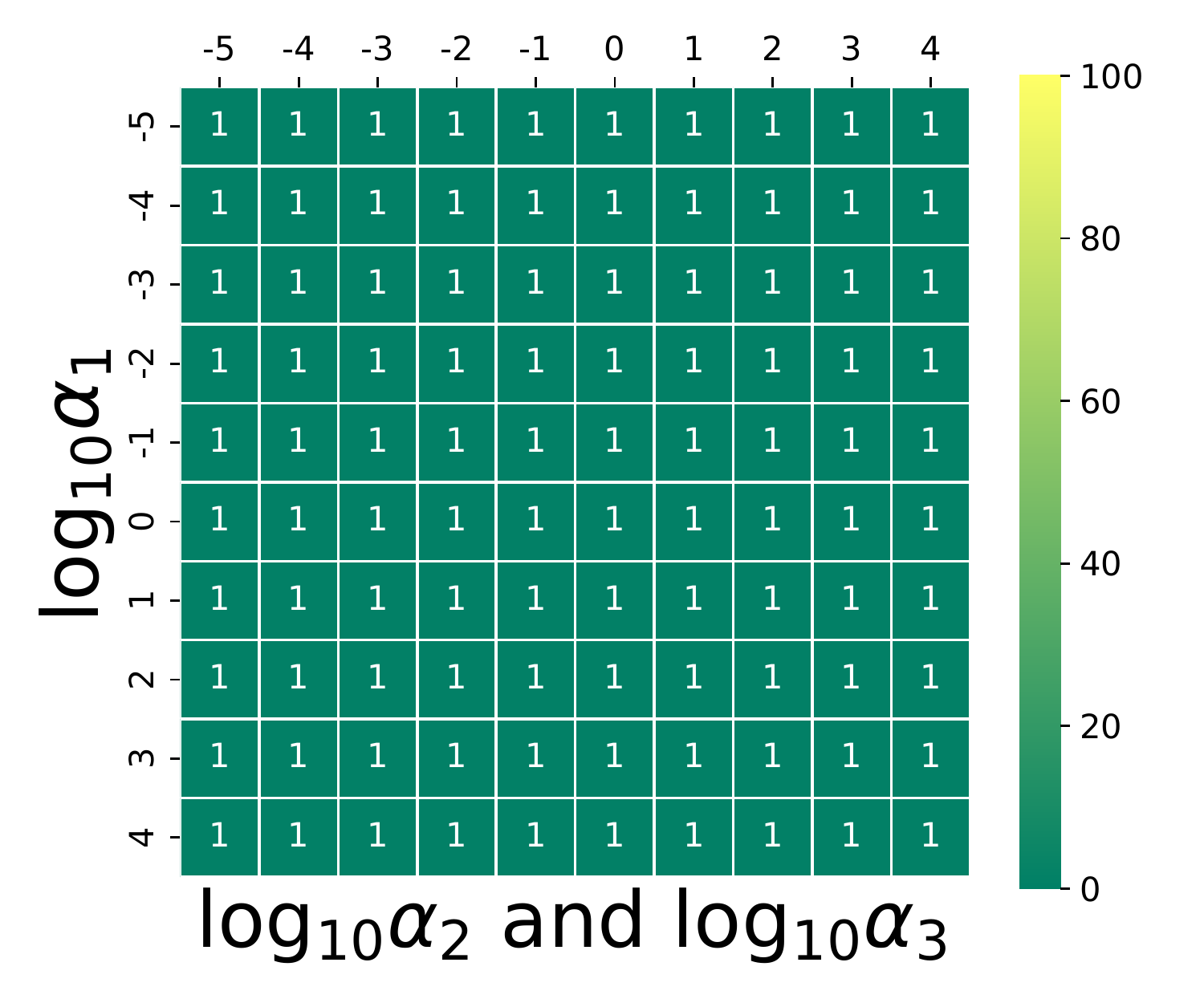}\label{fig:stability_badmme1}
}
\subfigure[ROTP$_{\text{B-E/Q}}$ ($\alpha_0=0.1$)]{
    \includegraphics[height=3.1cm]{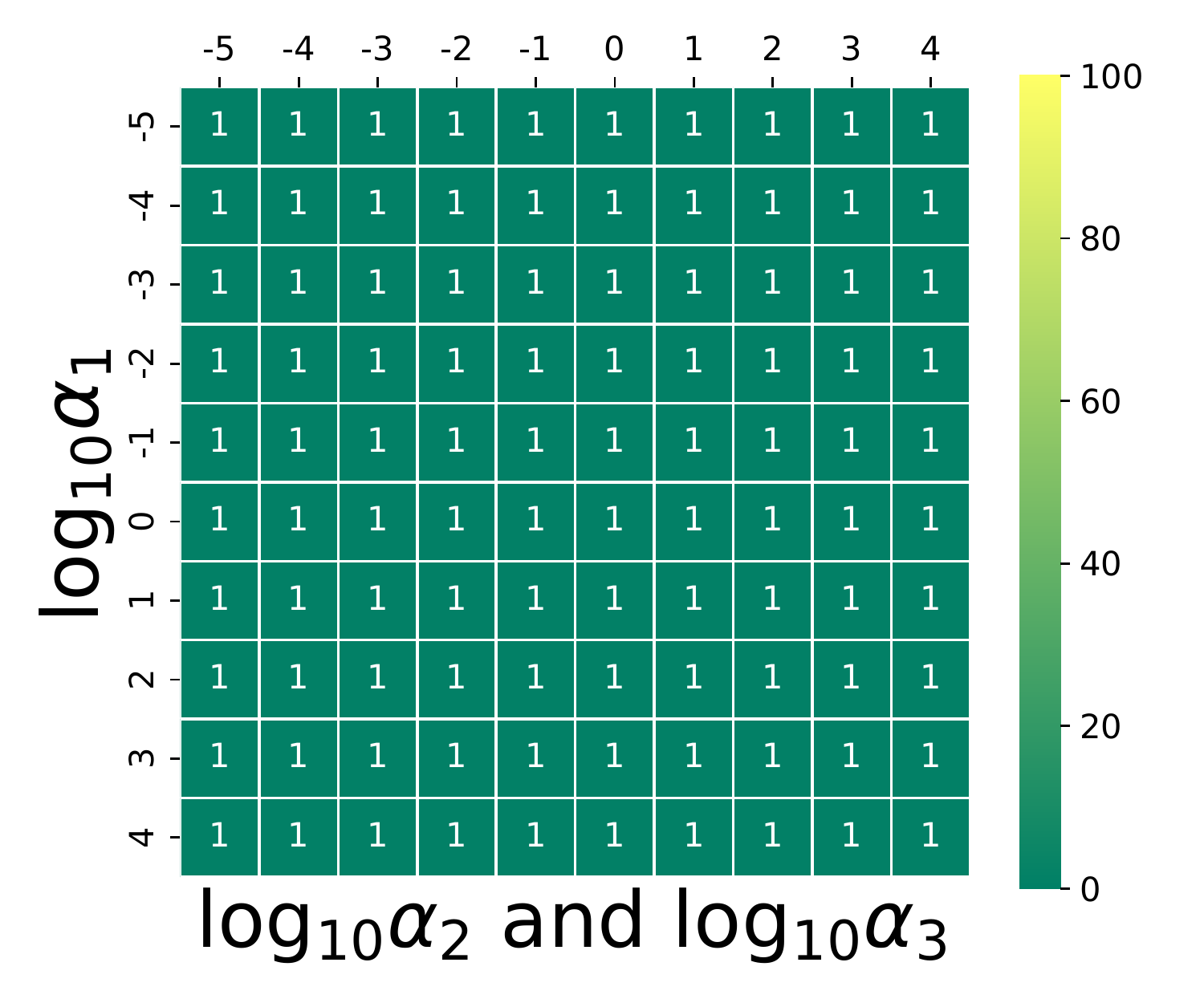}\label{fig:stability_badmme2}
}
\caption{Given $\bm{X}\in\mathbb{R}^{5\times 10}$, we learn $\bm{P}^*$'s under different configurations and calculate $\|\bm{P}^*\|_1$'s. 
Each subfigure shows the $\|\bm{P}^*\|_1$'s, and the white regions correspond to \textsf{NaN}'s.
}\label{fig:stability}
\end{figure}

\subsubsection{Numerical stability}
As aforementioned, one primary motivation for designing the BADMM-based ROTP layer is to overcome the numerical instability of the Sinkhorn-based ROTP layer. 
For the two layers, we set $\alpha_2=\alpha_3$ and select $\alpha_1,\alpha_2,\alpha_3$ from $\{10^{-5},...,10^{4}\}$. 
Under such configurations, we derive 100 $\bm{P}$'s accordingly. 
For each layer, we verify its numerical stability by checking whether $\|\bm{P}^*\|_1=\sum_{d,n}|p_{dn}|\approx 1$ and whether $\bm{P}^*$ contains \textsf{NaN} elements. 
Figs.~\ref{fig:stability_sinkhorn1} and~\ref{fig:stability_sinkhorn2} show that the Sinkhorn-based ROTP merely works under some configurations, which obtains \textsf{NaN} in many cases. 
When $\alpha_0>0$ and solving the ROT problem, the numerical stability of the Sinkhorn-based method becomes even worse. 
In the following experiments, we have to set $\alpha_0=0$, $\alpha_1 > 0.1$, $\alpha_{2},\alpha_3\in (10^{-5}, 10)$ for the Sinkhorn-based ROTP layer to ensure the stability of its training process. 
On the contrary, our BADMM-based ROTP layer owns much better numerical stability. 
As shown in Figs.~\ref{fig:stability_badmme1}-\ref{fig:stability_badmme2}, no matter whether $\alpha_0=0$ or not and which smoothness regularizer is applied, our BADMM-based ROTP layer not only keeps $\|\bm{P}^*\|_1\approx 1$ under all configurations but also avoids \textsf{NaN} elements successfully.

\subsection{Comparisons for various OT-based methods}
Compared with existing optimal transport-based pooling methods, e.g., OTK~\cite{mialon2020trainable}, WEGL~\cite{kolouri2020wasserstein}, and SWE~\cite{naderializadeh2021pooling}, our ROTP layers have better flexibility and generalization power. 
The main differences between our ROTP layers and other OT-based pooling methods can be categorized into the following three points. 

Firstly, our ROTP layers are based on an expectation-maximization framework in principle.
This framework defines optimal transport plans across sample indices and feature dimensions. 
The optimal transport plans can be interpreted as the joint distributions of the sample indices and the feature dimensions, which indicates the significant ``sample-feature'' pairs. 
On the contrary, existing OT-based pooling methods define optimal transport plans in the sample space. 
Their optimal transport plans work for pushing observed samples forward to some learnable sample clusters rather than weighting ``sample-feature'' pairs.

Secondly, in the implementation aspect, our ROTP layers correspond to a generalized ROT problem, which considers the smoothness of the objective function, the uncertainty of prior distributions, and the structural relations between samples jointly. 
The ROT problem leads to a generalized pooling framework that can unify typical pooling operations, and the ROTP layers can be used to build hierarchical ROTP modules. 
Moreover, the design of the ROTP layers is flexible and based on various algorithms. 
The weights of the ROT problem's regularizers and the prior distributions are learnable parameters in our ROTP layers.
On the contrary, existing OT-based pooling methods only consider the typical entropic OT problem (whose entropy term is with a predefined weight)~\cite{mialon2020trainable,kim2021differentiable} or the sliced Wasserstein problem~\cite{naderializadeh2021pooling}. 
Accordingly, these methods are only based on the Sinkhorn-scaling algorithm or the random projection method, and they cannot provide a generalized framework as our ROTP does.

Finally, existing OT-based pooling methods are designed for specific applications or data formats, such as graph classification~\cite{kolouri2020wasserstein,naderializadeh2021pooling} and set fusion~\cite{mialon2020trainable,kim2021differentiable}. 
In the following experiments, we will test our ROTP layers and demonstrate their feasibility in various applications.

%% file: tex/experiments.tex
\section{Experiments}\label{sec:exp} 
We demonstrate the effectiveness and superiority of our ROTP layers (\textbf{ROTP$_{\text{S}}$}, \textbf{ROTP$_{\text{B-E}}$}, and \textbf{ROTP$_{\text{B-Q}}$}) in various machine learning tasks, including multi-instance learning, graph classification, and image classification. 
Additionally, we build HROTP modules based on the ROTP layers and apply the modules in a typical graph set prediction task --- drug-drug interaction classification. 
For each learning task, we consider several datasets. 
For ROTP$_{\text{S}}$, we set $\alpha_0=0$ to avoid numerical instability. 
For ROTP$_{\text{B-E}}$ and ROTP$_{\text{B-Q}}$, we make the $\alpha_0$ learnable. 

The baselines we considered include $i)$ classic pooling operations like \textbf{Add-Pooling}, \textbf{Mean-Pooling}, and \textbf{Max-Pooling}; $ii)$ the mixed pooling operations like the \textbf{Mixed Mean-Max} and the \textbf{Gated Mean-Max} in~\cite{lee2016generalizing}; $iii)$ the learnable global pooling layers like \textbf{DeepSet}~\cite{zaheer2017deep}, \textbf{Set2Set}~\cite{vinyals2015order}, \textbf{DynamicPooling}~\cite{yan2018deep}, \textbf{GNP}~\cite{ko2021learning}, and the \textbf{Attention-Pooling} and \textbf{Gated Attention} in~\cite{ilse2018attention};
$iv)$ the attention-pooling methods for graphs, i.e., \textbf{SAGPooling}~\cite{lee2019self}, \textbf{ASAPooling}~\cite{ranjan2020asap};
and $v)$ OT-based pooling methods, i.e., OTK~\cite{mialon2020trainable}, WEGL~\cite{kolouri2020wasserstein}, and SWE~\cite{naderializadeh2021pooling}. 
The above pooling methods are trained and tested on a server with two Nvidia RTX3090 GPUs, whose key hyperparameters are set by grid search. 
For our ROTP layers, the number of the feed-forward modules $T$ is the key hyperparameter. 
According to the convergence analysis in Fig.~\ref{fig:convergence}, we set the number of the modules in the range $[4, 16]$ in the following experiments. 

\subsection{Evaluation of ROTP layers}
\subsubsection{Multi-instance learning}
We consider three MIL tasks, which correspond to a disease diagnosis dataset (Messidor~\cite{decenciere2014feedback}) and two gene ontology categorization datasets (Component and Function~\cite{blaschke2005evaluation}).
For each dataset, we learn a bag-level classifier, which embeds a bag of instances as input, merges the instances' embeddings via pooling, and finally, predicts the bag's label by a classifier. 
We use the AttentionDeepMIL in~\cite{ilse2018attention}, a representative bag-level classifier, as the backbone model and plug different pooling layers into it. 
When training the model, we apply the Adam optimizer~\cite{kingma2014adam} with a weight decay regularizer. 
The hyperparameters of the optimizer are set as follows: the learning rate is 0.0005, the weight decay is 0.005, the number of epochs is 50, and the batch size is 128. 
In this experiment, we apply four feed-forward modules to build each ROTP layer. 

\begin{table}[t]
\caption{Comparison on MIL accuracy$\pm$Std. (\%) for different pooling layers.}\label{tab:mil}
\centering
\small{
\begin{threeparttable}
    \begin{tabular}{@{\hspace{4pt}}c|c@{\hspace{8pt}}c@{\hspace{8pt}}c@{\hspace{4pt}}}
    \hline\hline
    Dataset  &
    Messidor &
    Component       & 
    Function       \\ 
    % Process     \\ 
    \hline
    $D$ &
    687 &
    200 &
    200 \\
    % 200 \\
    \#Positive bags &
    654 &
    423 &
    443 \\
    % 757 \\
    \#Negative bags &
    546 &
    2,707 &
    4,799 \\
    % 10,961 \\
    \#Instances &
    12,352 &
    36,894 &
    55,536 \\
    % 118,417 \\
    Min. bag size &
    8 &
    1 &
    1 \\
    % 1 \\
    Max. bag size &
    12 &
    53 &
    51 \\
    % 57 \\
    \hline
    Add  &  
    74.33$_{\pm\text{2.56}}$
    &
    93.35$_{\pm\text{0.98}}$
    &
    96.26$_{\pm\text{0.48}}$
    % & 
    % \textbf{97.41$_{\pm\text{0.21}}$}
    \\
    Mean & 
    74.42$_{\pm\text{2.47}}$
    &       
    93.32$_{\pm\text{0.99}}$
    &      
    96.28$_{\pm\text{0.66}}$
    % &    
    % 97.20$_{\pm\text{0.14}}$
    \\
    Max  & 
    73.92$_{\pm\text{3.00}}$
    &   
    93.23$_{\pm\text{0.76}}$
    &       
    95.94$_{\pm\text{0.48}}$
    % &      
    % 96.71$_{\pm\text{0.40}}$
    \\
    DeepSet  &  
    74.42$_{\pm\text{2.87}}$
    &     
    93.29$_{\pm\text{0.95}}$
    &  
    96.45$_{\pm\text{0.51}}$
    % & 
    % \color{red}{\textbf{97.64$_{\pm\text{0.18}}$}}
    \\
    Mixed  &  
    73.42$_{\pm\text{2.29}}$
    &
    \color{red}{\textbf{93.45$_{\pm\text{0.61}}$}}
    &   
    96.41$_{\pm\text{0.53}}$
    % &      
    % 96.96$_{\pm\text{0.25}}$
    \\
    GatedMixed  &
    73.25$_{\pm\text{2.38}}$
    &    
    93.03$_{\pm\text{1.02}}$
    &    
    96.22$_{\pm\text{0.65}}$
    % &      
    % 97.01$_{\pm\text{0.23}}$
    \\
    Set2Set  & 
    73.58$_{\pm\text{3.74}}$
    &
    93.19$_{\pm\text{0.95}}$
    &   
    96.43$_{\pm\text{0.56}}$
    % &    
    % 97.16$_{\pm\text{0.25}}$
    \\
    Attention   & 
    74.25$_{\pm\text{3.67}}$
    &
    93.22$_{\pm\text{1.02}}$
    &   
    96.31$_{\pm\text{0.66}}$
    % &    
    % \textbf{97.24$_{\pm\text{0.16}}$}
    \\
    GatedAtt  &
    73.67$_{\pm\text{2.23}}$
    &
    \textbf{93.42$_{\pm\text{0.91}}$}
    &   
    \textbf{96.51$_{\pm\text{0.77}}$}
    % &    
    % 97.18$_{\pm\text{0.14}}$
    \\
    DynamicP      & 
    73.16$_{\pm\text{2.12}}$
    &
    93.26$_{\pm\text{1.30}}$
    &   
    \textbf{96.47$_{\pm\text{0.58}}$}
    % &    
    % 97.03$_{\pm\text{0.14}}$
    \\
    GNP &
    73.54$_{\pm\text{3.68}}$
    &
    92.86$_{\pm\text{1.96}}$
    &   
    96.10$_{\pm\text{1.03}}$
    % &    
    % 96.03$_{\pm\text{0.67}}$
    \\
    OTK &
    74.78$_{\pm\text{2.89}}$
    &
    93.19$_{\pm\text{0.93}}$
    &
    96.31$_{\pm\text{1.02}}$
    \\
    SWE &
    74.46$_{\pm\text{3.72}}$
    &
    93.32$_{\pm\text{1.26}}$
    &
    96.42$_{\pm\text{0.88}}$
    \\
    \hline
    ROTP$_{\text{S}}$  &  
    \color{red}{\textbf{75.42$_{\pm\text{2.96}}$}}
    &
    93.29$_{\pm\text{0.83}}$
    &   
    \color{red}{\textbf{96.62$_{\pm\text{0.48}}$}}
    % &    
    % 97.08$_{\pm\text{0.11}}$
    \\
    ROTP$_{\text{B-E}}$ ($\alpha_0=0$) &
    74.83$_{\pm\text{2.07}}$
    &
    93.16$_{\pm\text{1.02}}$
    &   
    96.17$_{\pm\text{0.43}}$
    % &    
    % 97.15$_{\pm\text{0.16}}$
    \\
    ROTP$_{\text{B-Q}}$ ($\alpha_0=0$) &
    75.08$_{\pm\text{2.06}}$
    &
    93.13$_{\pm\text{0.94}}$
    &   
    96.09$_{\pm\text{0.46}}$
    % &    
    % 97.08$_{\pm\text{0.17}}$
    \\ 
    ROTP$_{\text{B-E}}$ (learn $\alpha_0$)  &
    \textbf{75.33$_{\pm\text{1.96}}$}
    &
    93.16$_{\pm\text{1.08}}$
    &   
    96.22$_{\pm\text{0.44}}$
    % &    
    % 97.06$_{\pm\text{0.15}}$
    \\
    ROTP$_{\text{B-Q}}$ (learn $\alpha_0$)  &
    \textbf{75.17$_{\pm\text{2.45}}$}
    &
    \textbf{93.45$_{\pm\text{0.96}}$}
    &   
    96.22$_{\pm\text{0.48}}$
    % &    
    % 97.04$_{\pm\text{0.18}}$
    \\ 
    \hline\hline
    \end{tabular}
    \begin{tablenotes}
    \item[*] The top-3 results are bolded and the best result is in red.
    \end{tablenotes}
\end{threeparttable}
}
\end{table}

\begin{table*}[t]
\caption{Comparison on graph classification accuracy$\pm$Std. (\%) for different pooling layers.}\label{tab:adgcl}
\centering
\small{
\begin{threeparttable}
    \begin{tabular}{c|c@{\hspace{8pt}}c@{\hspace{8pt}}c@{\hspace{8pt}}c@{\hspace{8pt}}c@{\hspace{8pt}}c@{\hspace{8pt}}c@{\hspace{8pt}}c}
    \hline\hline
    Dataset &
    NCII        & 
    PROTEINS     & 
    MUTAG       & 
    % DD  & 
    COLLAB       & 
    RDT-B       & 
    RDT-M5K        & 
    IMDB-B    & 
    IMDB-M   \\ 
    \hline
    \#Graphs &
    4,110 &
    1,113 &
    188 &
    5,000 &
    2,000 &
    4,999 &
    1,000 &
    1,500 \\
    Average \#Nodes &
    29.87 &
    39.06 &
    17.93 &
    74.49 &
    429.63 &
    508.52 &
    19.77 &
    13.00 \\
    Average \#Edges &
    32.30 &
    72.82 &
    19.79 &
    2,457.78 &
    497.75 &
    594.87 &
    96.53 &
    65.94 \\
    \#Classes &
    2 &
    2 &
    2 &
    3 &
    2 &
    5 &
    2 &
    3 \\
    \hline
    Add  &  
    67.96$_{\pm\text{0.43}}$
    & 
    72.97$_{\pm\text{0.54}}$
    & 
    \color{red}{\textbf{89.05$_{\pm\text{0.86}}$}}
    &  
    % \textbf{75.21$_{\pm\text{0.51}}$}
    % &
    71.06$_{\pm\text{0.43}}$
    & 
    80.00$_{\pm\text{1.49}}$
    &  
    50.16$_{\pm\text{0.97}}$
    &
    70.18$_{\pm\text{0.87}}$  
    & 
    47.56$_{\pm\text{0.56}}$
    \\
    Mean & 
    64.82$_{\pm\text{0.52}}$
    &
    66.09$_{\pm\text{0.64}}$
    & 
    86.53$_{\pm\text{1.62}}$
    &  
    % 63.58$_{\pm\text{0.61}}$
    % &
    72.35$_{\pm\text{0.44}}$
    & 
    83.62$_{\pm\text{1.18}}$
    &  
    52.44$_{\pm\text{1.24}}$
    &
    70.34$_{\pm\text{0.38}}$  
    & 
    48.65$_{\pm\text{0.91}}$ 
    \\
    Max  & 
    65.95$_{\pm\text{0.76}}$
    &  
    72.27$_{\pm\text{0.33}}$
    & 
    85.90$_{\pm\text{1.68}}$
    &  
    % 74.54$_{\pm\text{1.13}}$
    % &
    73.07$_{\pm\text{0.57}}$
    & 
    82.62$_{\pm\text{1.25}}$
    &
    44.34$_{\pm\text{1.93}}$
    &
    70.24$_{\pm\text{0.54}}$
    & 
    47.80$_{\pm\text{0.54}}$  
    \\
    DeepSet  &  
    66.28$_{\pm\text{0.72}}$
    & 
    \color{red}{\textbf{73.76$_{\pm\text{0.47}}$}}
    & 
    87.84$_{\pm\text{0.71}}$
    &  
    % 75.01$_{\pm\text{0.65}}$
    % &
    69.74$_{\pm\text{0.66}}$
    & 
    82.91$_{\pm\text{1.37}}$
    &  
    47.45$_{\pm\text{0.54}}$
    &
    70.84$_{\pm\text{0.71}}$ 
    & 
    48.05$_{\pm\text{0.71}}$  
    \\
    Mixed  &  
    66.46$_{\pm\text{0.74}}$
    &  
    72.25$_{\pm\text{0.45}}$
    & 
    87.30$_{\pm\text{0.87}}$
    & 
    % \color{red}{\textbf{75.54$_{\pm\text{0.87}}$}}
    % &
    73.22$_{\pm\text{0.35}}$
    & 
    84.36$_{\pm\text{2.62}}$
    & 
    46.67$_{\pm\text{1.63}}$
    &
    71.28$_{\pm\text{0.26}}$
    & 
    48.07$_{\pm\text{0.25}}$  
    \\
    GatedMixed  &
    63.86$_{\pm\text{0.76}}$
    &  
    69.40$_{\pm\text{1.93}}$
    & 
    87.94$_{\pm\text{1.28}}$
    &  
    % 70.12$_{\pm\text{2.66}}$
    % &
    71.94$_{\pm\text{0.40}}$
    & 
    80.60$_{\pm\text{3.89}}$
    &  
    44.78$_{\pm\text{4.53}}$  
    &
    70.96$_{\pm\text{0.60}}$
    & 
    48.09$_{\pm\text{0.44}}$
    \\
    Set2Set  & 
    65.10$_{\pm\text{1.12}}$
    &  
    68.61$_{\pm\text{1.44}}$
    & 
    87.77$_{\pm\text{0.86}}$
    &  
    % 66.23$_{\pm\text{0.96}}$
    % &
    72.31$_{\pm\text{0.73}}$
    & 
    80.08$_{\pm\text{5.72}}$
    &  
    49.85$_{\pm\text{2.77}}$
    &
    70.36$_{\pm\text{0.85}}$   
    & 
    48.30$_{\pm\text{0.54}}$
    \\
    Attention   & 
    64.35$_{\pm\text{0.61}}$
    & 
    67.70$_{\pm\text{0.95}}$
    & 
    88.08$_{\pm\text{1.22}}$
    &  
    % 64.28$_{\pm\text{1.16}}$
    % &
    72.57$_{\pm\text{0.41}}$
    & 
    81.55$_{\pm\text{4.39}}$
    &  
    51.85$_{\pm\text{0.66}}$
    &
    70.60$_{\pm\text{0.38}}$ 
    & 
    47.83$_{\pm\text{0.78}}$ 
    \\
    GatedAtt  &
    64.66$_{\pm\text{0.52}}$
    &  
    68.16$_{\pm\text{0.90}}$
    & 
    86.91$_{\pm\text{1.79}}$
    &  
    % 63.56$_{\pm\text{0.91}}$
    % &
    72.31$_{\pm\text{0.37}}$
    & 
    82.55$_{\pm\text{1.96}}$
    & 
    51.47$_{\pm\text{0.82}}$
    &
    70.52$_{\pm\text{0.31}}$  
    & 
    48.67$_{\pm\text{0.35}}$
    \\
    DynamicP      & 
    62.11$_{\pm\text{0.27}}$
    &  
    65.86$_{\pm\text{0.85}}$
    & 
    85.40$_{\pm\text{2.81}}$
    &  
    % 63.39$_{\pm\text{1.17}}$
    % &
    70.78$_{\pm\text{0.88}}$
    & 
    67.51$_{\pm\text{1.82}}$
    & 
    32.11$_{\pm\text{3.85}}$
    &
    69.84$_{\pm\text{0.73}}$ 
    &  
    47.59$_{\pm\text{0.48}}$
    \\
    GNP &
    \textbf{68.20$_{\pm\text{0.48}}$}
    &
    \textbf{73.44$_{\pm\text{0.61}}$}
    &
    88.37$_{\pm\text{1.25}}$
    &
    % 74.99$_{\pm\text{0.52}}$
    % &
    72.80$_{\pm\text{0.58}}$
    &
    81.93$_{\pm\text{2.23}}$
    &
    51.80$_{\pm\text{0.61}}$
    &
    70.34$_{\pm\text{0.83}}$
    &
    48.85$_{\pm\text{0.81}}$
    \\
    ASAP &
    68.09$_{\pm\text{0.42}}$
    &
    70.42$_{\pm\text{1.45}}$
    &
    87.68$_{\pm\text{1.42}}$
    &
    % 74.96$_{\pm\text{0.49}}$
    % &
    68.20$_{\pm\text{2.37}}$
    &
    73.91$_{\pm\text{1.50}}$
    &
    44.58$_{\pm\text{0.44}}$
    &
    68.33$_{\pm\text{2.50}}$
    &
    43.92$_{\pm\text{1.13}}$
    \\
    SAGP &
    67.48$_{\pm\text{0.65}}$
    &
    72.63$_{\pm\text{0.44}}$
    &
    87.88$_{\pm\text{2.22}}$
    &
    % 75.01$_{\pm\text{0.44}}$
    % &
    70.19$_{\pm\text{0.55}}$
    &
    74.12$_{\pm\text{2.86}}$
    &
    46.00$_{\pm\text{1.74}}$
    &
    70.34$_{\pm\text{0.74}}$
    &
    47.04$_{\pm\text{1.22}}$
    \\
    OTK &
    67.96$_{\pm\text{0.55}}$
    &
    69.52$_{\pm\text{0.76}}$
    &
    86.90$_{\pm\text{1.83}}$
    &
    71.35$_{\pm\text{0.91}}$
    &
    74.28$_{\pm\text{1.39}}$
    &
    50.57$_{\pm\text{1.20}}$
    &
    70.94$_{\pm\text{0.79}}$
    &
    48.41$_{\pm\text{0.89}}$
    \\
    SWE &
    68.06$_{\pm\text{0.98}}$
    &
    70.09$_{\pm\text{1.22}}$
    &
    85.68$_{\pm\text{2.07}}$
    &
    72.17$_{\pm\text{1.29}}$
    &
    79.30$_{\pm\text{3.94}}$
    &
    51.11$_{\pm\text{1.55}}$
    &
    70.34$_{\pm\text{1.05}}$
    &
    48.93$_{\pm\text{1.34}}$
    \\
    WEGL &
    \textbf{68.16$_{\pm\text{0.62}}$}
    &
    71.58$_{\pm\text{0.94}}$
    &
    \textbf{88.68$_{\pm\text{1.66}}$}
    &
    72.55$_{\pm\text{0.69}}$
    &
    82.80$_{\pm\text{1.73}}$
    &
    52.03$_{\pm\text{0.60}}$
    &
    71.94$_{\pm\text{0.75}}$
    &
    49.20$_{\pm\text{0.87}}$
    \\
    \hline
    ROTP$_{\text{S}}$  &  
    \color{red}{\textbf{68.27$_{\pm\text{1.06}}$}}
    &  
    \textbf{73.10$_{\pm\text{0.22}}$}
    & 
    \textbf{88.84$_{\pm\text{1.21}}$}
    &  
    71.20$_{\pm\text{0.55}}$
    % &
    
    & 
    81.54$_{\pm\text{1.38}}$
    &
    51.00$_{\pm\text{0.61}}$
    &
    70.74$_{\pm\text{0.80}}$
    &  
    47.87$_{\pm\text{0.43}}$
    \\
    % \color{red}{\textbf{68.27$_{\pm\text{1.06}}$}}
    % &  
    % % \textbf{73.10$_{\pm\text{0.22}}$}
    % ---
    % & 
    % \textbf{88.84$_{\pm\text{1.21}}$}
    % &  
    % ---
    % % &
    
    % & 
    % ---
    % &
    % ---
    % &
    % ---
    % &  
    % ---
    % \\
    ROTP$_{\text{B-E}}$ ($\alpha_0=0$)  &
    66.23$_{\pm\text{0.50}}$
    &  
    67.71$_{\pm\text{1.70}}$
    & 
    86.82$_{\pm\text{2.02}}$
    &
    73.86$_{\pm\text{0.44}}$
    & 
    \textbf{86.80$_{\pm\text{1.19}}$}
    &
    52.25$_{\pm\text{0.75}}$
    &
    71.72$_{\pm\text{0.88}}$
    &  
    \color{red}{\textbf{50.48$_{\pm\text{0.14}}$}}
    \\
    ROTP$_{\text{B-Q}}$ ($\alpha_0=0$)  &
    66.18$_{\pm\text{0.76}}$
    &  
    69.88$_{\pm\text{0.87}}$
    & 
    85.42$_{\pm\text{1.10}}$
    &  
    
    % &
    {\textbf{74.14$_{\pm\text{0.24}}$}}  
    & 
    {\textbf{87.72$_{\pm\text{1.03}}$}}
    &
    {\textbf{52.79$_{\pm\text{0.60}}$}}
    &
    \textbf{72.34$_{\pm\text{0.50}}$}
    &  
    49.36$_{\pm\text{0.52}}$
    \\ 
    ROTP$_{\text{B-E}}$ (learn $\alpha_0$)  &
    65.90$_{\pm\text{0.94}}$
    &  
    70.19$_{\pm\text{0.66}}$
    & 
    88.01$_{\pm\text{1.51}}$
    &
    \textbf{74.05$_{\pm\text{0.34}}$}
    & 
    86.78$_{\pm\text{1.14}}$
    &
    \textbf{52.77$_{\pm\text{0.69}}$}
    &
    \textbf{71.76$_{\pm\text{0.62}}$}
    &  
    {\textbf{50.28$_{\pm\text{0.86}}$}}
    \\
    ROTP$_{\text{B-Q}}$ (learn $\alpha_0$)  &
    65.96$_{\pm\text{0.32}}$
    &  
    70.12$_{\pm\text{1.17}}$
    & 
    86.79$_{\pm\text{1.81}}$
    &  
    \color{red}\textbf{74.27$_{\pm\text{0.47}}$}
    % &
     
    & 
    \color{red}\textbf{88.67$_{\pm\text{0.99}}$}
    &
    \color{red}\textbf{52.84$_{\pm\text{0.60}}$}
    &
   \color{red}{\textbf{71.78$_{\pm\text{1.00}}$}}
    &  
    \textbf{49.44$_{\pm\text{0.46}}$}
    \\ 
    \hline\hline
    \end{tabular}
    \begin{tablenotes}
    \item[*] For each dataset, the top-3 results are bolded and the best result is in red.
    \end{tablenotes}
\end{threeparttable}
}
\end{table*}

For each model with a specific pooling operation, we train and test it through 5-fold cross-validation. 
Accordingly, we evaluate the global pooling methods based on the averaged testing classification accuracy achieved by the corresponding models. 
Table~\ref{tab:mil} presents the statistics of the MIL datasets and the learning results. 
None of the baselines perform consistently well across all the datasets. 
Our ROTP layers outperform their competitors in most situations.
Especially, the ROTP$_{\text{S}}$ layer achieves the best performance on two of the three datasets. 
For the ROTP$_{\text{B-E/Q}}$ layers, their performance is comparable to its competitors. 
Additionally, we can find that when making $\alpha_0$ learnable, the performance of the ROTP$_{\text{B-E/Q}}$ layers is improved. 
It means that although the corresponding GW discrepancy term increases the complexity of the layer, it takes the structural information of samples into account and helps to improve the learning results indeed.

\subsubsection{Graph embedding and classification}
We further evaluate our ROTP layers in graph embedding and classification tasks. 
In this experiment, we consider eight representative graph classification datasets in the TUDataset~\cite{Morris+2020}, including three biochemical molecule datasets (NCII, MUTAG, and PROTEINS) and five social network datasets (COLLAB, RDT-B, RDT-M5K, IMDB-B, and IMDB-M). 
For each dataset, we implement the adversarial graph contrastive learning method (ADGCL)~\cite{suresh2021adversarial}, learning a five-layer graph isomorphism network (GIN)~\cite{xu2018powerful} to represent graphs. 
At the end of the GIN, we apply different global pooling methods to aggregate node embeddings as graph embeddings. 
After learning the graph embeddings, we train an SVM classifier to classify the graphs.

Following the setting used in the ADGCL work~\cite{suresh2021adversarial}, we apply learnable edge drop operations to augment observed graphs. 
For each model with a specific global pooling layer, we use the Adam optimizer to train it. 
The learning rate is 0.001, and the batch size is 32. 
We train each model with 100 epochs for the COLLAB dataset and 150 epochs for the RDT-B dataset, respectively.
For the remaining dataset, we set the number of epochs to 20. 
Following the setting used in the MIL experiment, we use four feed-forward modules to build each ROTP layer. 
We train and test each model in five trials.
The averaged classification accuracy and the standard deviation are recorded. 
Table~\ref{tab:adgcl} shows the statistics of the datasets and the learning results achieved by different global pooling methods. 
Similar to the above MIL experiment, our ROTP layers perform well in most situations. 
Especially our BADMM-based ROTP layers achieve the best performance on the five social network datasets. 
Note that the graph structure plays a central role in graph classification tasks. 
Therefore, we should take the GW discrepancy term into account when applying our ROTP layers. 
The results in Table~\ref{tab:adgcl} support our claim --- making $\alpha_0$ learnable improves the learning results in most situations.

The experiments on MIL and graph classification indicate that our ROTP layers can simplify the design and selection of global pooling to some degree. 
In particular, none of the baselines perform consistently well across all the datasets, while our ROTP layers are comparable to the best baselines in most situations, whose performance is more stable and consistent. 
Applying our ROTP layers, the design and selection of pooling operations are reformulated as the selection of optimization algorithms.  
Instead of testing various global pooling methods empirically, we just need to select an algorithm (i.e., Sinkhorn-scaling or Bregman ADMM) to implement the ROTP layer, which can achieve encouraging performance. 

\begin{table}[t]
\caption{Comparisons on graph set classification accuracy$\pm$Std. (\%) for different pooling layers.}\label{tab:drugs}
\centering
\small{
\begin{threeparttable}
    \begin{tabular}{@{}c|c@{\hspace{4pt}}c@{\hspace{4pt}}c@{\hspace{4pt}}c@{}}
    \hline\hline
    \multirow{2}{*}{Dataset} & 
    DECAGON    &
    DECAGON    &
    DECAGON    &
    \multirow{2}{*}{FEARS}\\
    & 
    DiBr-APND      &
    Anae-Fati      &
    PleuP-Diar     &
    \\
    \hline
    \#Graph sets &
    6,309 &
    2,922 &
    2,842 &
    6,338 \\
    \#Positive sets &
    3,189 &
    1,526 &
    1,422 &
    3,169 \\
    \multirow{2}{*}{Positive label} &
    Difficulty &
    \multirow{2}{*}{Anaemia} &
    Pleural &
    Non- \\
    &
    breathing &
    &
    pain &
    myopathy\\
    \#Negative sets &
    3,120 &
    1,396 &
    1,420 &
    3,169 \\
    \multirow{2}{*}{Negative label} &
    Pressure &
    \multirow{2}{*}{Fatigue} &
    \multirow{2}{*}{Diarrhea} &
    \multirow{2}{*}{Myopathy} \\
    &
    decreased &
    &
    &
    \\
    Set size &
    2 &
    2 &
    2 &
    2$\sim$52\\
    \hline
    Add  &
    50.86$_{\pm\text{0.97}}$
    &
    63.15$_{\pm\text{1.79}}$
    &
    \textbf{62.32$_{\pm\text{1.08}}$}
    &
    75.89$_{\pm\text{1.33}}$
    \\
    Mean &
    51.10$_{\pm\text{1.09}}$
    &
    61.95$_{\pm\text{2.60}}$
    &
    61.30$_{\pm\text{2.68}}$
    &
    72.42$_{\pm\text{1.51}}$
    \\
    Max  & 
    50.59$_{\pm\text{0.77}}$
    &
    61.88$_{\pm\text{2.03}}$
    &
    60.11$_{\pm\text{2.03}}$    
    &
    \textbf{82.02$_{\pm\text{0.72}}$}
    \\
    DeepSet  &
    49.83$_{\pm\text{1.07}}$
    &
    56.24$_{\pm\text{5.20}}$
    &
    51.78$_{\pm\text{3.10}}$
    &
    \textbf{82.40$_{\pm\text{1.56}}$}
    \\
    Mixed & 
    51.13$_{\pm\text{0.99}}$
    &
    \textbf{63.83$_{\pm\text{1.19}}$}
    &
    60.91$_{\pm\text{2.12}}$
    &
    81.54$_{\pm\text{1.13}}$
    \\
    GatedMixed  &
    51.39$_{\pm\text{0.63}}$
    &
    61.50$_{\pm\text{1.61}}$
    &
    59.12$_{\pm\text{2.12}}$
    &
    81.88$_{\pm\text{1.14}}$
    \\
    Set2Set  & 
    50.72$_{\pm\text{1.71}}$
    &
    59.35$_{\pm\text{2.04}}$
    &
    55.01$_{\pm\text{3.59}}$
    &
    79.29$_{\pm\text{0.84}}$
    \\
    Attention   &
    50.52$_{\pm\text{1.10}}$
    &
    61.40$_{\pm\text{2.03}}$
    &
    61.33$_{\pm\text{2.40}}$
    &
    75.98$_{\pm\text{0.74}}$
    \\
    GatedAtt   &
    50.74$_{\pm\text{0.61}}$
    &
    62.15$_{\pm\text{0.77}}$
    &
    58.80$_{\pm\text{1.18}}$
    &
    75.84$_{\pm\text{1.29}}$
    \\
    DynamicP      & 
    51.01$_{\pm\text{1.88}}$
    &
    55.93$_{\pm\text{1.56}}$
    &
    52.58$_{\pm\text{2.91}}$
    &
    74.00$_{\pm\text{1.61}}$
    \\
    GNP      &
    50.00$_{\pm\text{1.88}}$
    &
    53.98$_{\pm\text{6.34}}$
    &
    52.58$_{\pm\text{4.68}}$
    &
    62.71$_{\pm\text{15.55}}$
    \\
    ASAP      & 
    50.89$_{\pm\text{0.82}}$
    &
    63.66$_{\pm\text{1.81}}$
    &
    60.67$_{\pm\text{2.69}}$
    &
    77.15$_{\pm\text{1.13}}$
    \\
    SAGP      &
    49.87$_{\pm\text{0.77}}$
    &
    63.62$_{\pm\text{1.28}}$
    & 
    59.86$_{\pm\text{2.43}}$
    &
    77.29$_{\pm\text{1.04}}$
    \\
    OTK &
    50.96$_{\pm\text{1.11}}$
    &
    63.68$_{\pm\text{1.59}}$
    & 
    \textbf{61.66$_{\pm\text{2.39}}$}
    &
    79.40$_{\pm\text{1.08}}$
    \\
    SWE &
    51.05$_{\pm\text{2.15}}$
    &
    63.21$_{\pm\text{2.02}}$
    & 
    61.37$_{\pm\text{3.13}}$
    &
    80.64$_{\pm\text{1.86}}$
    \\
    WEGL &
    \textbf{51.67$_{\pm\text{0.85}}$}
    &
    \textbf{63.79$_{\pm\text{2.54}}$}
    & 
    61.36$_{\pm\text{2.30}}$
    &
    81.98$_{\pm\text{0.77}}$
    \\
    \hline
    ROTP$_{\text{S}}$  &
    \textbf{51.96$_{\pm\text{0.71}}$}
    &
    62.91$_{\pm\text{1.13}}$
    &
    59.40$_{\pm\text{0.90}}$
    &
    79.75$_{\pm\text{0.71}}$
    \\
    ROTP$_{\text{B-E}}$ 
    &
    51.26$_{\pm\text{0.84}}$
    &
    \color{red}{\textbf{63.86$_{\pm\text{2.41}}$}}
    &
    \color{red}{\textbf{62.57$_{\pm\text{1.34}}$}}
    &
    \color{red}{\textbf{82.55$_{\pm\text{0.42}}$}}
    \\ 
    ROTP$_{\text{B-Q}}$  &
    \color{red}{\textbf{52.72$_{\pm\text{0.66}}$}}
    &
    63.15$_{\pm\text{1.27}}$
    &
    60.88$_{\pm\text{1.65}}$
    &
    81.43$_{\pm\text{1.12}}$
    \\ 
    \hline\hline
    \end{tabular}
    \begin{tablenotes}
    \item[*] The top-3 results are bolded and the best result is in red.
    \end{tablenotes}
\end{threeparttable}
}
\end{table}

\subsection{Evaluation of HROTP modules}
To demonstrate the usefulness of the proposed HROTP modules, we apply them in drug-drug interaction (DDI) classification task~\cite{zhang2016drug,zhao2018similarity,deepika2018meta,peng2019deep}. 
In particular, the DDI classification task aims to predict different drug combinations' side effects.
Each drug is represented as a molecular graph, and a drug combination corresponds to a set of graphs. 
Therefore, the DDI classification task is a typical graph set prediction problem whose data have hierarchical clustering structures. 

In this experiment, we consider four drug side-effect datasets: three drug pair datasets (DiBr-APND, Anae-Fati, and PleuP-Diar) sampled from DECAGON~\cite{biosnapnets} and a drug combination dataset FEARS~\cite{peng2019deep}. 
Each dataset contains thousands of drug sets that may cause two different side effects. 
We implement a three-layer GIN model for each dataset to learn graph embeddings. 
A hierarchical pooling mechanism with two pooling layers is considered. 
The first pooling layer aggregates the node embeddings of each graph into a graph embedding.
The second pooling layer further aggregates the graph embeddings within a drug set to a set embedding. 
Finally, a classifier is trained based on the set embeddings. 
For the hierarchical pooling module, we implement its pooling layers based on different pooling methods. 
When using our ROTP layers, we obtain the proposed HROTP modules. 

We apply the Adam optimizer to train the models and set the learning rate to 0.001.
For the DiBr-APND and the FEARS, we train each model with 40 epochs.
For the remaining two datasets, we train each model with 100 epochs. 
The batch size is set to be 128 for the FEARS dataset and 32 for the remaining three datasets.
In this experiment, we set $\alpha_0=0$ for all three ROTP layers to achieve a trade-off between effectiveness and efficiency. 
The number of the feed-forward modules used in the ROPT layers is set from $\{4, 6, 8, 10\}$.
We train and test each model in five trials for each dataset to get the average classification accuracy. 
Table~\ref{tab:drugs} shows the statistics of the datasets and the learning results achieved by different global pooling methods.
Our HROTP modules achieve the best performance on all four datasets.

\begin{figure}[t]
    \centering
    \includegraphics[height=6.5cm]{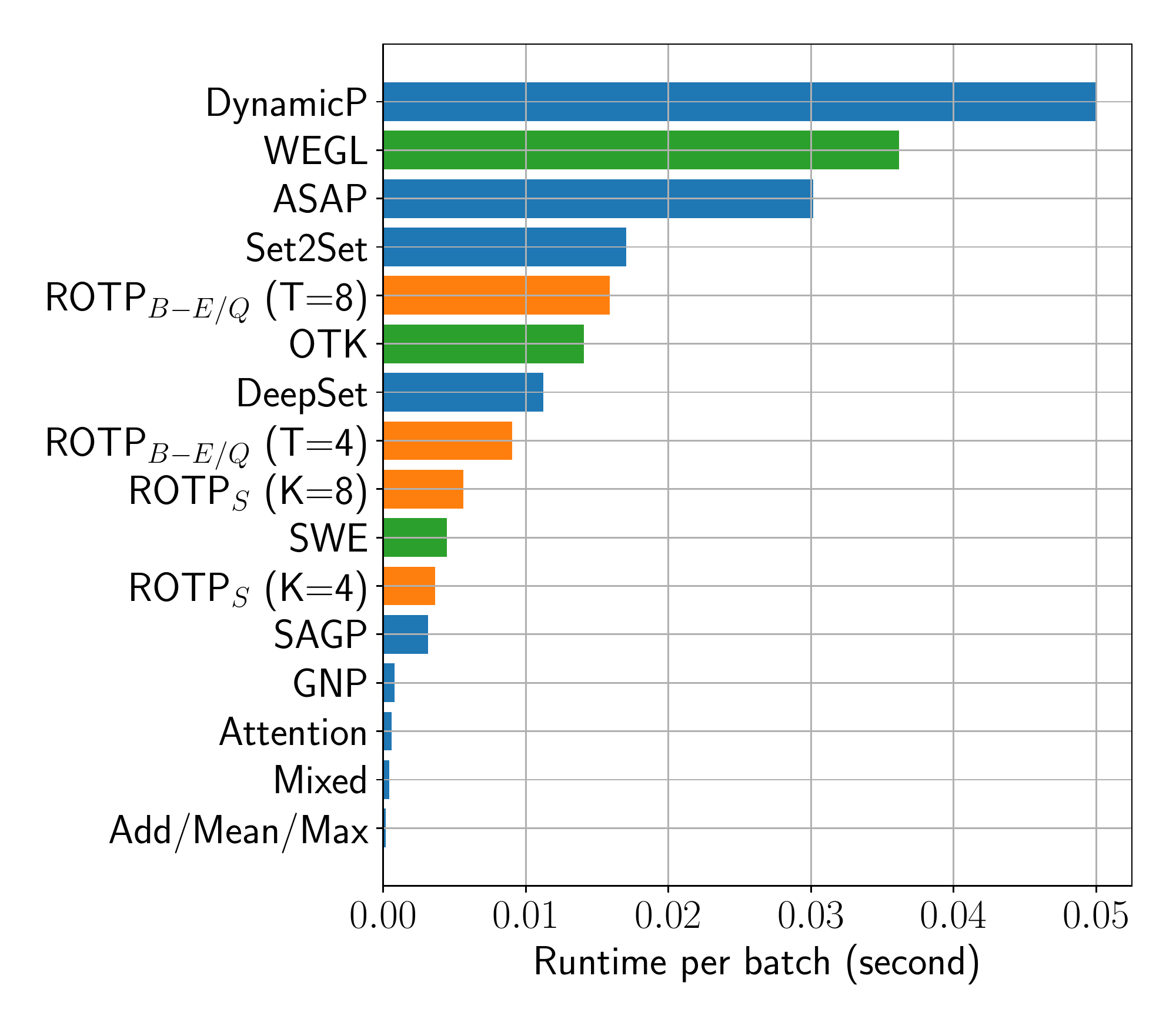}
    \caption{
    Given a batch of 50 sample sets, in which each sample set contains five hundred 100-dimensional samples, we plot the averaged feed-forward runtime of various pooling methods in 10 trials on a single GPU (RTX 3090).
    The proposed ROTP layers and existing OT-based methods are labeled in orange and green, respectively. 
    The remaining global pooling methods are labeled in blue.
    }
    \label{fig:allruntimes}
\end{figure}

\subsection{More analytic experiments}
\subsubsection{Runtime comparison}
The runtime of our ROTP layers is comparable to that of the learning-based global pooling methods (including existing OT-based methods).
Fig.~\ref{fig:allruntimes} shows the rank of various global pooling methods on their runtime per batch. 
We can find that when applying eight feed-forward modules, the runtime of the BADMM-based ROTP layer is almost the same as that of Set2Set~\cite{vinyals2015order}. 
When reducing the number of the feed-forward modules to four, its runtime can be less than that of DeepSet~\cite{zaheer2017deep}.
Because of setting $\alpha_0=0$, the Sinkhorn-based ROTP layer in this experiment is faster than the BADMM-based ROTP layers, which verifies the analysis shown in Section~\ref{sec:cmp}. 
When applying four Sinkhorn-scaling modules, the runtime of the Sinkhorn-based ROTP layer is comparable to that of SAGP~\cite{lee2019self}. 
Note that the BADMM-based ROTP layers are comparable to OTK~\cite{mialon2020trainable}, and the Sinkhorn-based ROTP layer is comparable to SWE~\cite{naderializadeh2021pooling} in the aspect of runtime.
All the ROTP layers are faster than the WEGL~\cite{kolouri2020wasserstein}. 

\subsubsection{Robustness to hyperparameter settings} 
\begin{figure}[t]
    \centering
    \includegraphics[height=4.5cm]{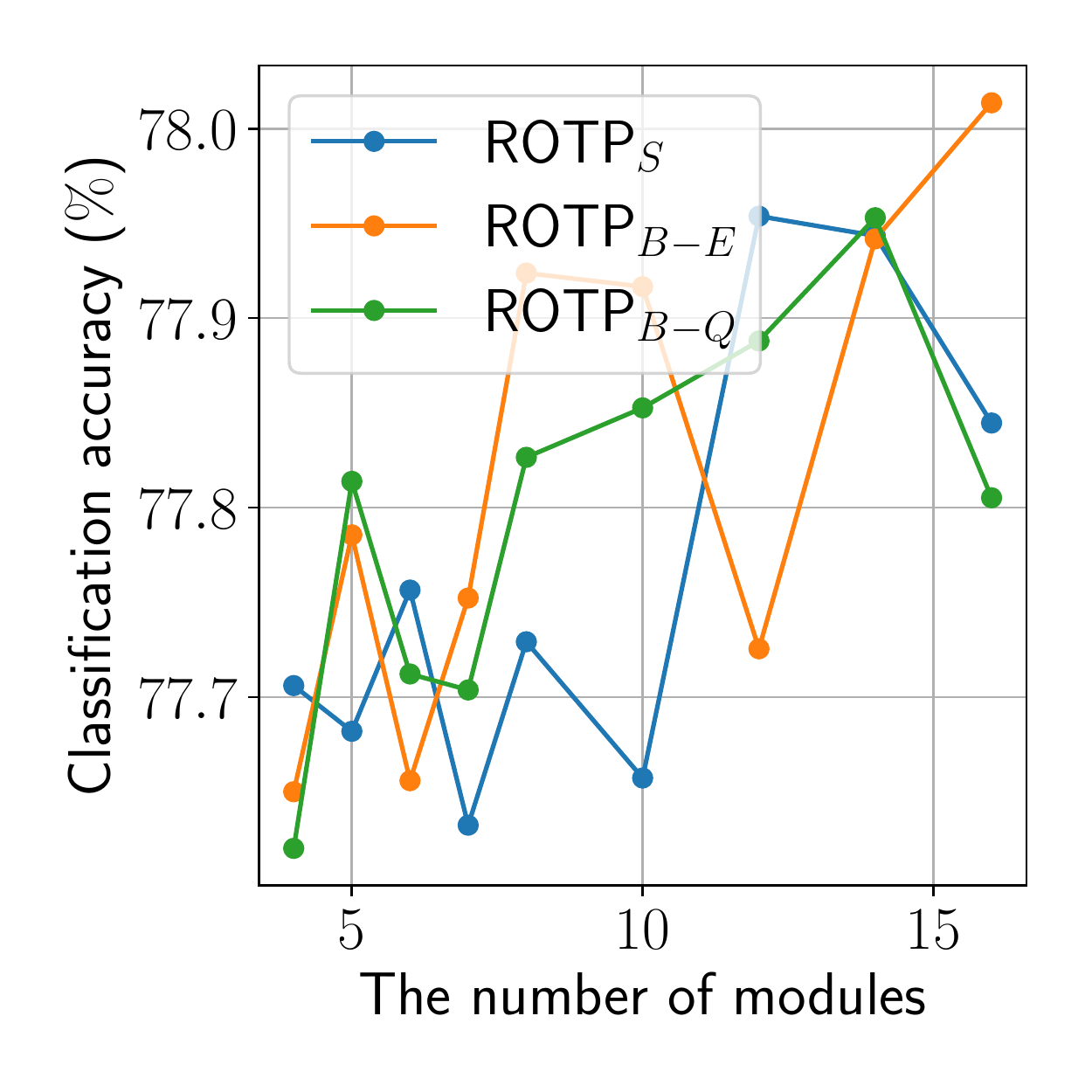}
    \caption{
    The averaged classification accuracy for the twelve datasets (MIL and Graph Classification) achieved by our ROTP layers with different numbers of feed-forward modules.
    }
    \label{fig:acc}
\end{figure}

Our ROTP layers have one key hyperparameter --- the number of feed-forward modules. 
Applying many modules will lead to highly-precise solutions to~\eqref{eq:uot} but take more time on both feed-forward computation and backpropagation. 
As aforementioned, we search the number of the modules in the range $[4, 16]$ in the above experiments, which can achieve a good trade-off between effectiveness and efficiency. 
To further demonstrate the robustness of our ROTP layers to the number of feed-forward modules, we consider the ROTP layers with 4-16 feed-forward modules and train the corresponding models on each of the twelve (MIL and graph classification) datasets. 
Fig.~\ref{fig:acc} shows the averaged classification accuracy on the twelve datasets with respect to the number of feed-forward modules. 
The performance of our ROTP layers is stable --- the dynamics of the average classification accuracy is smaller than 0.4\%.

\begin{table}[t]
\caption{The impacts of $\bm{p}_0$ and $\bm{q}_0$ on the classification accuracy (\%) of NCI1}\label{tab:prior}
\centering
\begin{small}
\begin{threeparttable}
    \begin{tabular}{cc|ccc}
    \hline\hline
    $\bm{p}_0$ & 
    $\bm{q}_0$ & 
    ROTP$_{\text{S}}$ &
    ROTP$_{\text{B-E}}$ &
    ROTP$_{\text{B-Q}}$
    \\
    \hline
    \multirow{4}{*}{} 
    Fixed
    & Fixed 
    & 68.27$_{\pm\text{1.06}}$
    & 65.90$_{\pm\text{0.94}}$
    & 65.96$_{\pm\text{0.32}}$
    \\
    Learned
    & Fixed
    & 67.97$_{\pm\text{0.48}}$
    & 66.57$_{\pm\text{0.54}}$
    & 66.45$_{\pm\text{0.82}}$
    \\
    Fixed
    & Learned 
    & 69.86$_{\pm\text{0.45}}$
    & 66.21$_{\pm\text{0.76}}$
    & 66.40$_{\pm\text{0.57}}$
    \\
    Learned
    & Learned 
    & 68.60$_{\pm\text{0.15}}$
    & 66.45$_{\pm\text{0.23}}$
    & 66.67$_{\pm\text{0.63}}$
    \\
    \hline\hline
    \end{tabular}
    \begin{tablenotes}
    \item[*] Each layer has four feed-forward modules.
    \end{tablenotes}
\end{threeparttable}
\end{small}
\end{table}

Besides the number of the feed-forward modules, we also consider the settings of the prior distributions ($i.e.$, $\bm{p}_0$ and $\bm{q}_0$). 
As mentioned in Section~\ref{sec:alg}, we can fix them as uniform distributions or learn them by a self-attention model. 
Take the NCI1 dataset as an example. 
Table~\ref{tab:prior} presents the learning results of our methods under different settings of $\bm{p}_0$ and $\bm{q}_0$. 
Our ROTP layers are robust to their settings --- the learning results do not change a lot under different settings. 
Therefore, we fix $\bm{p}_0$ and $\bm{q}_0$ as uniform distributions in the above experiments. 
Under this simple setting, our ROTP layers have already achieved encouraging results.

\begin{figure}[t]
    \centering
    \includegraphics[height=5.5cm]{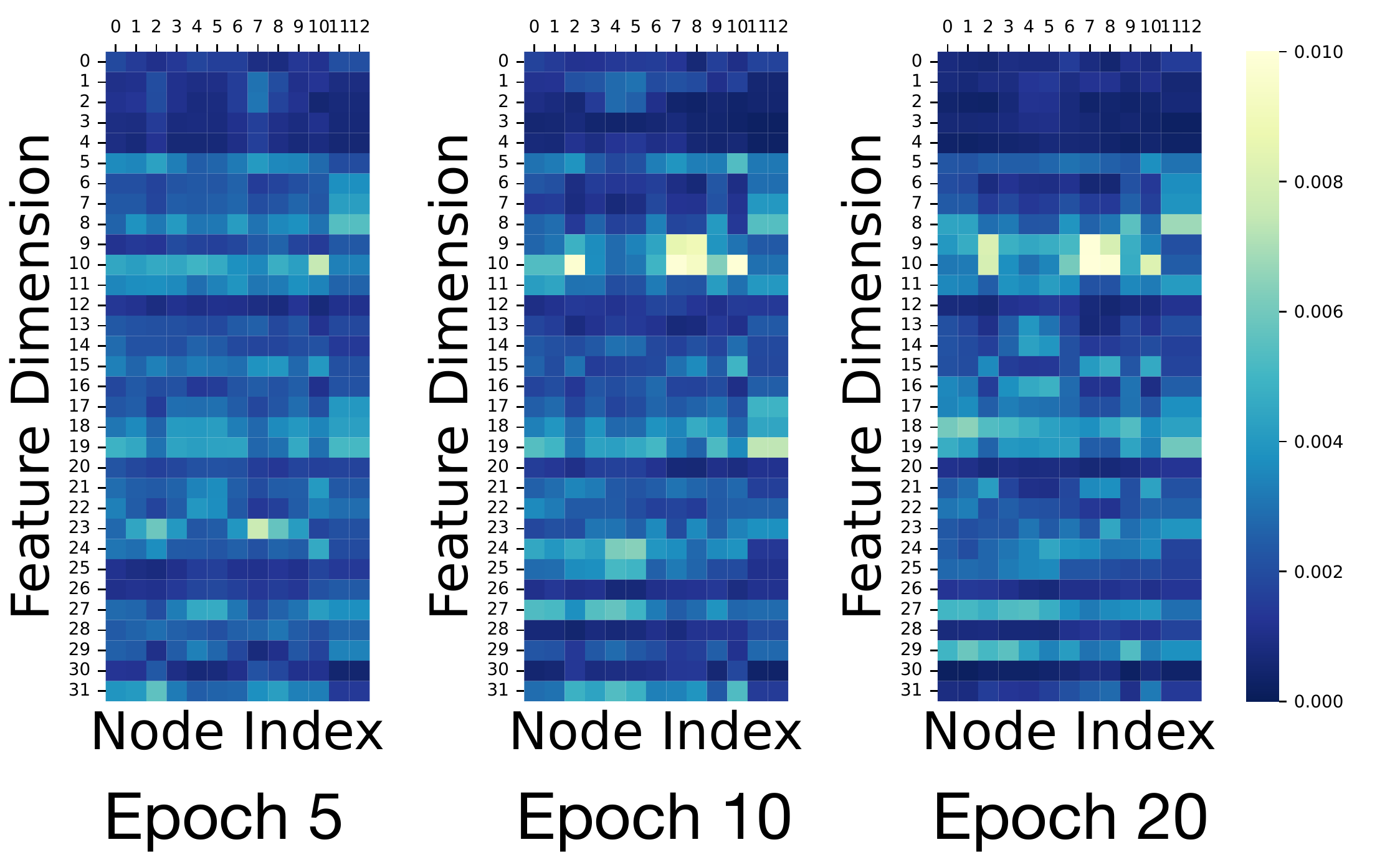}
    \caption{Illustrations of the $\bm{P}^*$'s of a MUTAG graph's node embeddings during training.}
    \label{fig:visual_p}
\end{figure}

\begin{figure}[t]
    \centering
    \subfigure[MUTAG]{
    \includegraphics[height=5.5cm]{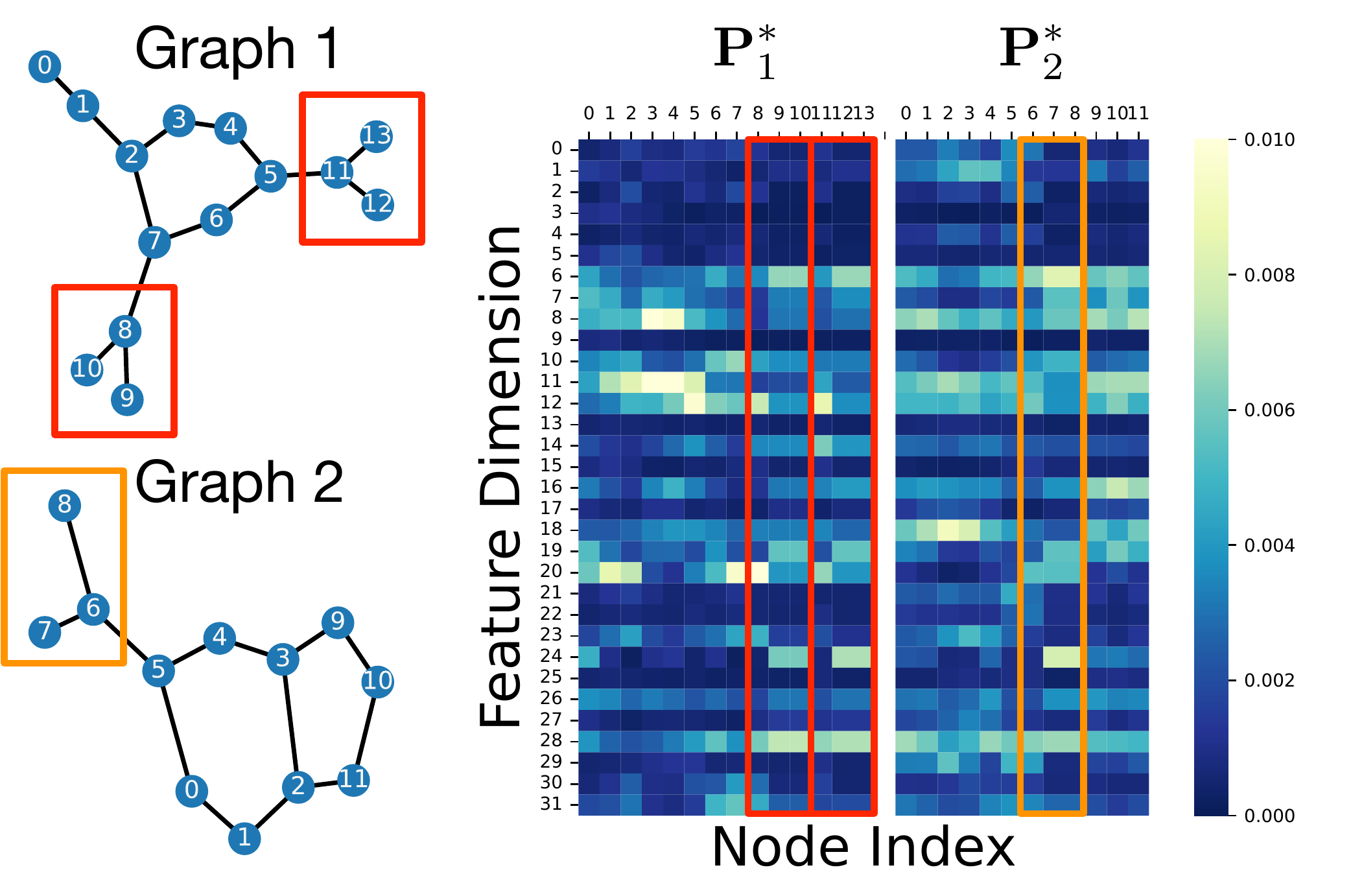}\label{fig:ration1}
    }
    \subfigure[IMDB-B]{
    \includegraphics[height=5.5cm]{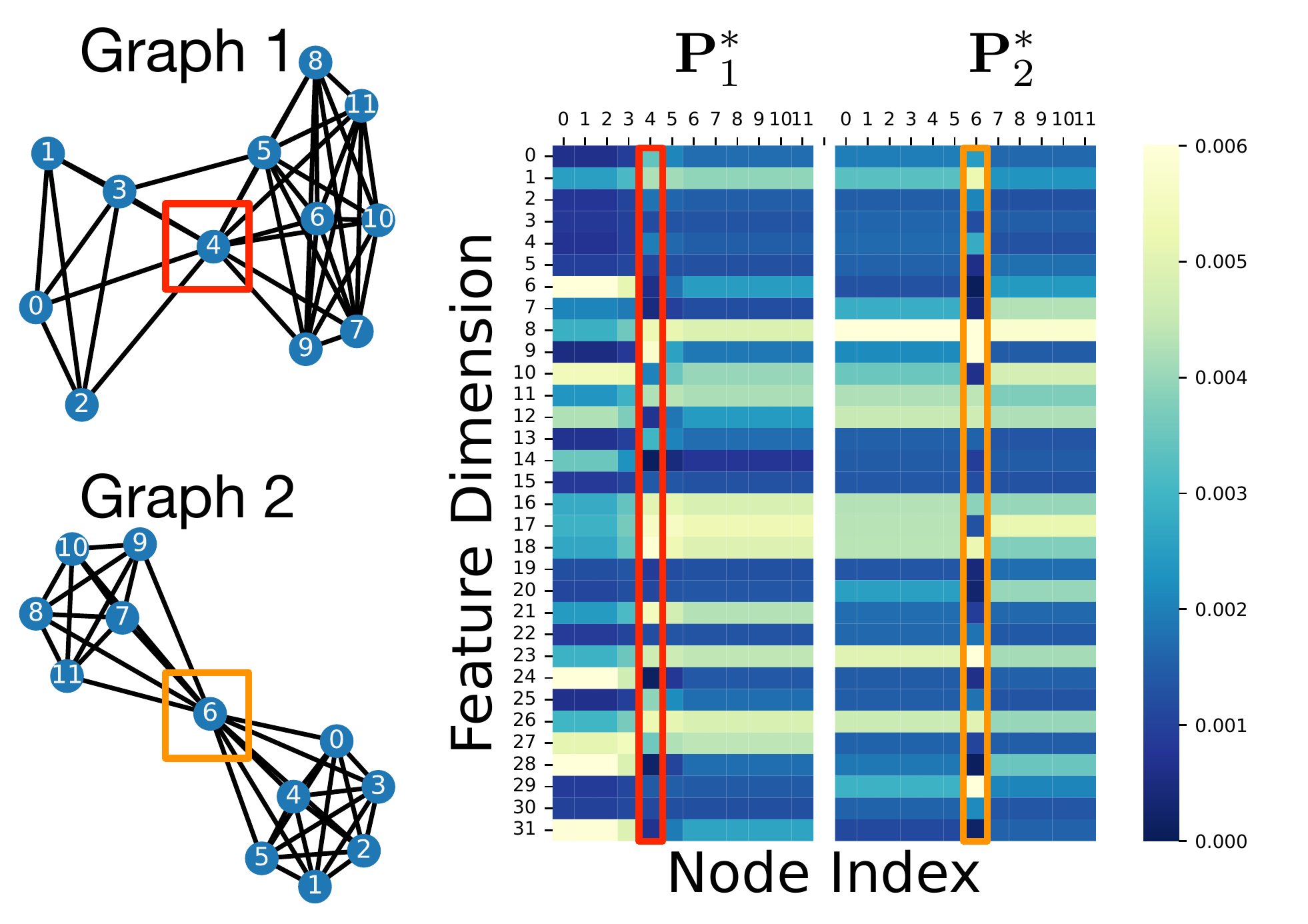}\label{fig:ration2}
    }
    \caption{
    (a) The visualizations of two MUTAG graphs and their $\bm{P}^*$'s. 
    For the ``V-shape'' subgraphs, their submatrices in the $\bm{P}^*$'s are marked by color frames.
    (b) The visualizations of two IMDB-B graphs and their $\bm{P}^*$'s. 
    For each graph, its key node connecting two communities and the corresponding column in the $\bm{P}^*$'s are marked by color frames.}\label{fig:ration}
\end{figure}

\begin{table*}[t]
    \caption{Comparisons for ResNets and our ResNets + ROTP$_{\text{B-E}}$ on validation accuracy (\%)}
    \label{tab:resnet}
    \centering
    \small{
    \begin{threeparttable}   
    \begin{small}
    \begin{tabular}{c|c|ccccc}
    \hline\hline
                   &Learning Strategy
                   &ResNet18
                   &ResNet34
                   &ResNet50
                   &ResNet101
                   &ResNet152\\
    \hline
    \multirow{2}{*}{Top-5}
    &100 Epochs (A2DP)
    &89.084
    &91.433
    &92.880
    &93.552 
    &94.048 \\
    &90 Epochs (A2DP) + 10 Epochs (ROTP$_{\text{B-E}}$)
    &\textbf{89.174}
    &\textbf{91.458}
    &\textbf{93.006}
    &\textbf{93.622}
    &\textbf{94.060}\\
    \hline
    \multirow{2}{*}{Top-1}
    &100 Epochs (A2DP)
    &69.762
    &73.320
    &76.142
    &77.386
    &78.324\\
    &90 Epochs (A2DP) + 10 Epochs (ROTP$_{\text{B-E}}$)
    &\textbf{69.906}
    &\textbf{73.426}
    &\textbf{76.446}
    &\textbf{77.522}
    &\textbf{78.446}\\
    \hline\hline
    \end{tabular}
    \end{small}
    \end{threeparttable}  
    }
\end{table*}

\subsubsection{Visualization and rationality}

Take the ROTP$_{\text{B-E}}$ layer used for the MUTAG dataset as an example. 
For a graph in the validation set, we visualize the dynamics of the corresponding $\bm{P}^*$'s in different epochs in Fig.~\ref{fig:visual_p}. 
In the beginning, the $\bm{P}^*$ is relatively dense because the node embeddings are not fully trained and may not be distinguishable.
With the increase of epochs, the $\bm{P}^*$ becomes sparse and focuses more on significant ``sample-feature'' pairs. 

Additionally, to verify the rationality of the learned $\bm{P}^*$, we take the ROTP$_{\text{S}}$ layer as an example and visualize some graphs and their $\bm{P}^*$'s in Fig.~\ref{fig:ration}.
For the ``V-shape'' subgraphs in the two MUTAG graphs, we compare the corresponding submatrices shown in their $\bm{P}^*$'s. 
These submatrices obey the same pattern, which means that for the subgraphs shared by different samples, the weights of their node embeddings will be similar. 
For the key nodes in the two IMDB-B graphs, their corresponding columns in the $\bm{P}^*$'s are distinguished from other columns. 
For the nodes belonging to different communities, their columns in the $\bm{P}^*$'s own significant clustering structures. 

\subsection{Limitations and discussions}
Besides MIL and graph classification, we further test our ROTP layers in image classification tasks.
Given a ResNet~\cite{he2016deep}, we replace its ``adaptive 2D mean-pooling layer (A2DP)'' with our ROTP$_{\text{B-E}}$ layer and finetune the modified model on ImageNet~\cite{deng2009imagenet}. 
In particular, given the output of the last convolution layer of the ResNet, i.e., $\bm{X}_{\text{in}}\in \mathbb{R}^{B\times C\times H\times W}$, our ROTP$_{\text{B-E}}$ layer fuses the data and outputs $\bm{X}_{\text{out}}\in \mathbb{R}^{B\times C\times 1\times 1}$. 
In this experiment, we apply a two-stage learning strategy: we first train a ResNet in 90 epochs, and then we replace its A2DP layer with our ROTP$_{\text{B-E}}$ layer; finally, we fix other layers and train our ROTP$_{\text{B-E}}$ layer in 10 epochs.
The learning rate is 0.001, and the batch size is 256. 
Because training on ImageNet is time-consuming, we set $\alpha_0=0$ for the ROTP$_{\text{B-E}}$ layer in this experiment to reduce the computational complexity.
Table~\ref{tab:resnet} shows that using our ROTP$_{\text{B-E}}$ layer helps to improve the classification accuracy, and the improvement is consistent for different ResNet architectures.

The improvements shown in Table~\ref{tab:resnet} are incremental because we just replaced a single global pooling layer with our ROTP layer. 
When training the ResNets with ROTP layers from scratch, the improvements are not so significant, either --- after training ``ResNet18+ROTP'' with 100 epochs, the top-1 accuracy is 69.920\%, and the top-5 accuracy is 89.198\%. 
Replacing more local pooling layers with our ROTP layers may bring better performance. 
However, given a tensor $\bm{X}_{\text{in}}\in \mathbb{R}^{B\times C\times H\times W}$, a local pooling merges each patch with size $(B\times C\times 2\times 2)$ and outputs $\bm{X}_{\text{out}}\in \mathbb{R}^{B\times C\times \frac{H}{2}\times \frac{W}{2}}$, which involves $\frac{BHW}{4}$ pooling operations.  
In other words, the current bottleneck of our ROTP layer is its computational efficiency. 
Developing an efficient CUDA version of the ROTP layers and extending them to local pooling operations will be our future work.